\documentclass{article} % For LaTeX2e
\usepackage{iclr2025_conference,times}

\usepackage{amsmath,amsfonts,bm}

\def\eqref#1{equation~\ref{#1}}
\def\1{\bm{1}}

\DeclareMathAlphabet{\mathsfit}{\encodingdefault}{\sfdefault}{m}{sl}
\SetMathAlphabet{\mathsfit}{bold}{\encodingdefault}{\sfdefault}{bx}{n}

\usepackage{enumitem}
\usepackage{hyperref}
\usepackage{url}
\usepackage{graphicx}
\usepackage{wrapfig}
\usepackage{xcolor}
\definecolor{mygreen}{RGB}{11,200,11}
\definecolor{mygreen2}{RGB}{0,150,0}
\usepackage{booktabs}
\usepackage{subfigure}
\usepackage{multirow}
\title{UniCBE: An Uniformity-driven Comparing Based Evaluation Framework with Unified Multi-Objective Optimization}

\author {
    \textbf{Peiwen Yuan}\textsuperscript{\rm 1}, \hspace{0.4cm}
    \textbf{Shaoxiong Feng}\textsuperscript{\rm 2}, \hspace{0.4cm}
    \textbf{Yiwei Li}\textsuperscript{\rm 1}, \hspace{0.4cm}
    \textbf{Xinglin Wang}\textsuperscript{\rm 1}, \hspace{0.4cm} 
    \textbf{Yueqi Zhang}\textsuperscript{\rm 1}, \hspace{0.4cm}\\
    \textbf{Jiayi Shi}\textsuperscript{\rm 1}, \hspace{0.4cm}
    \textbf{Chuyi Tan}\textsuperscript{\rm 1}, \hspace{0.4cm}
    \textbf{Boyuan Pan}\textsuperscript{\rm 2}, \hspace{0.4cm} 
    \textbf{Yao Hu}\textsuperscript{\rm 2}, 
    \hspace{0.4cm} \textbf{Kan Li}\textsuperscript{\rm 1}\footnotemark[3] \\
    \textsuperscript{\rm 1} School of Computer Science, Beijing Institute of Technology \\
    \textsuperscript{\rm 2} Xiaohongshu Inc \\
    \texttt{\{peiwenyuan,liyiwei,wangxinglin,zhangyq\}@bit.edu.cn} \\
    \texttt{\{shijiayi, tanchuyi, likan\}@bit.edu.cn} \\
    \texttt{\{shaoxiongfeng2023\}@gmail.com}  \quad \texttt{\{panboyuan,xiahou\}@xiaohongshu.com}
}

\iclrfinalcopy % Uncomment for camera-ready version, but NOT for submission.

\begin{document}

\maketitle
\renewcommand{\thefootnote}{\fnsymbol{footnote}} 
\footnotetext[3]{Corresponding author.} 

\begin{abstract}
%Human preference在衡量和引导LLM更好地与人类对齐方面发挥了重要作用。然而当前的COMPARING BASED EVALUATION方法通常只关注单一的目标而无法有效地利用珍贵的preference signals。为此，我们深入分析了提升CBE的准确性、收敛性和可扩展性的关键分别在于抑制采样bias、平衡不确定性下降过程和削弱updating uncertainty。遵循这些guidelines，我们提出了UniCBE方法，其通过解耦地构建三个采样概率矩阵并进行集成来平衡优化三个核心目标。我们还消融了最优的tuple sampling和PREFERENCE AGGREGATION策略，从而实现高效的CBE。在AlpacaEval上，UniCBE saves over 17% of evaluation budgets when 达到了与ground truth超过0.995的皮尔逊一致性，展现了良好的准确性与收敛性。在有新模型持续加入的场景中，UniCBE can even save over 50% of evaluation costs，showcasing excellent scalability.
Human preference plays a significant role in measuring large language models and guiding them to align with human values. Unfortunately, current comparing-based evaluation (CBE) methods typically focus on a single optimization objective, failing to effectively utilize scarce yet valuable preference signals. To address this, we delve into key factors that can enhance the accuracy, convergence, and scalability of CBE: suppressing sampling bias, balancing descending process of uncertainty, and mitigating updating uncertainty.
Following the derived guidelines, we propose \textsc{UniCBE}, a unified uniformity-driven CBE framework which simultaneously optimize these core objectives by constructing and integrating three decoupled sampling probability matrices, each designed to ensure uniformity in specific aspects. We further ablate the optimal tuple sampling and preference aggregation strategies to achieve efficient CBE.
On the AlpacaEval benchmark, \textsc{UniCBE} saves over 17\% of evaluation budgets while achieving a Pearson correlation with ground truth exceeding 0.995, demonstrating excellent accuracy and convergence. In scenarios where new models are continuously introduced, \textsc{UniCBE} can even save over 50\% of evaluation costs, highlighting its improved scalability.
\end{abstract}
\section{Introduction}

% arena chatbot
% RLHF
% 基于概率

% 语言模型能力的持续进化使得评估它们与人类期望之间的对齐性变得愈发重要而困难。一方面，人类提供的评估信号是正确衡量并引导模型通往安全、可靠的AGI的核心依据；另一方面，模型在训练和应用场景的快速迭代催生了大量的评估需求，这使得获取足够的劳动密集型的人类preference变得困难。因此，高效地利用珍贵的人类preference来评估模型是一项重要的挑战。
The ongoing evolution of large language models (LLMs) has made it increasingly important to assess their alignment with human preferences \citep{alpacaeval,judging}. 
The preference signals provided by humans are crucial for accurately assessing and guiding models toward safe and reliable AGI \citep{safe,survey}. 
However, the rapid iteration of LLMs in training and application scenarios has created a substantial demand for evaluation, complicating the acquisition of sufficient labor-intensive human preferences \citep{arena,ultra}. 
Therefore, exploring the use of precious preference signals for efficient model alignment evaluation is of great significance and requires long-term research.

% scoring-based 方法和comparing-based 方法是当前两类主流评估模型能力的范式。其中前者需要judger每次判断单个模型response的质量而后者要求judger每次给出多个候选模型responses间的质量比较结果。
% 由于能够直接对不同模型生成的内容进行细粒度的对比，\citet{judging, role} 证实在judger的评估能力一定的情况下，CBE能够更准确地衡量模型能力。
% However，当有M条数据和N个待评估模型时，O(MN^2)的评测开销限制了CBE的实用性\cite{all}。为此，一些研究\cite{arena，alpacaeva, speeding}探索了基于特定优化目标的[待比较模型，样本]组合采样策略。
% 它们在各自优化目标的指导下根据已有的观测结果来选择(待评测模型，样本)组合，并记录judge在该组合上的preference结果，并重复该过程。
% 模型能力的估值可以通过特定的preference aggregation方式作用于preference的观测结果来获得
% 以期在更少的评测开销下实现高效的CBE。（selector）
% Nevertheless, 我们分析发现这些方法所设定的优化目标往往是单一的，无法兼顾所获得评测结果的准确性、稳定性和可扩展性，并通过实验进行了证实。

\begin{figure}[ht]
\begin{center}
\includegraphics[width=0.9\textwidth]{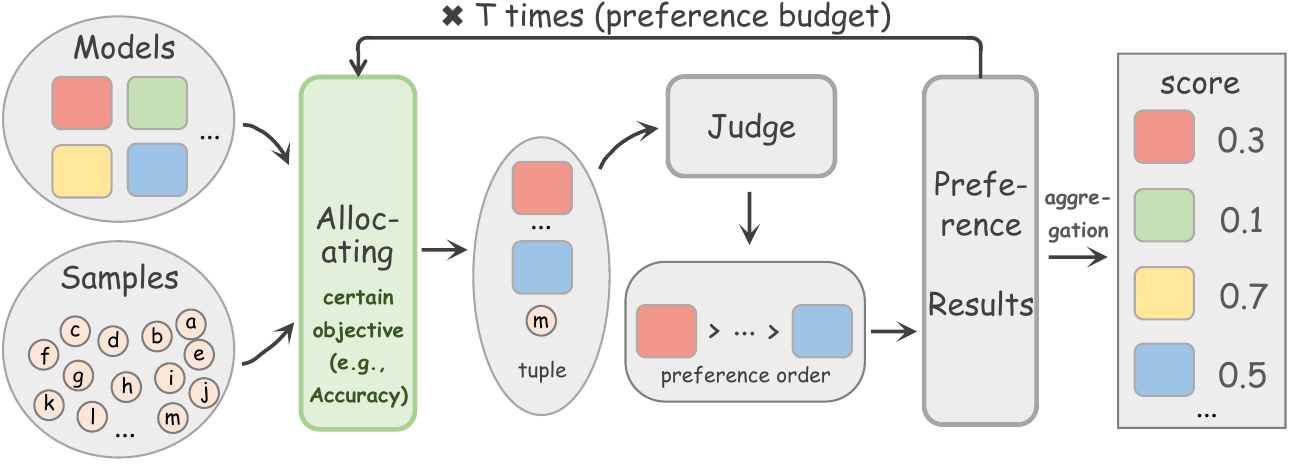}
\end{center}
\vspace{-0.2cm}
\caption{Flowchart of the process for comparing-based evaluation. }
\vspace{-0.2cm}
\label{fig:intro}
\end{figure}

Current mainstream model evaluation paradigms include scoring-based evaluation (SBE) \citep{geval,pointwise} and comparing-based evaluation (CBE) \citep{arena,alpacaeval}. The former requires the judge to offer preference scores for individual responses, while the latter needs the judge to establish a preference order among multiple candidate model responses. 
By directly comparing the responses of different models, \citet{judging,role} confirm that CBE can more accurately assess model performance.
However, the $O(NM^2)$ evaluation overhead limits the practicality of CBE when there are $M$ models to evaluate on $N$ samples \citep{allpair}. 
To achieve efficient CBE, various methods have been explored \cite{arena,speeding,alpacaeval}. 
As shown in Figure~\ref{fig:intro}, based on existing observational results, these methods iteratively allocate preference budget to the next (models, sample) tuple according to respective optimization objectives. 
Specific preference aggregation methods (e.g., ELO rating \citep{elo}) are then applied to predict the model capability scores based on these preference results. 
Nevertheless, as shown in Table~\ref{tb:intro}, the optimization objectives of these methods are often singular, failing to simultaneously achieve the accuracy, convergence, and scalability well. We will discuss this in detail in \S\ref{sec:2.1} and conduct experimental validation in \S\ref{sec:exp-1}.

% \begin{wraptable}{r}{0.6\textwidth}
%     \vspace{-22pt}
%     \centering
%     \small
%     \caption{Optimization Objectives of Different Methods.}
%     \begin{tabular}{l c c c}
%     \toprule
%        Methods & Accuracy & Convergence & Scalability \\ \toprule
%       \citet{allpair}  & +  & - & - \\
%       \citet{arena}   & - & + & -\\
%       \citet{alpacaeval} & - & - &++\\ 
%       \textsc{UniCBE} & ++ & ++  &++\\
%         \bottomrule
%     \end{tabular}
%     \vspace{-22pt}
%     \label{tb:intro}
% \end{wraptable}
% \begin{wraptable}{r}{0.6\textwidth}
\begin{table}[ht]
    \vspace{-8pt}
    \centering
    \small
    \caption{Optimization Objectives of widely applied CBE Methods. The number of '+' indicates the degree of optimization for the objective, which is discussed in \S\ref{sec:2.1} and measured in Table~\ref{tab:beta}.}
    \begin{tabular}{l c c c c}
    \toprule
       \multirow{2}{*}{Methods} & \citet{allpair} & \citet{arena} & \citet{alpacaeval} & Ours \\
&\textsc{Random}&\textsc{Arena}&\textsc{AlpacaEval}&\textsc{UniCBE} \\ \midrule
      Accuracy  & +  & - & - & ++ \\
      Convergence   & - & + & - &++\\
      Scalability & - & - &++&++\\ 
        \bottomrule
    \end{tabular}
    % \vspace{-22pt}
    \label{tb:intro}
\end{table}
% \end{wraptable}
% 为了设计能快速收敛并准确评测模型性能、当新的待评测模型加入时实现良好可扩展性的采样方法，我们从理论上分析总结了以下insights：提升评测结果准确性的关键在于均匀化采样元素组合，从而削弱潜在的观测偏差；提升评测结果稳定性的关键在于均匀化胜率矩阵不确定性的梯度下降过程，从而在全局层面削弱观测方差；而提升可扩展性的关键则在于对新加入的待评测模型倾斜足够的评估开销，从而加速削弱更新偏差、实现评测结果稳态化的进程。基于这些insights的指导，我们提出了\textsc{EfficientArena}方法。我们首先建立采样过程中的状态变量分别与观测偏差、观测方差、更新偏差间的映射关系；然后我们联立这些解耦的映射关系来获取兼顾准确性、稳定性与可扩展性的采样概率矩阵，并据此进行采样。

To develop a method that can accurately assess model performance, quickly converge evaluation results, and ensure good scalability when new models are introduced, we theoretically analyze and summarize the following guidelines:
\begin{itemize}[leftmargin=20pt]
\setlength{\itemsep}{0pt}
\setlength{\parsep}{0pt}
\setlength{\parskip}{0pt}
% \item The key to improving the \textbf{accuracy} of evaluation results depends on the uniform sampling of item (model and sample) tuples, which helps mitigate the sampling bias.
% \item The key to accelerating the \textbf{convergence} process lies in balancing the gradient descent process of the win rate uncertainty matrix, thereby globally reducing observation variance.
% \item The key to strengthening the \textbf{scalability} is to allocate sufficient evaluation budgets to newly introduced models under assessment, thereby accelerating the reduction of updating bias.
\item Improving the \textit{\textbf{accuracy}} of evaluation results relies on completely \textbf{uniform} sampling of tuple combinations, so as to mitigate sampling bias.
\item Accelerating the \textit{\textbf{convergence}} process involves ensuring the \textbf{uniformity} of the win rate uncertainty matrix during its descending process to reduce observation variance.
\item Enhancing \textit{\textbf{scalability}} requires sufficient budgets being allocated to new added models to ensure the \textbf{uniform} allocation among models, which helps reduce updating uncertainty.
\end{itemize}
Based on these insights, we propose \textsc{UniCBE}, a unified uniformity-driven framework that can achieve CBE with better accuracy, convergence and scalability. 
In each iteration of the evaluation process, we first establish sampling probability matrices under different optimization objectives respectively based on real-time preference results.
Afterwards, we integrate these matrices to obtain a global sampling probability matrix. 
Furthermore, we explore various tuple sampling strategies and preference aggregation methods to achieve optimal evaluation results.

% 为了全面地验证所提出方法的有效性和泛化性，我们设置了多组实验，其中包括不同的judger（GPT-3.5-turbo, GPT-4o, human），不同的数据集（AlpacaEval、MTBench），不同的待评测模型组合（随机地从20个主流LLM中进行选择），不同的场景（包括静态的和不断有新的待评测模型加入的场景），不同的评测指标（模型能力预测值与真实值之间的一致性系数、模型间胜率预测值与真实值之间的MAE距离）。Main results显示，相比于随机采样的基线，在达到相同的评测精度时（超过0.995的斯皮尔曼系数），\textsc{EfficientArena}能够节省超过20%的评测开销，展现了显著高于其他各基线的良好稳定性和准确性。此外在有新的待评测模型加入的场景下，\textsc{EfficientArena}能够显著更快地完成对其能力的准确评测，相较于随机采样能够节省超过50%的评测开销，展现了优异的可扩展性。

To comprehensively validate the effectiveness and generalizability of \textsc{UniCBE}, we conduct multiple experiments involving various types of judges (LLMs and humans), different benchmarks, varied model sets to be evaluated, diverse scenarios (static and dynamic), and multiple evaluation metrics. 
The main results indicate that, compared to the random sampling baseline, \textsc{UniCBE} saves over 17\% of evaluation budgets when achieving the same assessment accuracy (with a Pearson coefficient exceeding 0.995 with the ground truth), demonstrating significantly better convergence and accuracy than other baselines. Furthermore, in scenarios where new models are continuously introduced, \textsc{UniCBE} can even save over 50\% of evaluation costs compared to random sampling, showcasing excellent scalability.

%\cite{all}考虑直接利用所有模型在所有samples上的pair-wise比较结果来比较模型能力，这也是理论上能得到的最准确评测结果。 
% \cite{arena}以模型间胜率关系观测值的方差下降梯度为模型对的采样概率，试图在每次调用judger时最大程度地减少评估结果的不确定性。
% \cite{alpacaeval}通过选择固定的reference model并调用judger比较待评测模型与reference model来将\cite{all}所需的O(N^2)的评估开销缩减到O(N)。

\section{Related Work}
% 我们在本节讨论现有的CBE方法和Preference Aggregation方法
% 由于能够区分质量差异微小的responses，基于对比的preference信号长期被用来进行模型训练和测试。
% Due to the ability of distinguishing between subtle difference in response quality, 
Comparative preference signals have long been used for model training \citep{insgpt,llama2} and evaluation \citep{arena,batcheval}.
Centered around comparing-based evaluation, we will discuss existing budget allocation strategies and preference aggregation methods below.
\paragraph{Budget Allocation}
\label{sec:2.1}
% 最naive的Allocation方法是每次随机选择(models,sample)组合，并由judge提供preference，直到达到预设的Preference Budget. according to our 推导 in section3，该方法在期望上保证了对于不同样本组合的均匀采样从而保证了评测结果准确性。
% Arena 考虑在每一步依据胜率方差梯度对模型对采样，旨在通过greedy的方式减小观测胜率矩阵的不确定性，加速evaluation的收敛过程。
% AlpacaEval 通过比较待评测模型与固定的reference model之间的胜负关系来衡量模型性能。不同的待评测模型被要求在相同的set上进行评测，这意味着在动态场景下当新模型被加入时会优先分配preference给新模型使对其的能力估计快速稳定，从而实现良好的可扩展性。
% 尽管这些方法都能在期望的aspect保障良好的表现，但却不能同时兼顾准确性、收敛性和可扩展性，这对探索能平衡多方面性质的Preference Budget Allocation方法提出了迫切需求。
Many efforts have been made to explore preference budget allocation approaches.
The most naive allocation method is to randomly select (models, sample) tuple for judging each time until the preset preference budget is reached \citep{allpair}. 
This method ensures a relatively uniform sampling across different tuple combinations in expectation, thereby guaranteeing the accuracy of evaluation results according to our derivation in \S\ref{sec:3.2}.
Arena \citep{arena} aims to sample model pairs proportionally to the variance gradient of win rate at each step, seeking to accelerating the convergence of evaluation by reducing the uncertainty of the observed win rate matrix in a greedy manner. 
AlpacaEval \citep{alpacaeval} measures model performance by comparing the models under evaluation with a fixed reference model. Different models under evaluation are expected to be assessed on the same set. Thus, when new models are introduced, preference budget is prioritized for them to stabilize the estimation of their capabilities, thereby achieving good scalability.
Despite these methods performing well in their intended objectives, they cannot achieve a balance among accuracy, convergence, and scalability. This makes it imperative to explore better preference budget allocation strategy that can effectively reconcile all these attributes.
\paragraph{Preference Aggregation}
\label{sec:2-pa}
% 由于同一组模型可能在不同的samples上呈现不同的胜负关系，因此我们需要find the optimal preference ranking 在收集一定量的non-transitive comparison preference results among models under evaluation后。
Due to the possibility that the same group of models may exhibit different ranking relationships across different samples, it is essential to estimate the global model capability scores to better fit these non-transitive preference results.
% \citet{alpacaeval,judging} 直接用每个模型的平均胜率作为该模型能力的度量。\citet{elo2,elo1} 采用经典的Elo rating system（见附录 for 细致的计算过程）将评测过程视为序列化的模型battle来获得模型score。\citet{BTapp,arena}采用Bradley-Terry model \citep{BT} 估计模型score来最大似然模型间的比较结果。我们将在实验中系统比较这些Preference Aggregation方法在efficient CBE中的效果。
\citet{alpacaeval,judging} directly use the average pair-wise win rate of each model as a measure of its capability. 
\citet{elo2,elo1} apply the classical Elo rating system \citep{elo} (see the Appendix~\ref{app:elo} for detailed introduction) by treating the evaluation process as a sequence of model battles in order to derive model scores. 
\citet{BTapp,arena} employ the Bradley-Terry model \citep{BT} (see Appendix~\ref{app:bt} for detailed introduction) to estimate model scores by maximizing the likelihood of the comparison results between models. We will systematically compare the effectiveness of these preference aggregation methods in \S\ref{sec:exp-2}.

\section{Preliminary}
\label{sec:3}
% 在本节，我们将分别定义实现高效CBE的三个要素：accuracy，convergence, scalability，并分析影响它们的要素。
In this section, we start by symbolically introducing the working process of CBE. Afterwards, we introduce the key objectives for achieving efficient CBE: accuracy, convergence, and scalability, and analyze the factors that influence them.
We mainly discuss the pair-wise evaluation scenario (where the judge provides preference between two models per time) for its wide applications \citep{paireval,allpair}. Actually, list-wise preferences can be easily converted into pair-wise ones, as demonstrated in \S\ref{sec:5.4-2}, so the discussions below are general for CBE.
% 考虑到pair-wise comparison（Judge每次给出对两个models的preference）是CBE的最典型场景，我们在本节围绕pair-wsie evaluation展开讨论。事实上，list-wise preference results可以轻松地转化成pair-wise，正如第三届所示的那样。
\subsection{Process of CBE}
%一个CBE method 可以被划分为三部分：preference allocation strategy, sampling strategy以及model score estimation strategy。
Generally, a CBE method $f$ can be divided into three parts: budget allocation strategy $f^{ba}$, tuple sampling strategy $f^{ts}$, and preference aggregation strategy $f^{pa}$.
Given benchmark $\mathcal{D}:s_{1:N}$ and models under evaluation $\mathcal{M}:m_{1:M}$, we iterate the following steps: 
\textit{step 1.} applying $f^{ba}$ to attain sampling matrix $P^l$ at iteration $l$, where $P_{i,j,k}^l$ denotes the probability to select tuple $(m_i,m_j,s_k)$ for judging; 
\textit{step 2.} applying $f^{ts}$ to sample certain tuple $(m^{l1},m^{l2},s^l)$ based on $P^l$; 
\textit{step 3.} attaining preference result $r^l$ from the judge, where $r^l \in [0,1]$ denotes the degree $m^{l1}$ wins over $m^{l2}$ (0.5 means tie). 
We stop this iterative process when the preset preference budget $T$ is achieved and then apply $f^{pa}$ on preference results $\{(m^{l1},m^{l2},s^l,r^l)\}_{l=1}^T$ to attain estimated model scores $u_{1:M}$.

\subsection{Accuracy}
\label{sec:3.2}
Theoretically, if we have a budget of $\hat{T}=\frac{NM(M-1)}{2}$, we can explore all tuples to obtain the ground truth estimation for the model scores $\hat{u}_{1:M}$. 
However, typically $T$ is much smaller than $\hat{T}$ in reality considering the preciousness of preference signals. 
% 然而，在现实中考虑到preference的珍贵性T通常远小于Y. 
Previous studies \citep{samplebias1,samplebias2} have discussed the risks of introducing sampling bias in incomplete sampling scenarios, which we believe could similarly lead to potential risks in CBE. % due to inappropriate $f^{ba}$. 
% 考虑到每次采样的内容是(ml-1, ml-2, sl)，我们认为sample bias存在两方面.
Considering that the content of each sample is $(m^{l1},m^{l2},s^l)$, we think the sample bias exists across both samples and models. 
% 首先，由于不同的模型可能擅长回答不同类型的问题，因此模型的能力值会随采样的sample而发生变化：

\begin{figure}[ht]
    \centering
    \subfigure[Sampling bias with different preference aggregation strategies across samples and models.]{
    \label{fig:bias1}
        \includegraphics[width=0.47\textwidth]{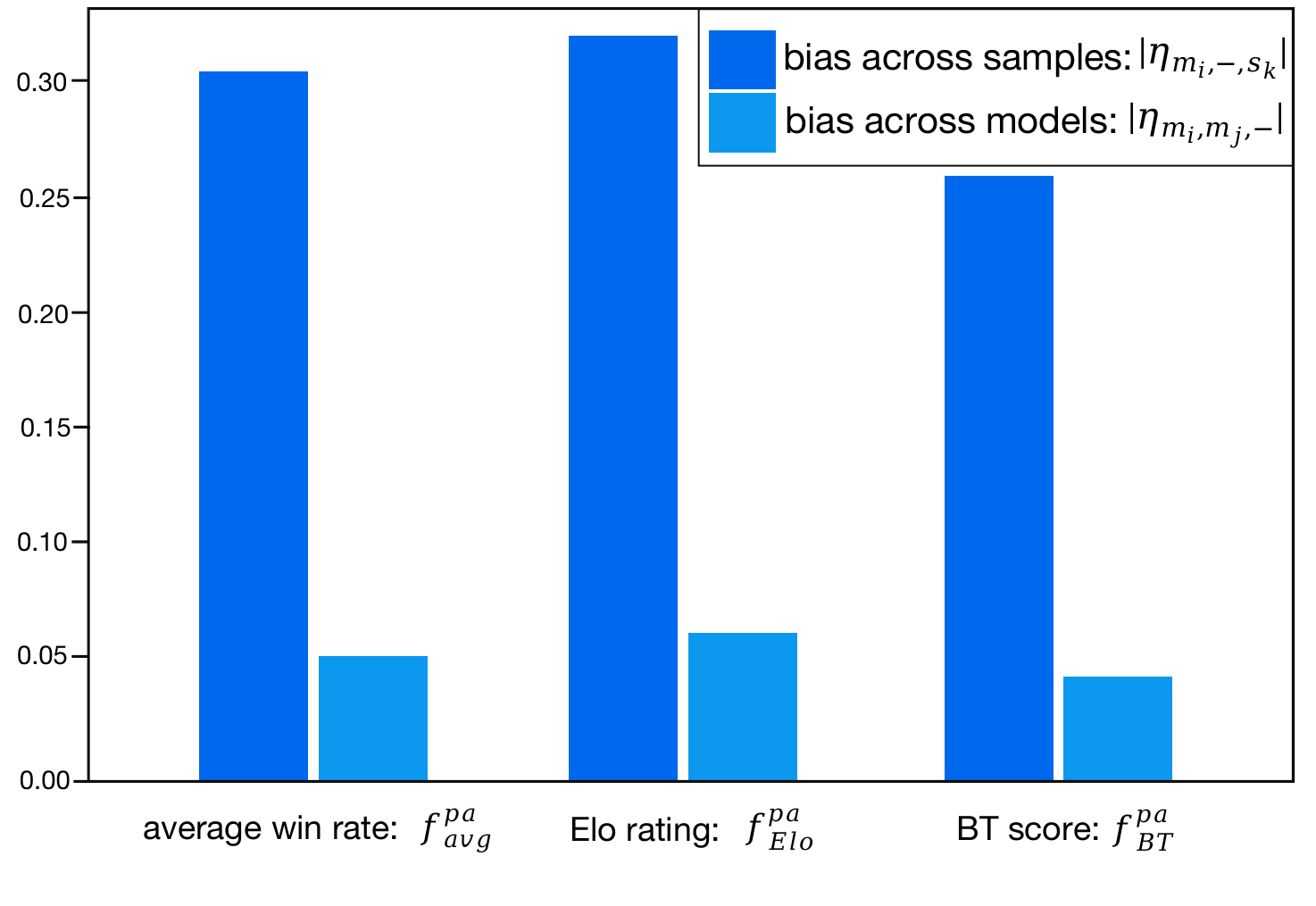}
    }
    \hfill
    \subfigure[Interval distribution of bias across samples with $f^{pa}_{BT}$ as preference aggregation strategy.]{
    \label{fig:bias2}
        \includegraphics[width=0.48\textwidth]{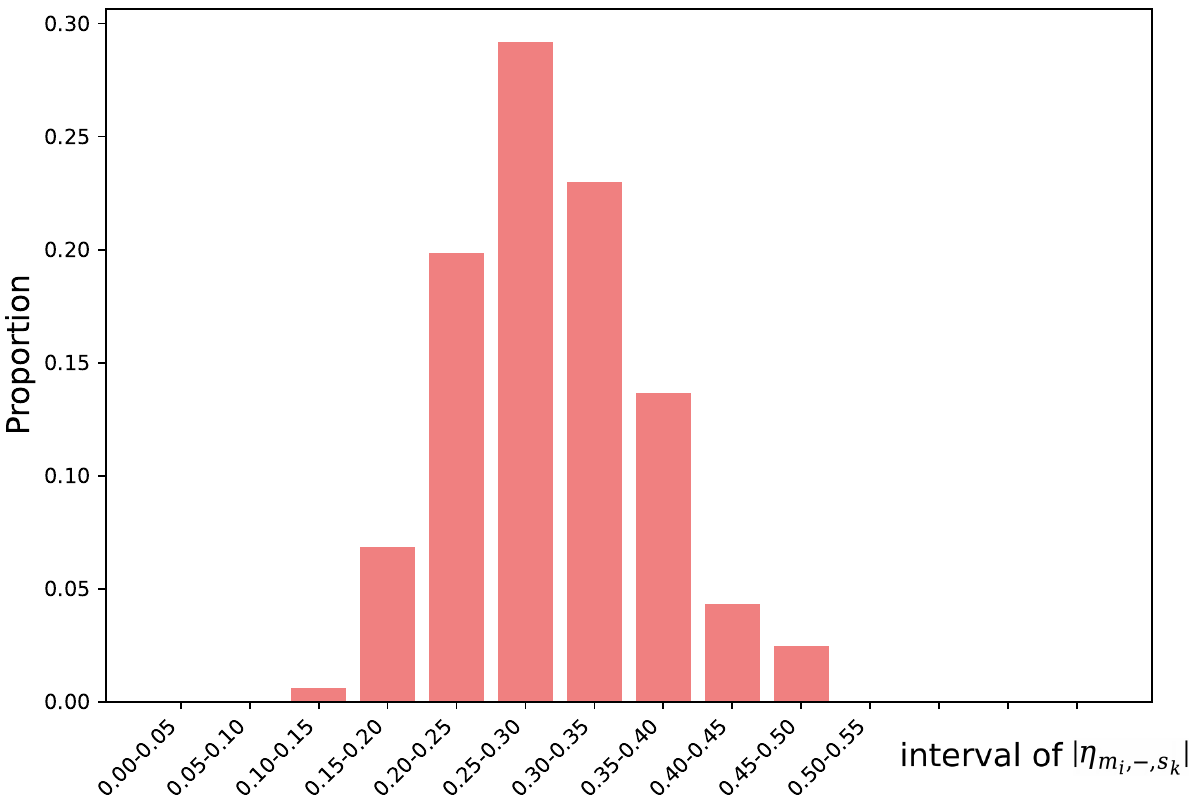}
    } \\
    \vspace{-0.2cm}
    \subfigure[Bias across models with $f^{pa}_{BT}$ as preference aggregation strategy.]{
    \label{fig:bias3}
        \includegraphics[width=0.95\textwidth]{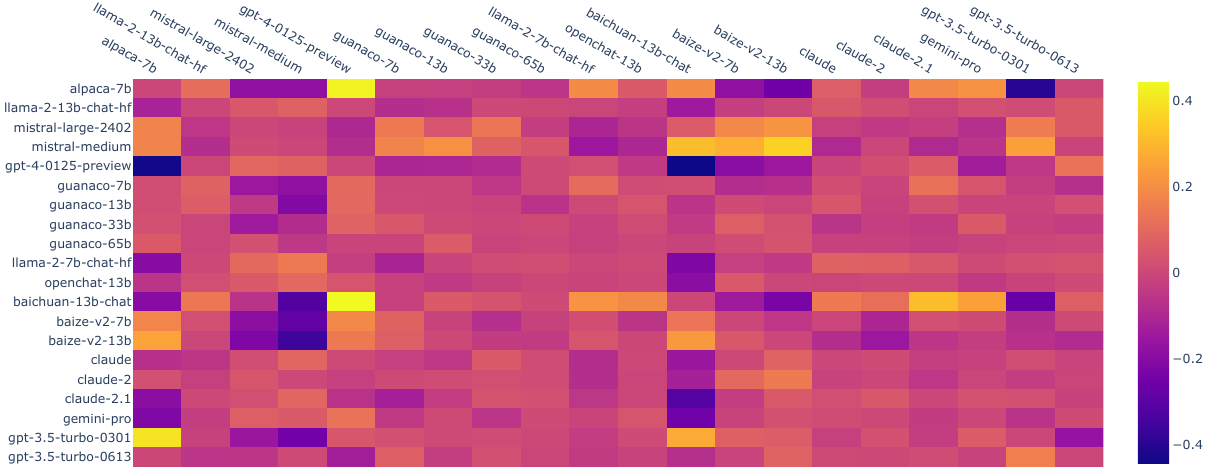}
    }
    \vspace{-0.2cm}
    \caption{Analyses of potential sampling bias risks in CBE.}
    \vspace{-0.4cm}
    \label{fig:samplebias}
\end{figure}

\paragraph{Bias across Samples.} Since different models may excel at answering different types of queries, the model scores can vary depending on the sampled data:
\begin{equation}
    \ \ \ \ \ \ \ \ \ \ \ \ \ \ \ \ \ \ \ u_t=f^{pa}(\{(m_i,m_j,s_k,r_{i,j,k})\}_{i\in 1:M,j\in i+1:M})_t = \hat{u}_t + \eta_{m_t,\text{-},s_k} \ \ \ \ \ \ \ \ \ \ \ \ \text{for} \ \ \ \forall \  t,k
    \label{eq:1}
\end{equation}
% 为了验证this，我们在alpacaeval benchmark上以GPT-4o为judge，随机选择了20个模型进行了如下实验：
% 我们首先在每个sample上遍历所有的model pair得到对应的N组preference results。然后应用f计算相应的u，并计算u与s之间差值的绝对值，也即k，并取均值。
% s中提到的三种Preference Aggregation方法计算对应的
where $\eta_{m_t,\text{-},s_k}$ represents the bias between the observed model score $u_t$ of $m_t$ and the ground truth $\hat{u}_t$ when sorely assessing on sample $s_k$.
To verify this, we conduct experiments on the AlpacaEval benchmark \citep{alpacaeval} using GPT-4o \citep{4o} as the judge across randomly selected 20 LLMs (listed in Figure~\ref{fig:bias3}). We first traversed all model pairs for samples $s_{1:N}$ to obtain corresponding $N$ sets of preference results and then calculate the respective $|\eta_{m_i,\text{-},s_k}|$ for $\ i\in 1:M$ and $\ k\in 1:N$ according to~\eqref{eq:1} (model scores are normalized to an average of 1). We calculate the average value of $|\eta_{m_i,\text{-},s_k}|$ across models and samples using different preference aggregation strategies $f^{pa}$ discussed in \S\ref{sec:2-pa}. As shown in Figure~\ref{fig:bias1}, with all kinds of $f^{pa}$, the average difference between the model scores estimated on single sample and the ground truth values exceeds 0.25, indicating a significant bias across samples. We further analyze the proportion of samples with different biases using $f^{pa}_{BT}$ in Figure~\ref{fig:bias2} and find that they overall follow a Gaussian distribution, showing the wide existence of sample bias in CBE.
\paragraph{Bias across Models.} Just as humans may perform differently when facing different opponents, models may also have varying scores when competing against different models:
\begin{equation}
    \ \ \ \ \ \ \ \ \ \ \ \ \ \ \ \ \ \ \ u_i=f^{pa}(\{(m_i,m_j,s_k,r_{i,j,k})\}_{k\in 1:N})_i = \hat{u}_i + \eta_{m_i,m_j,\text{-}} \ \ \ \ \ \ \ \ \ \ \ \ \text{for} \ \ \ \forall \  i,j
    \label{eq:2}
\end{equation}
We validate this from two perspectives:  
(1) We calculate the average $|\eta_{m_i,m_j,\text{-}}|$ according to~\eqref{eq:2} like the process above and show the results in Figure~\ref{fig:bias1}. Overall, although the bias across models is significantly lower than the bias across samples, it still exists at a scale around 0.05. We further visualize the pair-wise model score bias in Figure~\ref{fig:bias3} to validate its wide existence. 
(2) We obtain over 1.7 million pairwise preference results across 129 LLMs collected by Chatbot Arena \footnote{\url{https://storage.googleapis.com/arena_external_data/public/clean_battle_20240814_public.json}}. After excluding pairs with fewer than 50 comparisons, we calculate the pairwise win rates and find non-transitivity in 81 model triplets (win rate: $A > B$, $B > C$, $C > A$), which also verifies the existence of bias across models.

% 在

\paragraph{Uniform Allocation Brings the Least Bias.} 
Based on the discussions above, we analyze the budget allocation strategy that can introduce the least bias. 
Considering the presence of sampling bias, the estimation error of $u_i$ with $T$ evaluation budget can be expressed as follows:
%考虑到采样偏差的存在，在T次采样时，我们有以下equation：
\begin{equation}
    u_i-\hat{u}_i = \sum_{l=1}^T \1_{m^{l1}=m_i} \times \eta_{m^{l1},m^{l2},s^l}
    \label{eq:3}
\end{equation}

Considering that $u=\hat{u}$ when all the tuples are traversed, we have the following equation:
\begin{equation}
    \ \ \ \ \ \ \ \ \ \ 0 = u_i-\hat{u}_i = \sum_{j=1}^M\sum_{k=1}^N \eta_{i,j,k} \ \ \ \ \ \ \ \ \ \ \ \ \text{for} \ \ \ \forall \  i
    \label{eq:4}
\end{equation}
%此时获得最小的estimation error of $u_i$ 的目标转化为了从U个和为0的数中采样V个数，使这V个数的和的绝对值最小。
The goal of obtaining the minimum estimation error for $u_i$ is transformed into sampling \( T \) numbers (~\eqref{eq:3}) from \( MN \) numbers that sum to zero (~\eqref{eq:4}), such that the absolute value of the sum of these \( T \) numbers is minimized. We have provided a detailed proof in Appendix~\ref{app:proof} that the best strategy is completely uniform sampling. \textit{\textbf{This denotes that the score estimation error can be minimized when the preference budgets are uniformly distributed across models and samples to bring the least sampling bias.}}

\subsection{Convergence}
\label{sec:3.3}
% 在评测进行的过程中，随着新的preference results被观测到，模型间的胜负关系和对模型score的预估值都在发生变化。
During the evaluation process, as new preference results are continuously observed, the estimated values of the models win rate matrix and model scores also change constantly. 
To accelerate the convergence process, we analyze the uncertainty of the win rate matrix as follows.
Defining that:
\begin{equation}
    X^l_{i,j} = \frac{1}{P^l_{i,j}}r^l\1_{m^{l1}=m_i \And m^{l2}=m_j} + \frac{1}{P^l_{j,i}}(1-r^l)\1_{m^{l1}=m_j \And m^{l2}=m_i}
    \label{eq:6}
\end{equation}
The unbiased estimated win rate matrix $\Phi$ at iteration $L$ can be calculated as follows:
\begin{equation}
    \Phi^L = \frac{1}{L}\sum_{l=1}^L X^l
    \label{eq:7}
\end{equation}
We further estimate the variance matrix $\Theta$ as:
\begin{equation}
    \Theta^L = \frac{1}{L}\sum_{l=1}^L(X^l-\Phi^L)\circ (X^l-\Phi^L)
    \label{eq:8}
\end{equation}
Denoting if the model pair $(m_i,m_j)$ has been compared on sample $s_k$ after $l$ iterations as $C^l_{i,j,k}$, the uncertainty (standard deviation) of each element in the win rate matrix is as follows:
\begin{equation}
    \epsilon^l_{i,j} = \sqrt{\frac{\Theta^l_{i,j}}{\sum_{k=1}^{N} C^l_{i,j,k}}}
    \label{eq:9}
\end{equation}
Allocating the next preference budget on $(m_i,m_j)$ can reduce the uncertainty of their win rate by:
\begin{equation}
    \sqrt{\frac{\Theta^l_{i,j}}{\sum_{k=1}^{N} C^l_{i,j,k}}} - \sqrt{\frac{\Theta^l_{i,j}}{\sum_{k=1}^{N} C^l_{i,j,k}+1}}
    \label{eq:10}
\end{equation}
%在此基础上，考虑到我们的核心目标是对于所有的模型都进行准确的能力评估，因此我们需要globally平衡胜率矩阵不确定度的descent process以避免。
Considering that our core objective is to conduct accurate capability assessments for all models and estimate their ranking relationship, \textit{\textbf{we should globally ensure the uniformity of the win rate uncertainty matrix during its descending process to achieve smooth evaluation convergence}}.

\subsection{Scalability}
\label{sec:3.4}
Due to the continuous emergence of new LLMs, the demand for scalability in evaluation method is becoming increasingly prominent \citep{scalable}. 
Considering that we have evaluated $m_{1:M}$ with $T$ budgets, when model $m_{M+1}$ is introduced for assessment, a well-scalable CBE method should be able to quickly calibrate the capability estimates of $m_{1:M+1}$ with minimal additional preference budget. 
In this scenario, at the beginning stage when \( m_{M+1} \) is introduced, $\mathrm{avg}(C_{M+1,\text{-},\text{-}})$ is much smaller than $\mathrm{avg}(C_{\neq M+1,\text{-},\text{-}})$. 
According to~\eqref{eq:9}, the uncertainty at this point mainly arises from $\epsilon_{M+1}$, which is also intuitively easy to understand.
\textit{\textbf{Therefore, the key to improving scalability lies in allocating sufficient evaluation budgets to the newly added models to ensure
the uniform allocation among models, reducing the updating uncertainty.}}
%依据公式7可知此时评测的不确定性主要来自于对$m_{1:M+1}$，这也是直观上容易理解的。
%因此，提升可扩展性的关键在于sufficient evaluation budgets being allocated to new added models 从而快速减少更新带来的不确定性
% 

% 定义accuracy，convergence, scalability 
% 明确他们分别与sampling bias, observation variance, updating bias的关系
% 推导降低这三者的方法分别在于XXX
\section{\textsc{UniCBE}}
% \section{CBE with Better Scalability, Accuracy and Convergence}
% 以上的讨论揭示了为CBE带来Scalability, Accuracy and Convergence的guidelines，基于此我们在本节提出可以兼顾这些特性的CBE方法，UniCBE。
The discussions above reveal guidelines for strengthening scalability, accuracy, and convergence in CBE. Based on this, we propose \textsc{UniCBE}, a unified uniformity-driven framework that can simultaneously enhance these objectives well.

\subsection{Budget Allocation}
% 为了实现尽可能均匀的采样，从而最少地引入采样bias，我们如下构建P：
To ensure the uniformity of tuple combination sampling for minimizing the introduction of sampling bias according to \S\ref{sec:3.2}, we construct $P^{acc\text{-}l}$ 
 at iteration $l$ as follows:
\begin{equation}
    P^{acc\text{-}l}_{i,j,k} = \alpha^{-\sum_{k=1}^{N} C^l_{i,j,k}}\times \alpha^{-\sum_{i=1}^{M} C^l_{i,j,k}}\times \alpha^{-\sum_{j=1}^{M} C^l_{i,j,k}}
    \label{eq:11}
\end{equation}
where $\sum_{k=1}^{N} C^l_{i,j,k}$ denotes the times model pair $(m_i,m_j)$ has been compared, $\sum_{i=1}^{M} C^l_{i,j,k}$ and $\sum_{j=1}^{M} C^l_{i,j,k}$ denote the times model $m_i$ and $m_j$ has been tested on $s_k$ respectively. 
If certain model-model combination or model-sample combination have been sampled multiple times,~\eqref{eq:11} will reduce the probability of such combinations being selected again, thereby achieving sufficient uniformity to minimize the introduction of bias between models and samples, respectively.

To accelerate the convergence of evaluation results, we construct $P^{con\text{-}l}$ according to \S\ref{sec:3.3} as follows:
\begin{equation}
    P^{con\text{-}l}_{i,j,k} = \epsilon^l_{i,j}
    \label{eq:12}
\end{equation}
Sampling specific model pair helps reduce the uncertainty of their win rate estimation according to~\eqref{eq:10}. By sampling proportionally to the win rate uncertainty matrix, we can uniformly decrease the uncertainty for each model pair, thereby facilitating convergence.
% Based on~\eqref{eq:10} that sampling specific model pair help reduce the uncertainty of their win rate, sampling according to the win rate uncertainty matrix can ensure that the uncertainty of each pair of models is reduced simultaneously, thereby facilitating convergence.
% 
%如果模型pair或模型样本pair已经被多次采样了，（15）将减小这样的pair再次被选中的概率，从而实现足够的均匀性来减少模型间bias和样本间bias的引入。

We construct $P^{sca\text{-}l}$ to allocate more preference budget to the newly introduced model so as to improving the scalability according to \S\ref{sec:3.4} as follows:
\begin{equation}
    P^{sca\text{-}l}_{i,j,k} = \alpha^{-\sum_{k=1}^{N}\sum_{i=1}^{M} C^l_{i,j,k}}\times \alpha^{-\sum_{k=1}^{N}\sum_{j=1}^{M} C^l_{i,j,k}}
    \label{eq:13}
\end{equation}
%最后，我们集成上述三个矩阵来获得P，从而实现依据P采样时能够同时兼顾accuracy、convergence和scalability：
Finally, we integrate the matrices mentioned above to obtain $P^l$, ensuring that sampling according to $P^l$ can simultaneously balance the accuracy, convergence, and scalability of evaluation results:
\begin{equation}
\begin{aligned}
    P^{l} = \frac{P^{acc\text{-}l} \circ P^{con\text{-}l} \circ P^{sca\text{-}l}}{\sum (P^{acc\text{-}l} \circ P^{con\text{-}l} \circ P^{sca\text{-}l})}
    \label{eq:14}
\end{aligned}
\end{equation}

\subsection{Tuple Sampling}
% 在获得$P^l$后，我们要依据它采样tuple for judging. 我们考虑两种 tuple sampling策略：probabilistic sampling and greedy sampling.
After obtaining \( P^l \), we need to sample tuples for judging based on it. Two tuple sampling strategies are considered: 
\begin{itemize}[leftmargin=20pt]
\setlength{\itemsep}{0pt}
\setlength{\parsep}{0pt}
\setlength{\parskip}{0pt}
\item \textbf{probabilistic sampling $f^{ts}_{p}$} means sampling tuple directly according to \( P^l \).
\item \textbf{greedy sampling $f^{ts}_{g}$} means selecting the tuple with the maximum probability in \( P^l \).
\end{itemize}
The default tuple sampling strategy of \textsc{UniCBE} is $f^{ts}_{g}$, which can avoid the suboptimal achievement of objectives due to uncertainty in the sampling process.

\subsection{Preference Aggregation}
As discussed in \S\ref{sec:2-pa}, mainstream preference aggregation strategies include averaging win rate $f^{pa}_{avg}$, Elo rating system $f^{pa}_{Elo}$ and Bradley-Terry model $f^{pa}_{BT}$. In our preliminary experiment (Figure~\ref{fig:bias3}) we observe that $f^{pa}_{BT}$ can better alleviate sampling bias, for which we choose it as our default setting.

\section{Experiments}
Centered around \textsc{UniCBE}, we will empirically compare its performance with baselines and validate its scalability in \S\ref{sec:exp-1}, explore the optimal variants in \S\ref{sec:exp-2}, and demonstrate its generalizability under different settings in \S\ref{sec:exp-3}.

\subsection{Experimental Settings}
\paragraph{Benchmarks.} We choose AlpacaEval \citep{alpacaeval} and MT-Bench \citep{judging} benchmarks for our experiments. For AlpacaEval, we use its default version which includes 805 high-quality human annotated instructions and corresponding responses from multiple LLMs. We randomly choose 20 LLMs (listed in Figure~\ref{fig:bias3}) for experiments, with GPT-4o and GPT-3.5-turbo as judges (see Appendix~\ref{app:prompt} for the prompt). For MT-Bench, we use the released responses from the all 6 LLMs and corresponding human preferences for experiments.
\paragraph{Baselines.} We choose widely applied \footnote{\url{https://tatsu-lab.github.io/alpaca_eval/}, \url{https://lmarena.ai/}} methods \textsc{Random}, \textsc{Arena} and \textsc{AlpacalEval} as baselines, which have been discussed in \S\ref{sec:2.1} and listed in Table~\ref{tb:intro}.
\paragraph{Metrics.} 
% 为了衡量CBE方法的有效性，我们分别检验预估的胜率和模型score的准确性。我们计算预估的胜率与对应Ground truth（$T=\hat{T}$时的预估值）之间的平均误差一范数
To assess the effectiveness of the CBE methods, we evaluate the accuracy of both the estimated model pair-wise win rates and the model scores. We calculate the average absolute error between the estimated win rates and corresponding ground truth (the estimates when $T = \hat{T}$).
% \begin{equation}
%     \Delta = \rm{mean}(|\Phi^{T}-\Phi^{\hat{T}}|)
%     \label{eq:15}
% \end{equation}
%我们计算预估的模型分数和对应Ground truth之间的斯皮尔曼系数和皮尔逊系数来分别检验预估的模型能力的秩序关系和线性关系的准确性
We calculate the Spearman correlation coefficient $r_s$ between the predicted model scores and the corresponding ground truth to evaluate the accuracy of the model's rank-order relationship, and the Pearson correlation coefficient $r_p$ to assess the accuracy of the linear relationship.
\paragraph{Details.} 
To ensure the reliability of the experimental results, for each setting, we randomly select $M$ (default to 15 for AlpacaEval and 5 for MT-Bench) models and $N$ (default to 805 for AlpacaEval and 700 for MT-Bench) samples, and report the average results across 10,000 random seeds. We don't observe obvious performance difference in preliminary experiments when varying $\alpha$ within the range of $[1.5,3]$ (we conduct a detailed discussion about this in Appendix~\ref{app:dis-aff}), thus we set the default value of $\alpha$ as 2 in our experiments. 
\begin{figure}[h]
    \centering
    \subfigure{
    \label{fig:4o1}
        \includegraphics[width=0.31\textwidth]{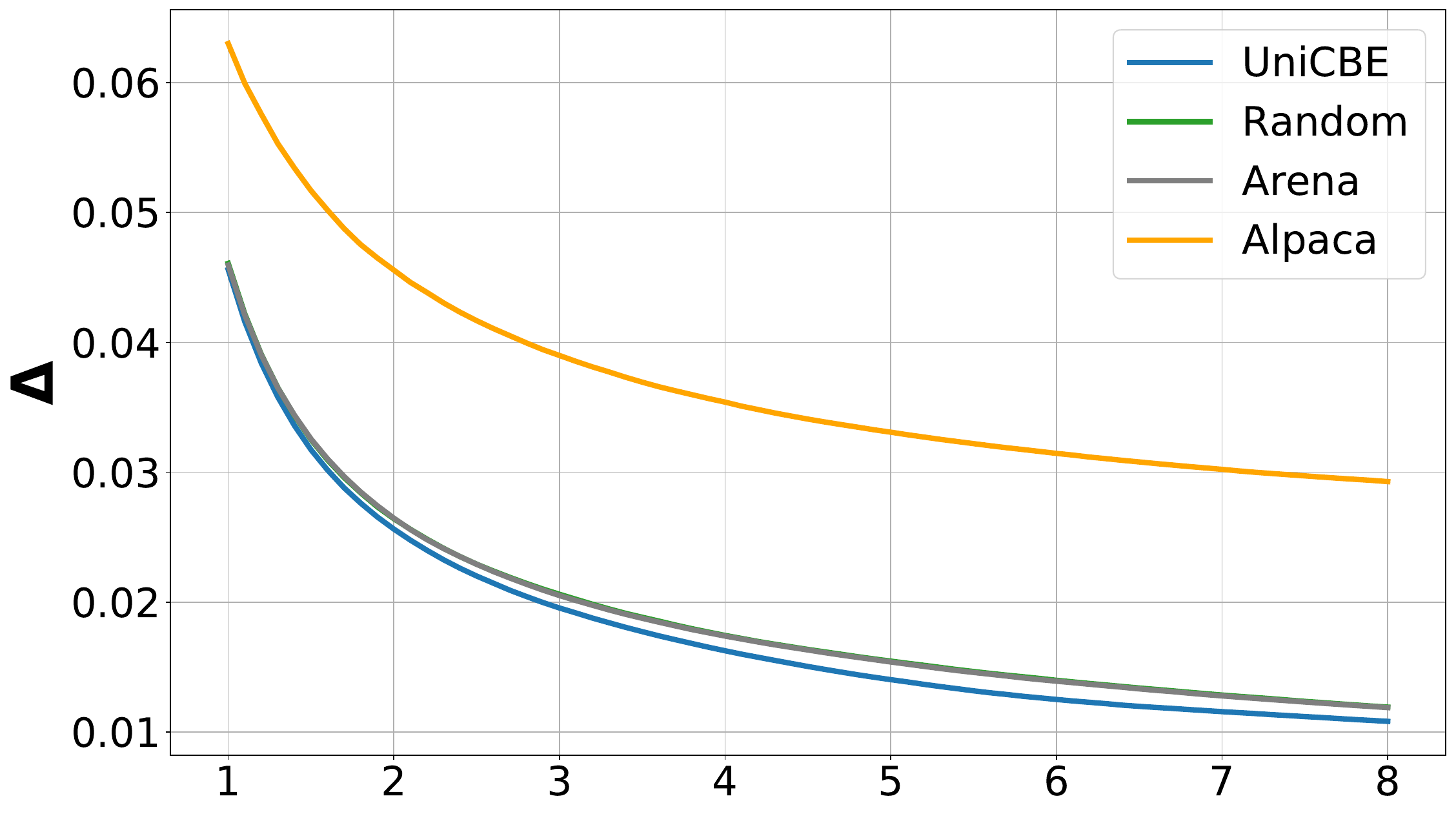}
    }
    \hfill
    \subfigure{
    \label{fig:4o2}
        \includegraphics[width=0.31\textwidth]{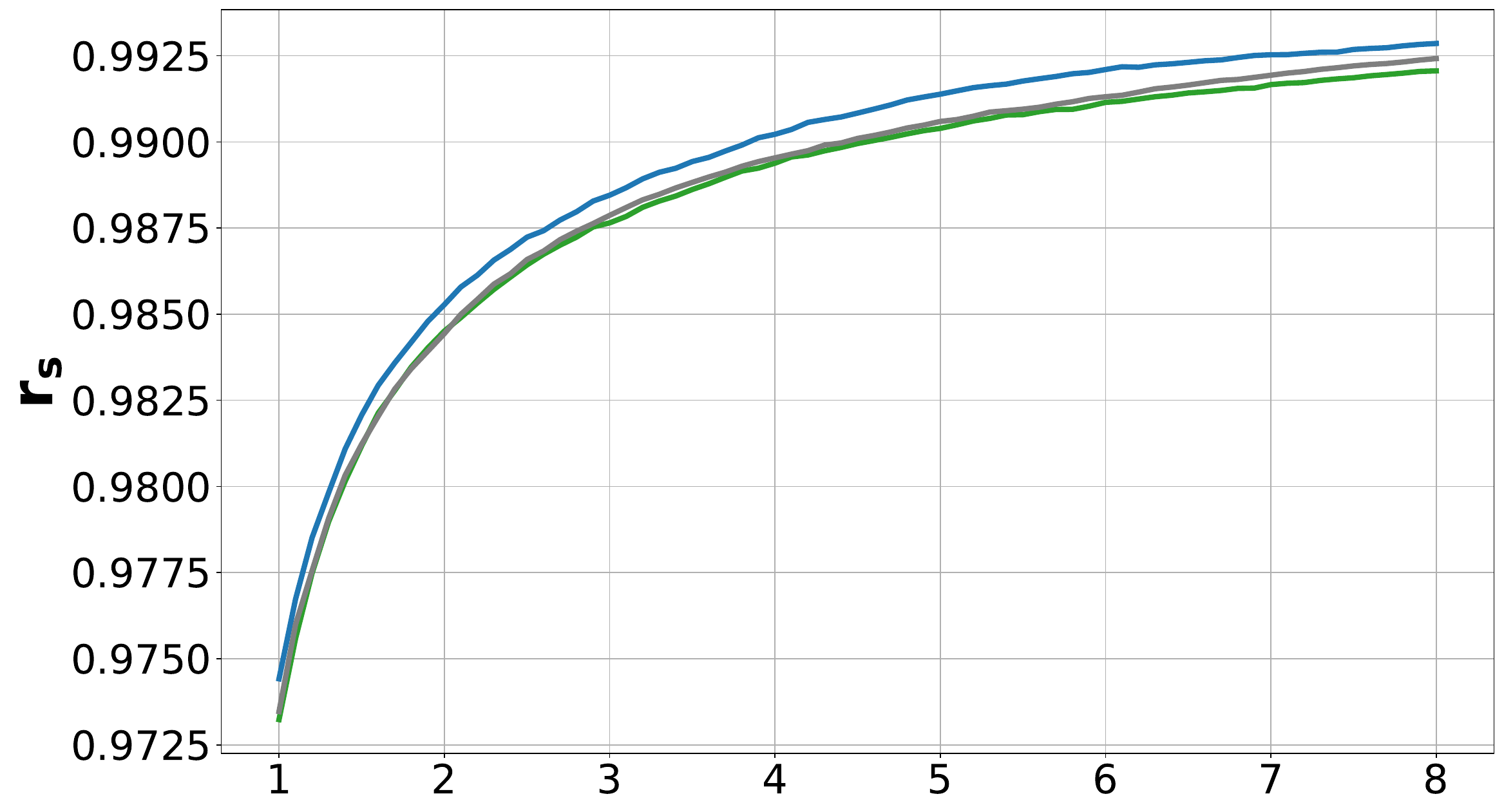}
    }
    \hfill
    \subfigure{
    \label{fig:4o3}
        \includegraphics[width=0.31\textwidth]{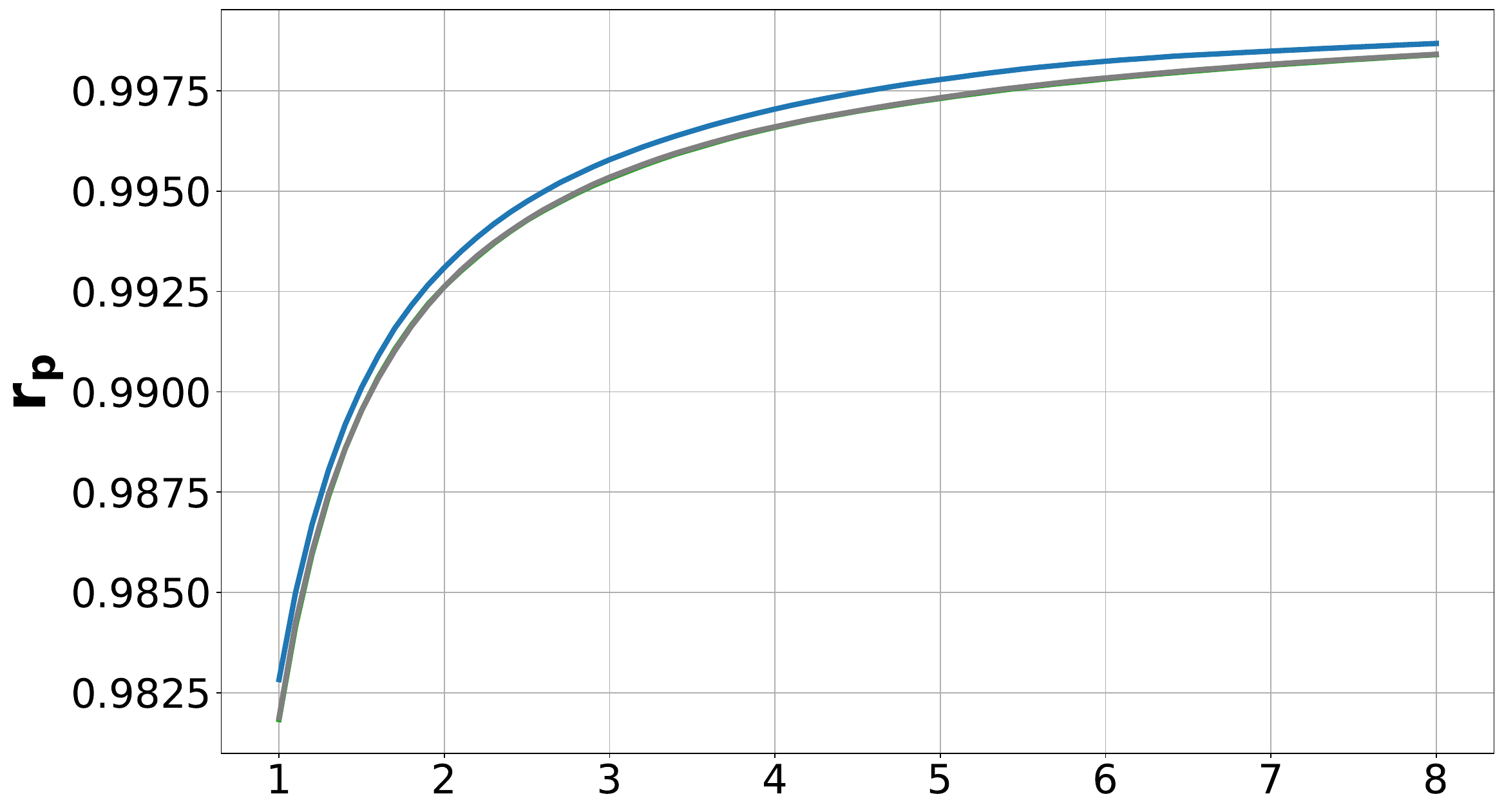}
    } \\
    \subfigure{
    \label{fig:4o4}
        \includegraphics[width=0.31\textwidth]{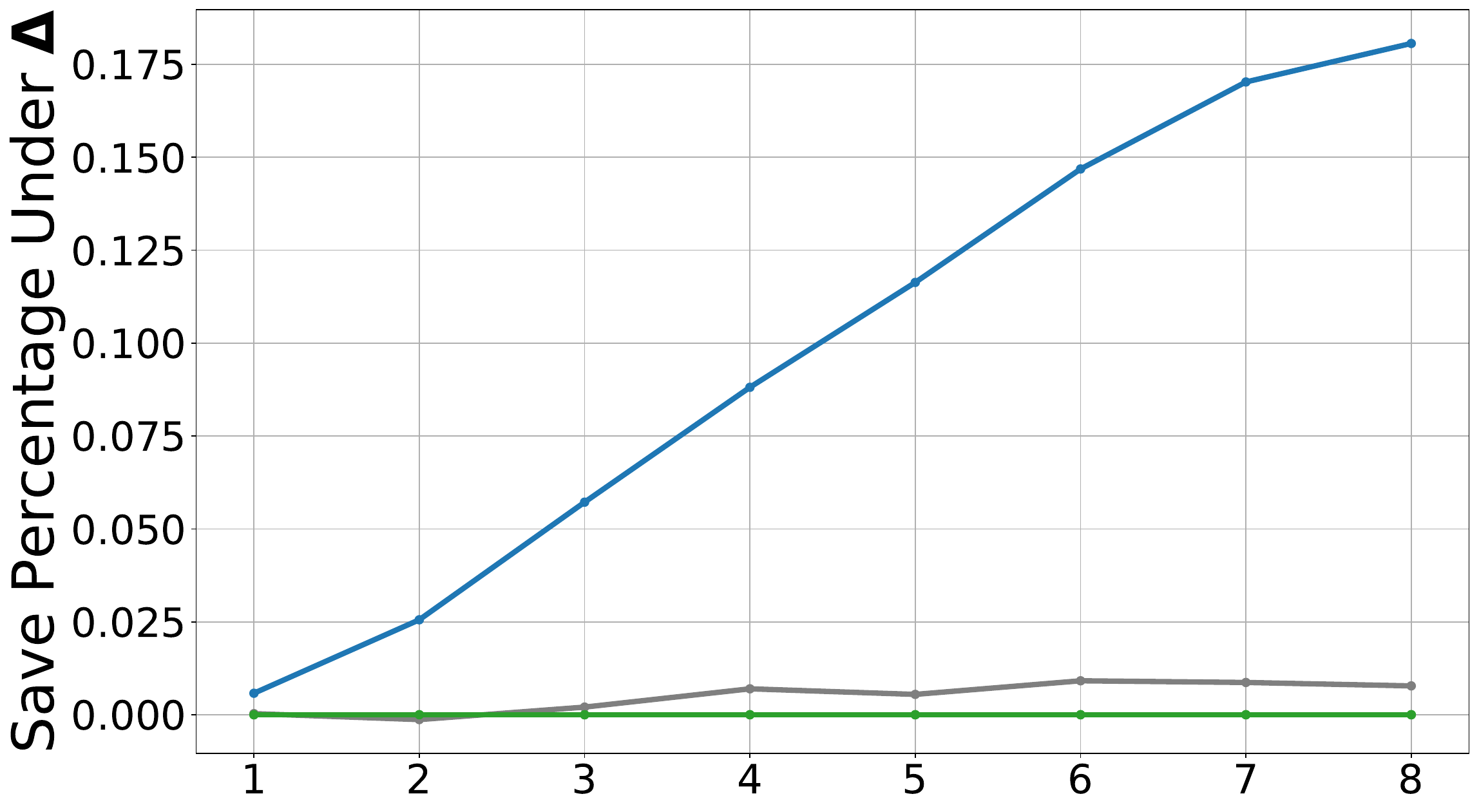}
    }
    \hfill
    \subfigure{
    \label{fig:4o5}
        \includegraphics[width=0.31\textwidth]{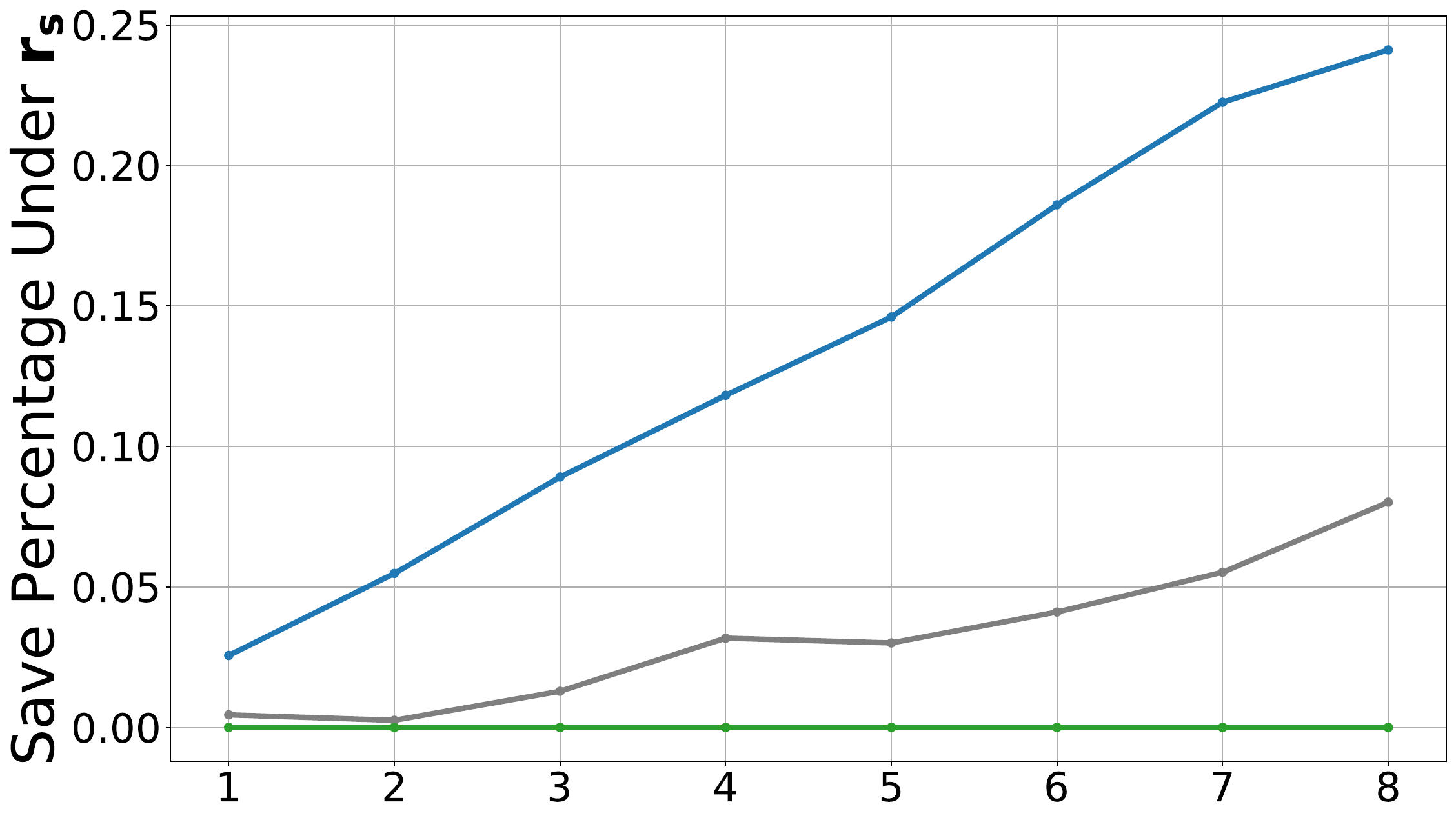}
    }
    \hfill
    \subfigure{
    \label{fig:4o6}
        \includegraphics[width=0.31\textwidth]{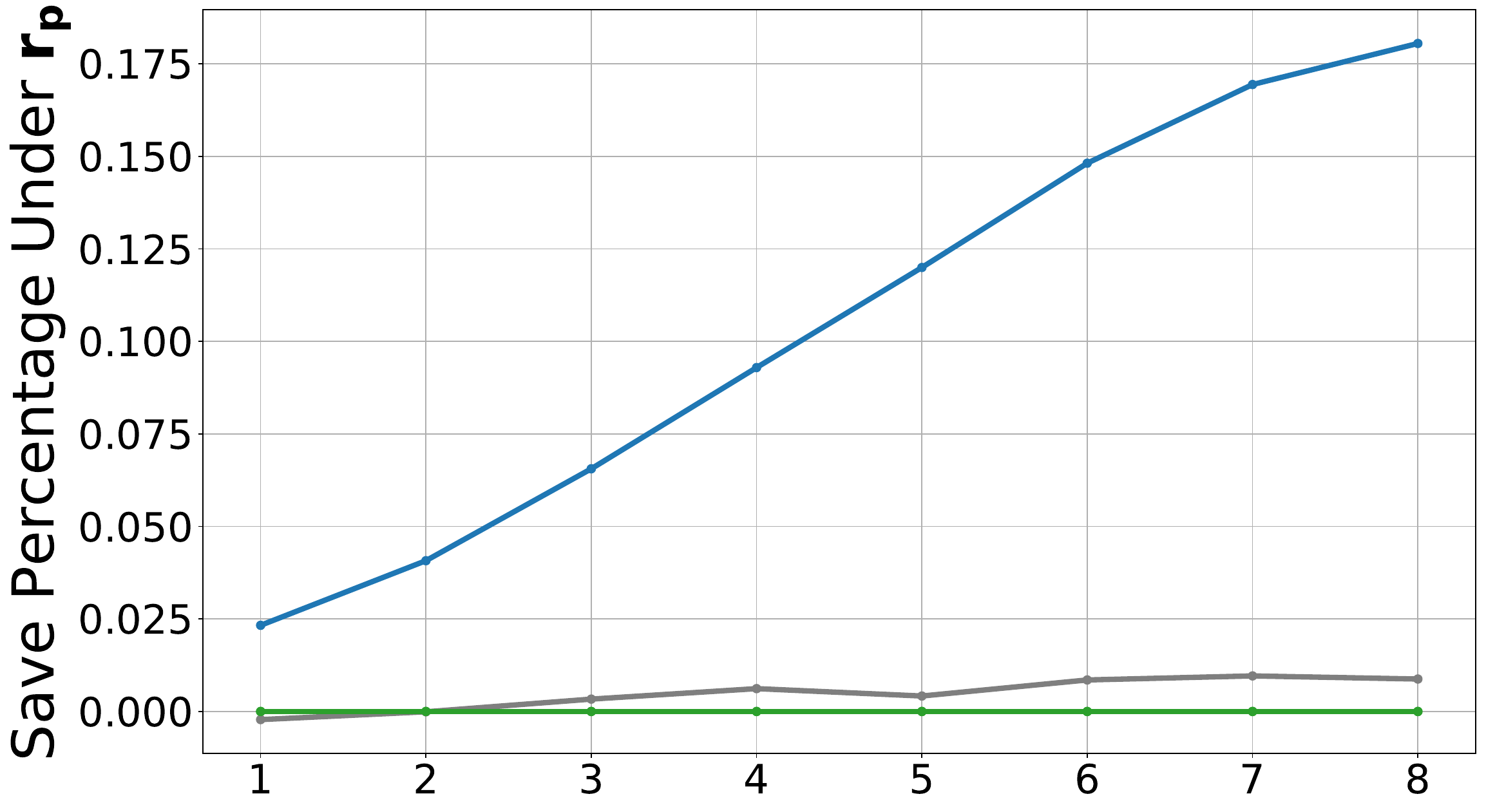}
    }
    \vspace{-0.4cm}
    \caption{Results of compared CBE methods with GPT-4o as the judge on AlpacaEval benchmark. The X-axis (applicable to all plots below) represents the preference budget ($k$). $\mathbf{\Delta}$ denotes the mean absolute error of the estimated win rate. $\mathbf{r_s}$ and $\mathbf{r_p}$ denote the Spearman and Pearson correlations between the the estimated model scores and the ground truth respectively.}
    \vspace{-0.4cm}
    \label{fig:main-4o}
\end{figure}
%为了确保实验结果的可靠性，对于每组实验，我们随机选择M个模型和N条数据并汇报10000个随机种子下的平均结果。

\subsection{Main Results}
\label{sec:exp-1}
\paragraph{Accuracy and Convergence.}
The results of compared CBE methods on AlpacaEval benchmark with GPT4-turbo as the judge are shown in Figure~\ref{fig:main-4o}. 
%为了更好地呈现，在第二行我们计算了在达到与Random baseline相同的性能时各方法所节省的preference budget比例。
To better illustrate the results, we also calculate the percentage of preference budget saved by each method compared to \textsc{Random} baseline when achieving the same performance. 
In terms of performance, \textsc{AlpacaEval} $<<$ \textsc{Random} $<$ \textsc{Arena} $<$ \textsc{UniCBE}.
%从表现而言，\textsc{AlpacaEval}<<\textsc{Random}<\textsc{Arena}<\textsc{UniCBE}.
To understand the differences in the performance of each method, we quantitatively analyze them based on the guidelines summarized in \S~\ref{sec:3}. 
% Specifically, we calculate the extent to which these methods satisfy the objectives for suppressing sampling bias, balancing the uncertainty descent process, and mitigating updating uncertainty, denoted as $\beta_{acc}$, $\beta_{con}$, and $\beta_{sca}$, respectively (See Appendix for detailed equations). 
To achieve accuracy, convergence, and scalability, it is supposed  to allocate the preference budget in a way that ensures uniformity across tuples, uniformity across model pairs in win-rate uncertainty, and uniformity across models. 
We calculate the cosine similarity between the allocation results of these methods and the corresponding expected uniform vectors for each objective as a measure, denoted as $\beta_{acc}$, $\beta_{con}$, and $\beta_{sca}$, respectively (see Appendix~\ref{app:beta} for calculating process).
%由于实现准确性、收敛性和可扩展性分别要求preference budget的分配满足tuple间的均匀性、胜率不确定度的均匀性、模型间的均匀性，我们计算了这些方法与对应期望的均匀向量之间的余弦相似度作为衡量，并分别记作$\beta_{acc}$,$\beta_{con}$,\beta_{sca}.
% 为了理解各方法表现的差异性，我们从第三节所总结的三条guidelines来量化分析它们。具体来说我们分别计算了这些方法对于suppressing sampling bias, balancing the uncertainty descent process, and mitigating updating uncertainty的满足程度，并分别记作$\beta_{acc}$,$\beta_{con}$,\beta_{sca}.
As shown in Table~\ref{tab:beta}, the fixed inclusion of the reference model in the tuple selection of \textsc{AlpacaEval} compromises uniformity across multiple aspects, thereby resulting in lower $\beta$ values and significantly poorer performance. 
%由于\textsc{AlpacaEval} 的tuple选择中固定包含reference model，因此在多方面都无法保证较好的均匀性，从而导致了较低的$\beta$值和明显较差的性能。
\textsc{Arena} and \textsc{Random} respectively improve the balance of uncertainty and suppression of sampling bias, resulting in higher $\beta_{con}$ and $\beta_{acc}$ values.
Following our guidelines, \textsc{UniCBE} improves $\beta_{con}$, $\beta_{acc}$, and $\beta_{sca}$ simultaneously and save over 17\% of the preference budget compared to \textsc{Random} with a $\Delta$ close to 0.01, showcasing improved accuracy and convergence.
\begin{table}[h]
\renewcommand\arraystretch{1.4}
\centering
\setlength{\tabcolsep}{0.6em} 
\vspace{-0.4cm}
\caption{The measurement results of the achievement of objectives in \S\ref{sec:3} for the compared methods.}
\begin{tabular}{lcccc}
\toprule
Methods&\textsc{Random}&\textsc{Arena}&\textsc{AlpacaEval}&\textsc{UniCBE}\\
\midrule
$\beta_{acc}$&.5803&.5725&.0925&.7364\\
$\beta_{con}$&.9081&.9172&.3515&.9228\\
$\beta_{sca}$&.9972&.9945&.4987&.9997\\
\bottomrule
\end{tabular}
\vspace{-0.4cm}
\label{tab:beta}
\end{table}

\begin{figure}[h]
    \centering
    \subfigure{
    \label{fig:4oadd1}
        \includegraphics[width=0.31\textwidth]{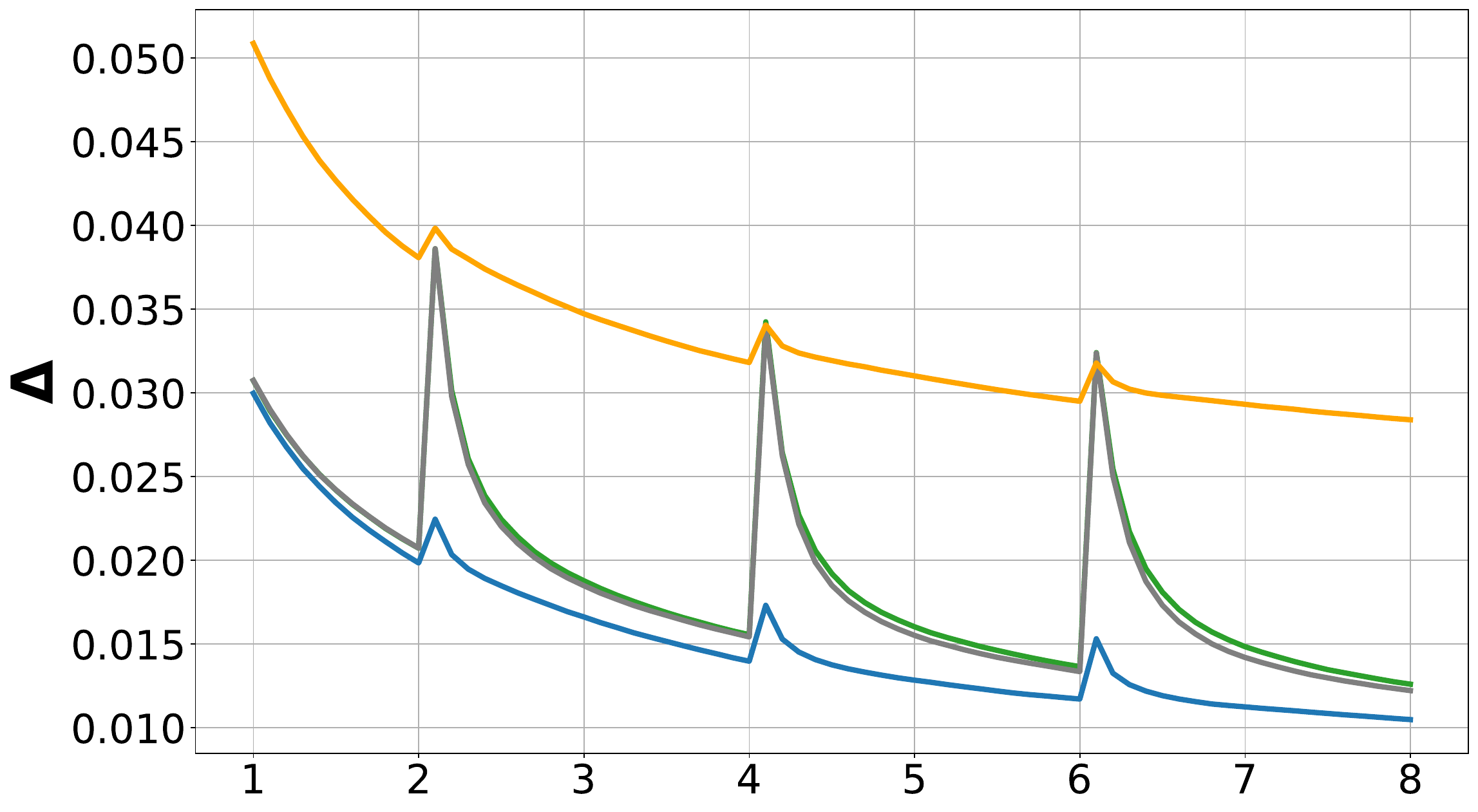}
    }
    \hfill
    \subfigure{
    \label{fig:4oadd2}
        \includegraphics[width=0.31\textwidth]{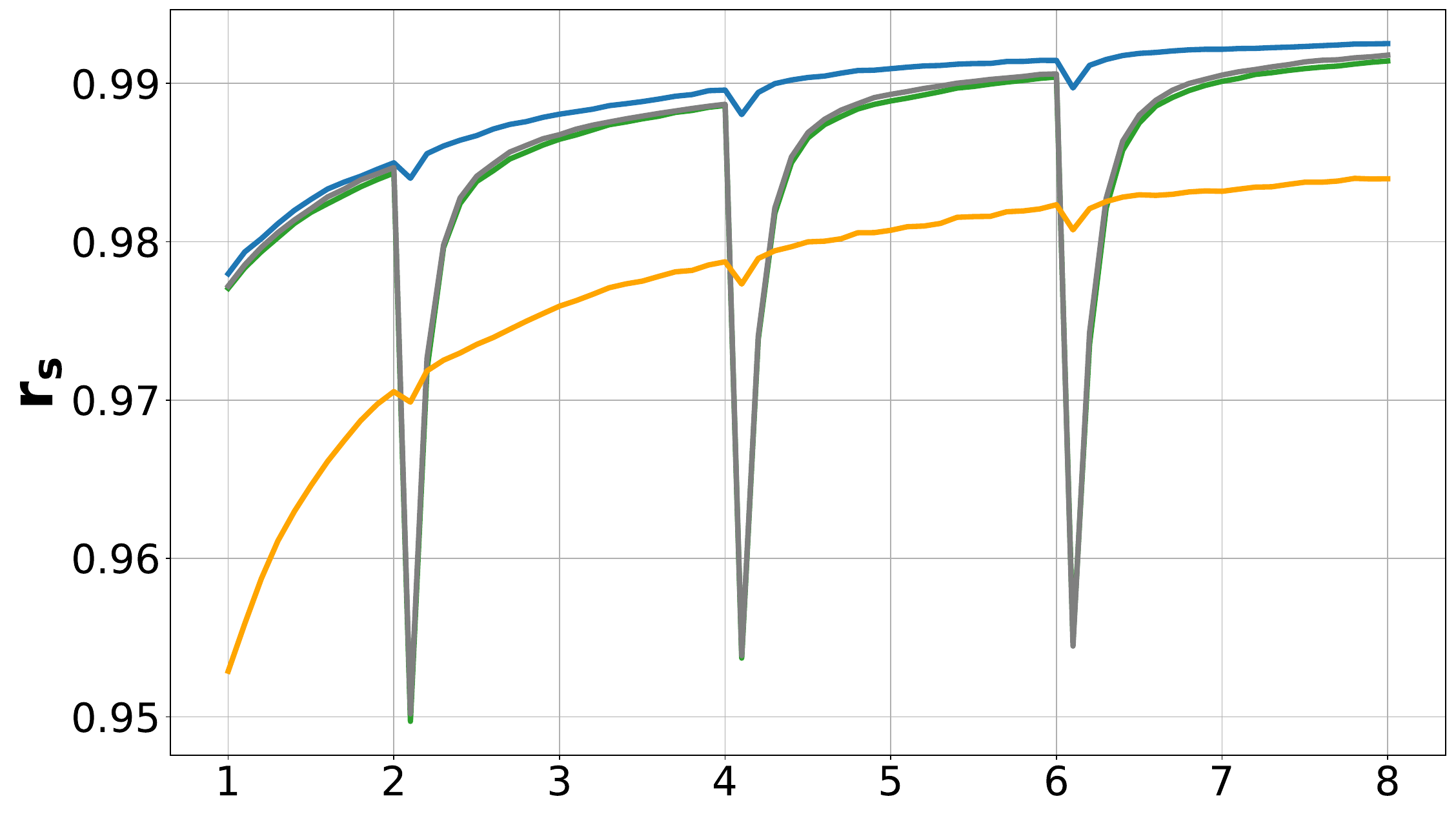}
    }
    \hfill
    \subfigure{
    \label{fig:4oadd3}
        \includegraphics[width=0.31\textwidth]{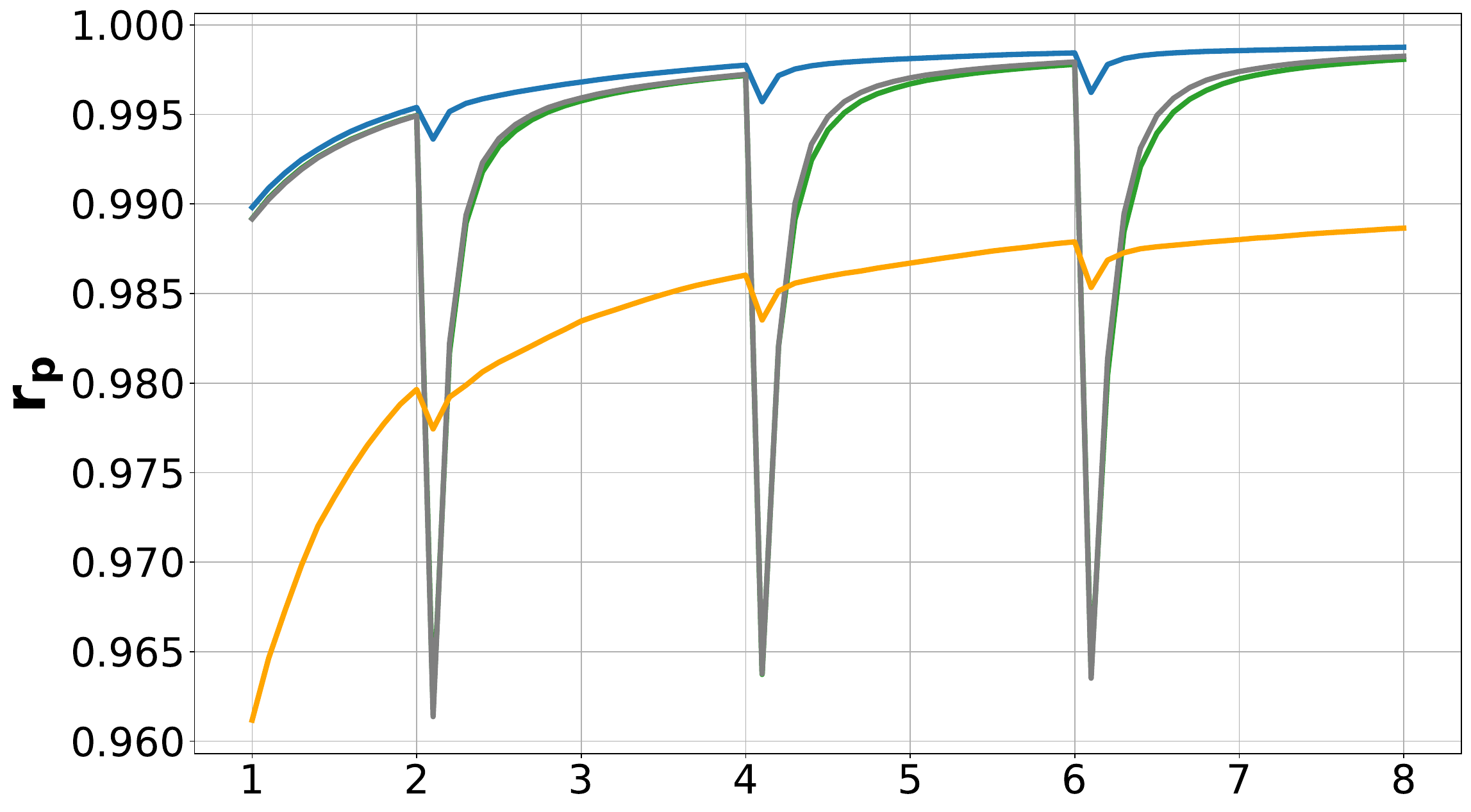}
    } \\
    \vspace{-0.2cm}
    \subfigure{
    \label{fig:4oadd4}
        \includegraphics[width=0.31\textwidth]{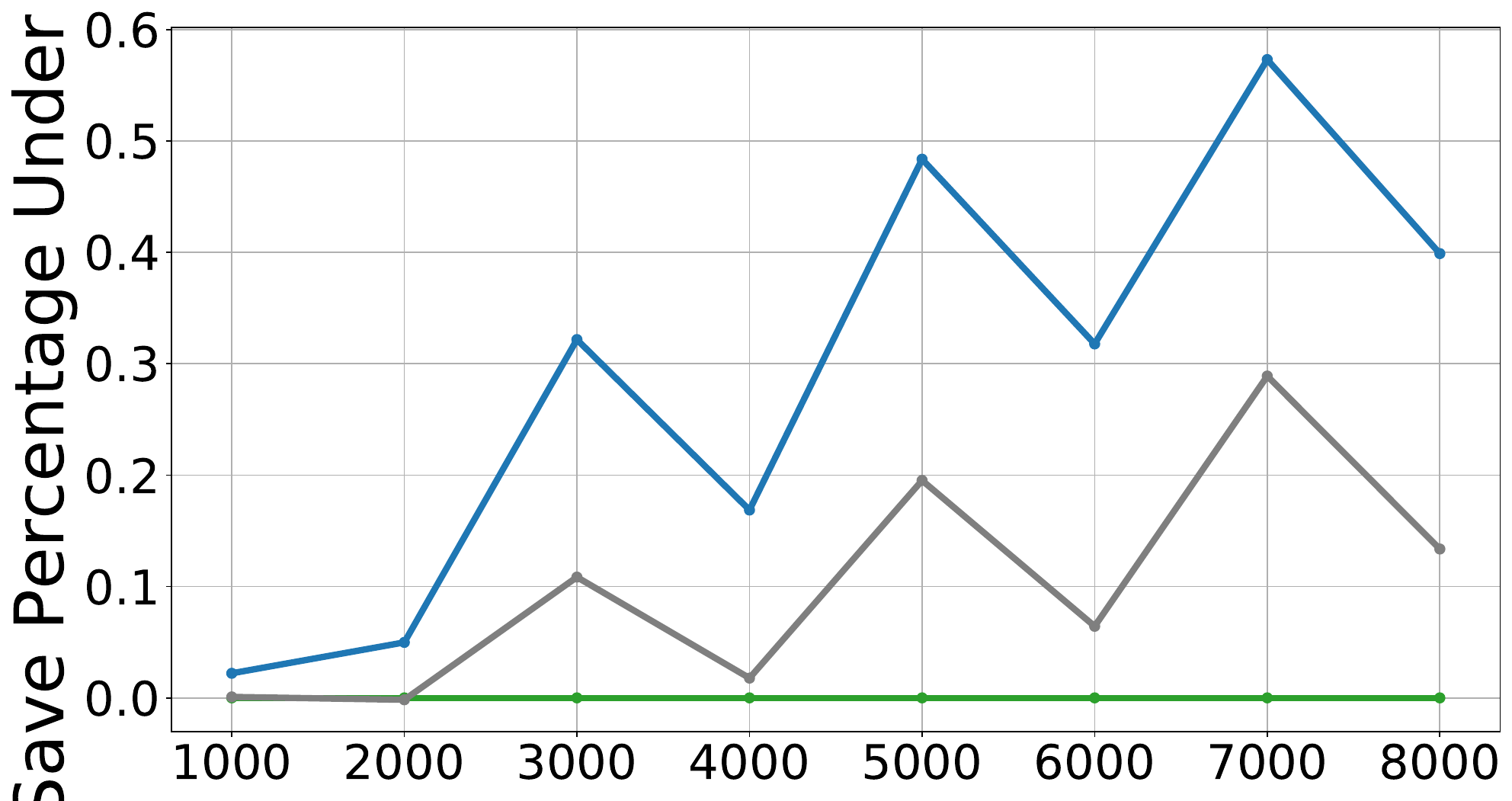}
    }
    \hfill
    \subfigure{
    \label{fig:4oadd5}
        \includegraphics[width=0.31\textwidth]{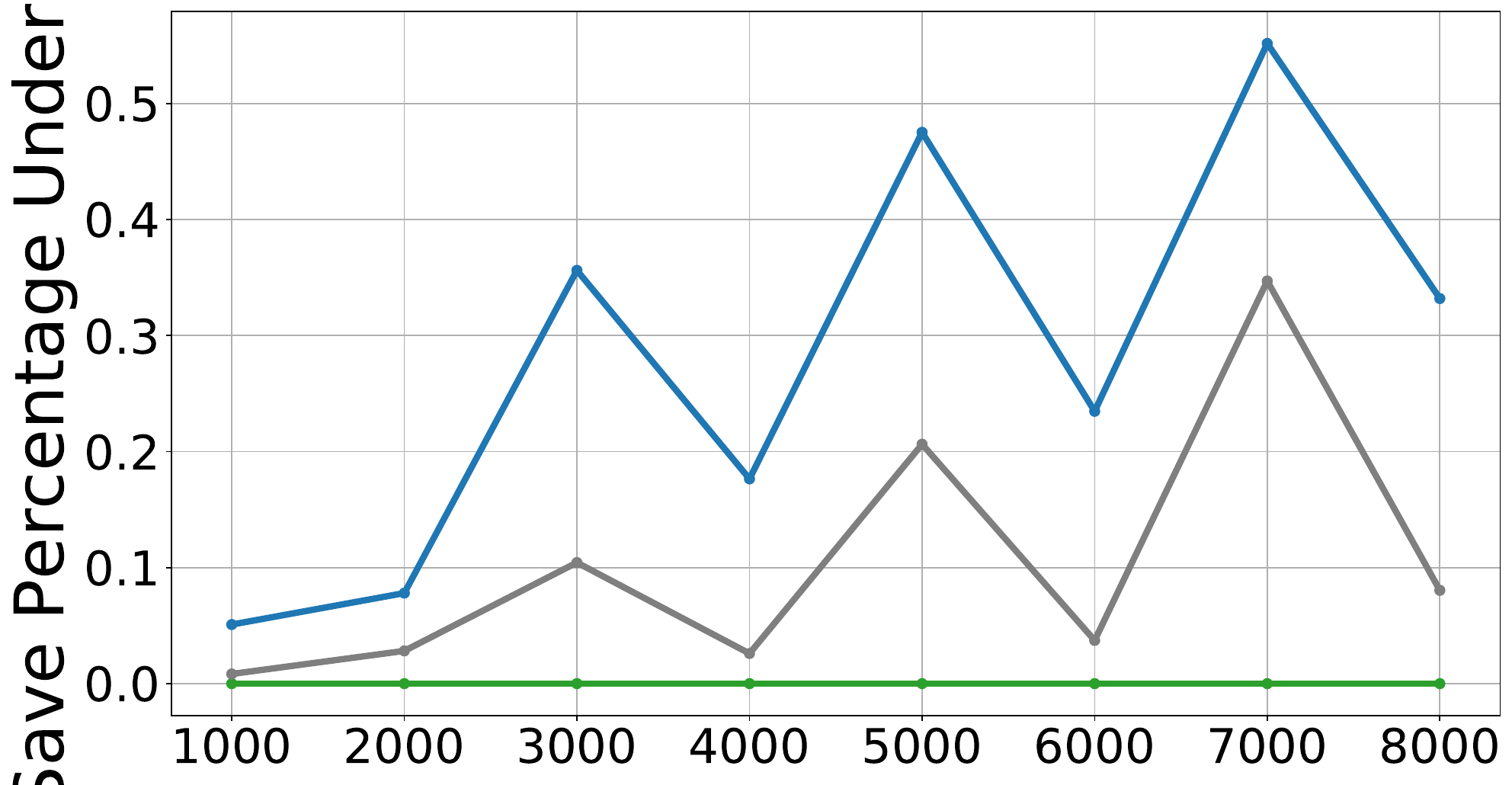}
    }
    \hfill
    \subfigure{
    \label{fig:4oadd6}
        \includegraphics[width=0.31\textwidth]{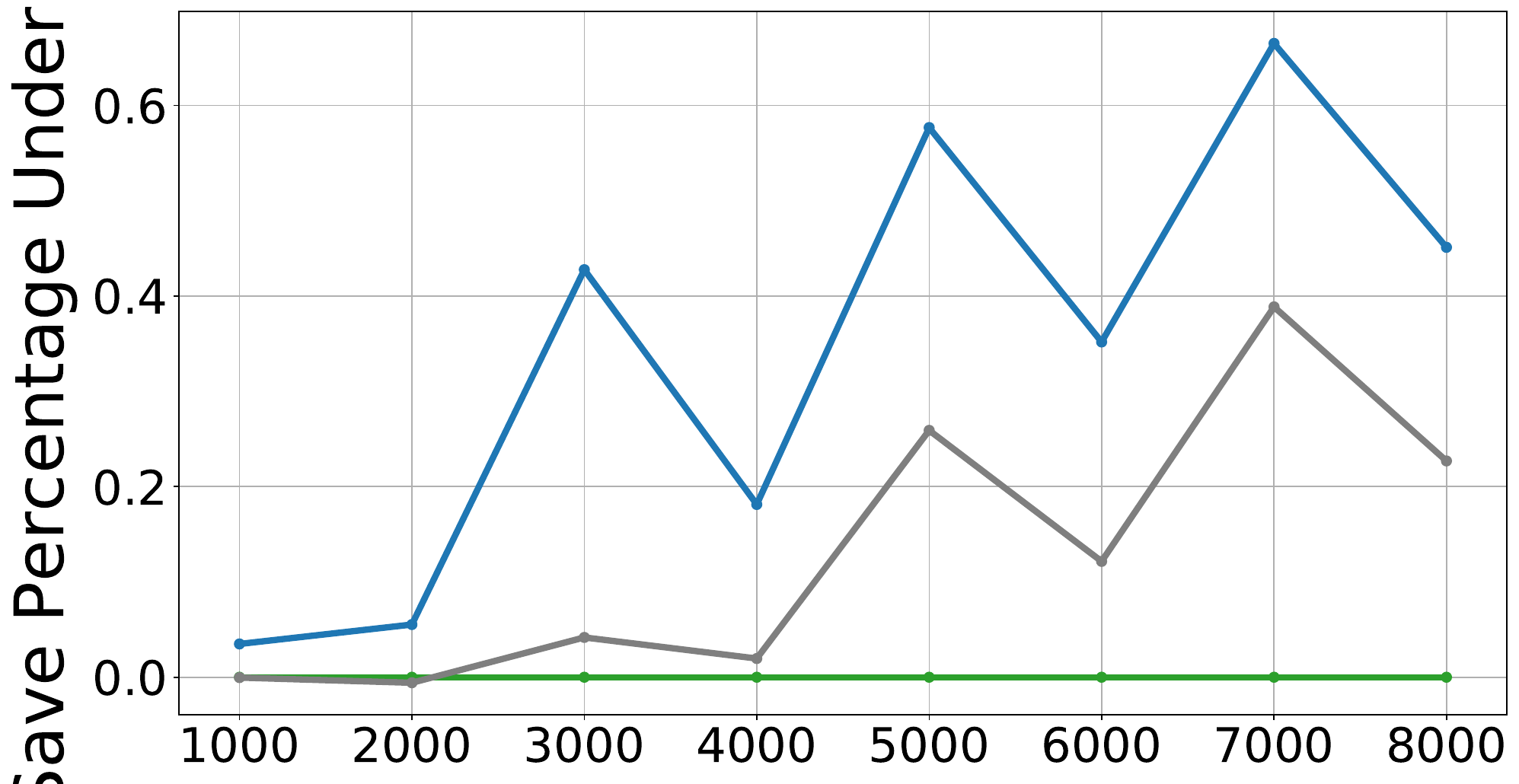}
    }
    \vspace{-0.2cm}
    \caption{Results of compared CBE methods in the scenario where new model are consistently introduced every 2000 iterations. }
    \vspace{-0.2cm}
    \label{fig:main-4o-add}
\end{figure}

% \textsc{UniCBE}，通过遵循guidelines同时提升了$\beta_{con}$， $\beta_{acc}$ 和 $\beta_{sca}$，从而在达到相同的性能时(接近0.01的$\Delta$)相比于\textsc{Random}节省了超过17%的preference budget，展现了良好的accuracy和convergence
\paragraph{Scalability.}
%To analyze the scalability, we 设置了一个场景where开始有11个待评测模型，然后每采样2000次就有一个新模型加入。
To analyze scalability, we establish a scenario where we initially have 11 models awaiting assessment, and new models are sequentially added every 2000 samplings.
As shown in Figure~\ref{fig:main-4o-add}, 
%在每次有新模型被引入时，UniCBE都能通过preference budget的自适应倾斜实现性能的快速稳定，相比于\textsc{Random} baseline甚至能节省超过50%的budget。相比之下arena和random因为没有考虑相应的优化目标展现出较差的scalability. 尽管分配给reference model的budget显著多于其他模型导致AlpacaEval的\beta_{sca}较低，但新模型被引入时会自动分配budget的策略也给它带来了较好的可扩展性。
Whenever a new model is introduced, \textsc{UniCBE} can rapidly stabilize the performance through adaptive preference allocation skewing for the new model, saving over 50\% of the budget compared to the \textsc{Random} baseline. In contrast, \textsc{Arena} and \textsc{Random} exhibit poorer scalability since they do not consider scalability as optimization objective. Although the budget allocated to the reference model is significantly more than that for other models, resulting in a lower \(\beta_{sca}\) for \textsc{AlpacaEval}, the strategy of automatically allocating the budget to the new introduced models also provides it with good scalability.

\subsection{Variants Ablations}
\label{sec:exp-2}
\begin{figure}[t]
    \centering
    \subfigure{
    \label{fig:4oa1}
        \includegraphics[width=0.31\textwidth]{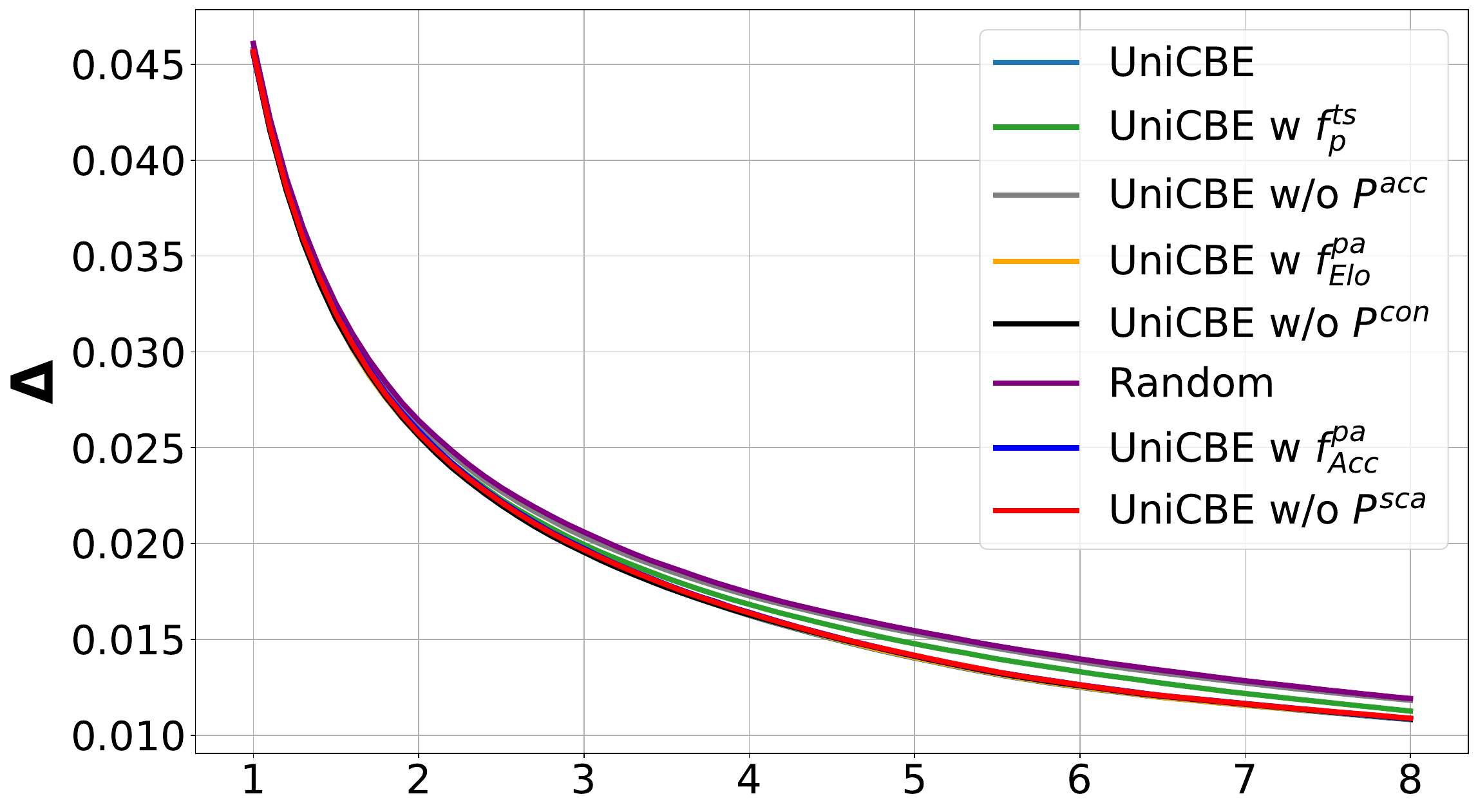}
    }
    \hfill
    \subfigure{
    \label{fig:4oa2}
        \includegraphics[width=0.31\textwidth]{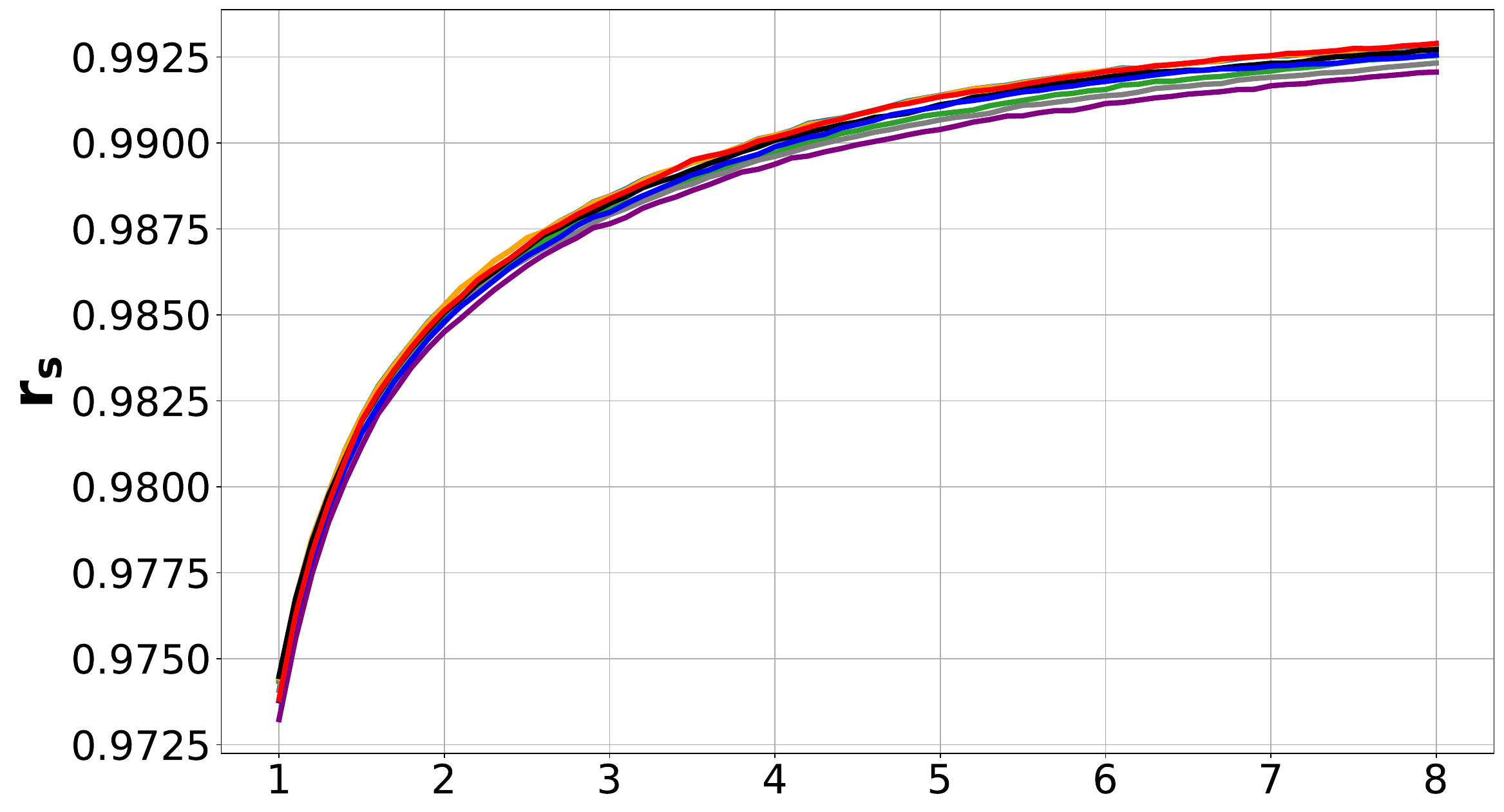}
    }
    \hfill
    \subfigure{
    \label{fig:4oa3}
        \includegraphics[width=0.31\textwidth]{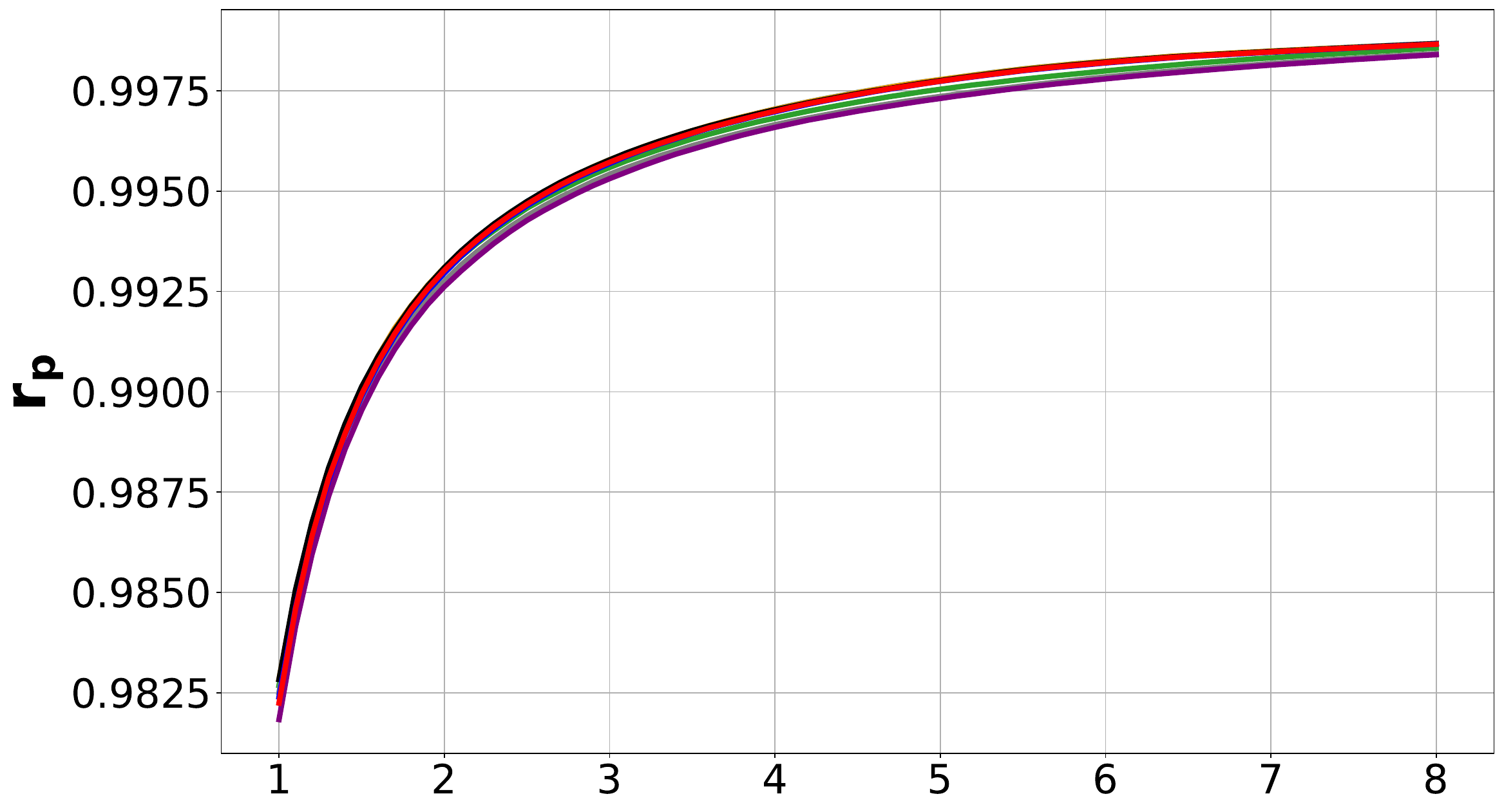}
    } \\
    \subfigure{
    \label{fig:4oa4}
        \includegraphics[width=0.31\textwidth]{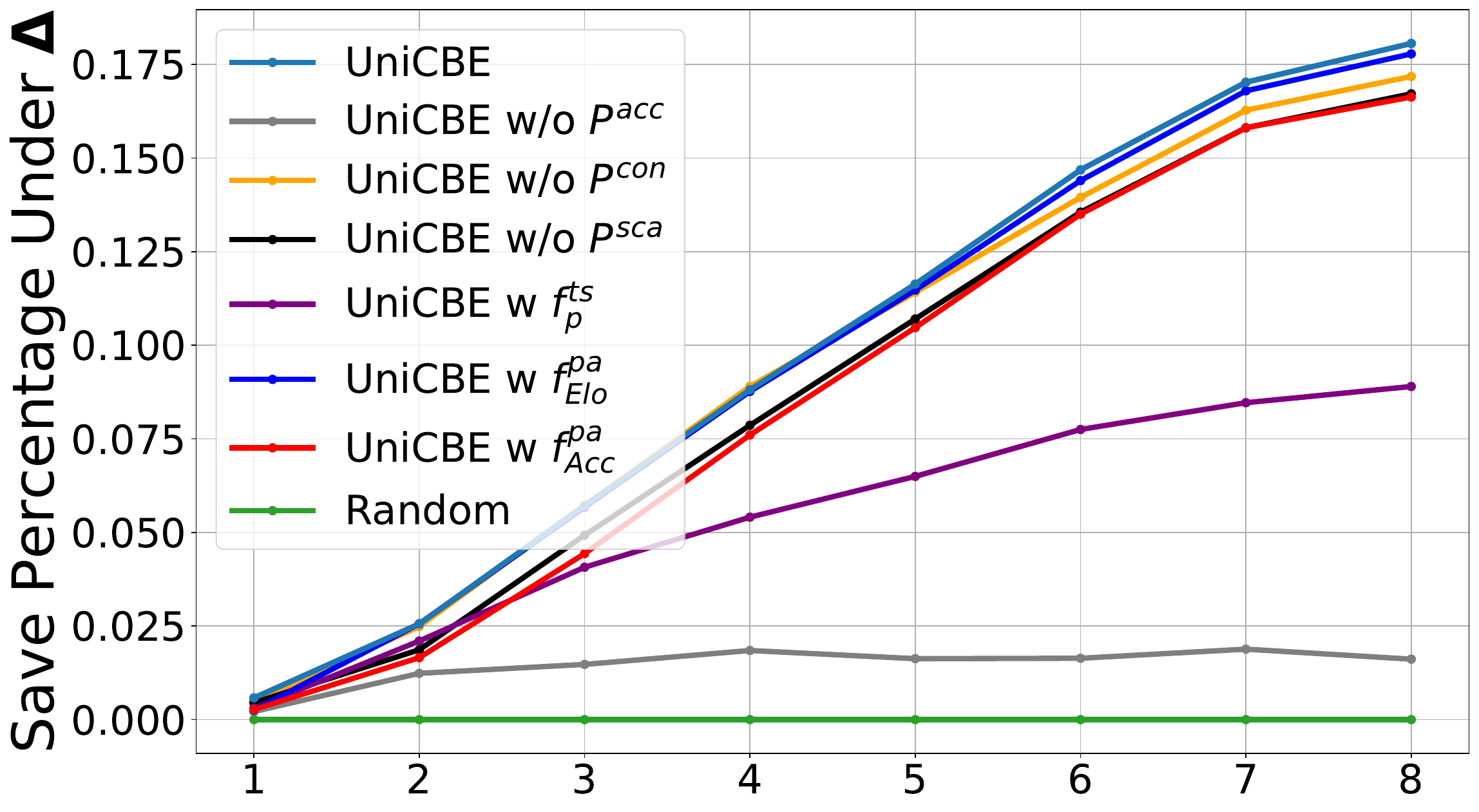}
    }
    \hfill
    \subfigure{
    \label{fig:4oa5}
        \includegraphics[width=0.31\textwidth]{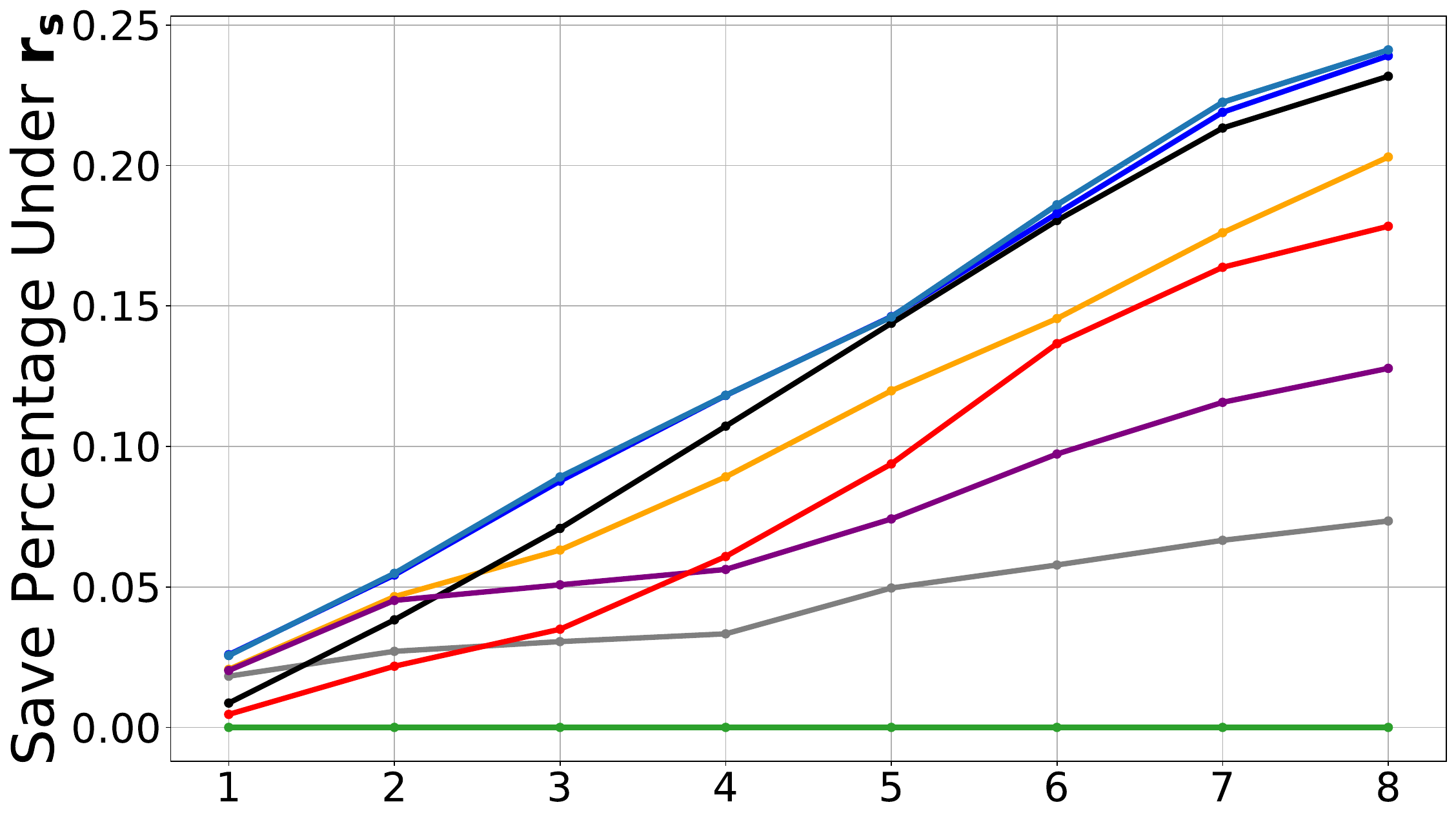}
    }
    \hfill
    \subfigure{
    \label{fig:4oa6}
        \includegraphics[width=0.31\textwidth]{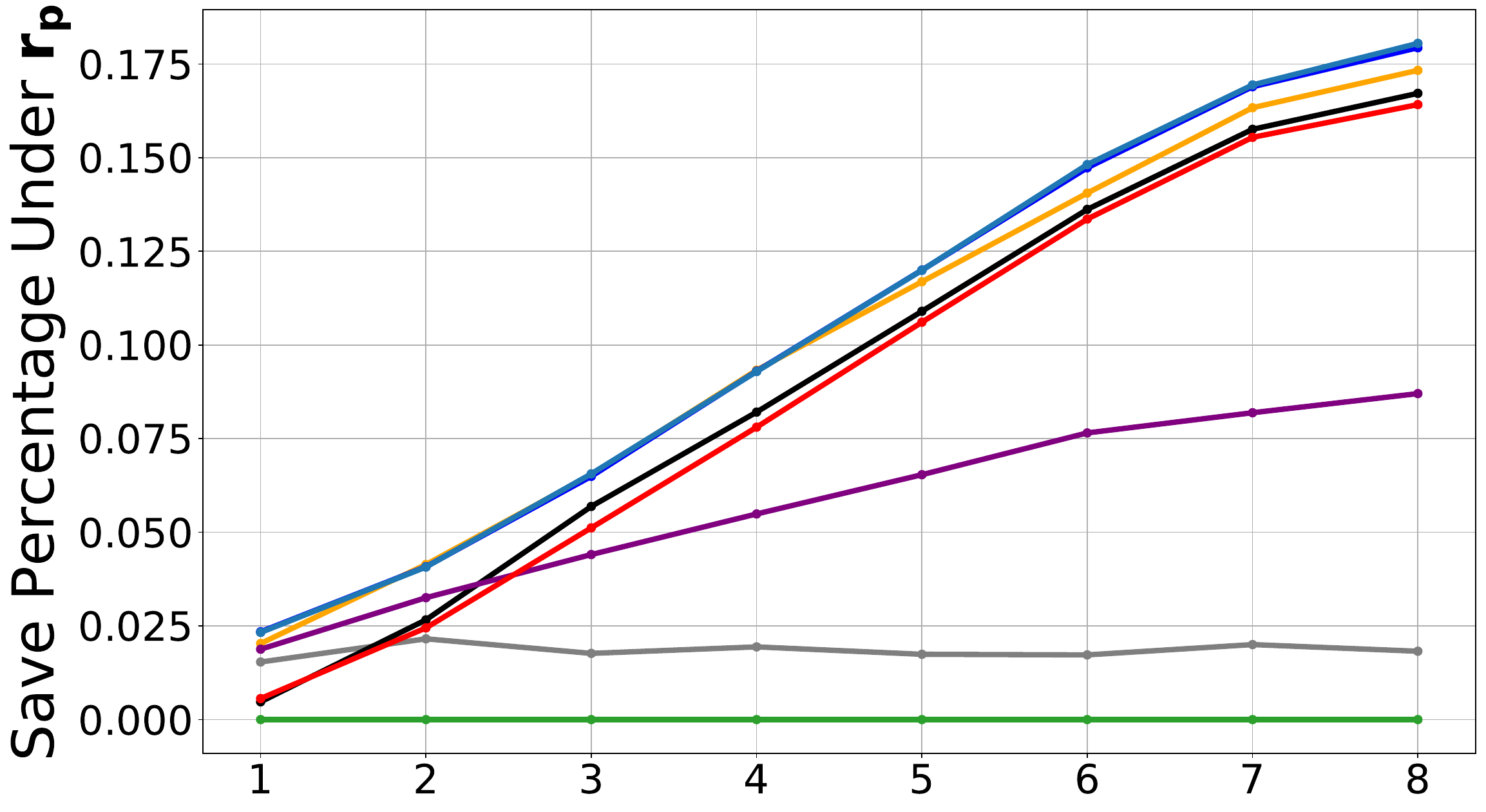}
    }
    \vspace{-0.2cm}
    \caption{Ablation studies of \textsc{UniCBE} with GPT-4o as the judge on AlpacaEval benchmark. }
    \vspace{-0.4cm}
    \label{fig:main-4o-ab}
\end{figure}

\paragraph{Budget Allocation Objectives.}
We test the impact of different optimization objectives by removing \(P^{acc}\), \(P^{con}\), and \(P^{sca}\) from~\eqref{eq:14} separately. As shown in Figure~\ref{fig:main-4o-ab}, the significant performance degradation observed when removing \( P^{acc} \) from \textsc{UniCBE} indicates that mitigating sampling bias to improve accuracy is the most critical factor in achieving efficient CBE. Furthermore, we find that \( P^{con} \) has a considerable impact on \( r_s \). We hypothesize that this is because balancing the uncertainty among different models helps prevent any one model from having a significant ranking bias due to its larger uncertainty. The performance drop when removing \( P^{sca} \) also suggests that ensuring uniformity in sampling across models not only enhances scalability but also further reduces sampling bias, thereby improving accuracy.
%\textsc{UniCBE}在去掉P^{acc}后出现了显著的表现下滑，这说明通过抑制采样偏差来提升准确性是实现高效CBE的最核心要素。此外，我们发现\(P^{con}\)对$r_s$的影响很大，我们猜想这是因为平衡各个模型之间的不确定度能够避免某个模型因较大的不确定度而出现明显的秩序偏差。去掉$P^{sca}$也会导致\textsc{UniCBE}下滑说明平衡模型间的采样均匀度除了能提升可扩展性，还可以进一步降低采样偏差从而提升准确性。

\paragraph{Tuple Sampling and Preference Aggregation Strategies.} 
%如图所示，如果将greedy sampling替换为按照概率采样，性能将会出现显著的下降。这很可能是因为按概率采样引入的随机性会阻碍多个优化目标的实现程度。在preference aggregation 策略方面，Elo rating由于具有较强的不稳定性导致了性能相比BT model有略微下降。而直接使用胜率平均值的策略则因没有考虑不同模型所面临对手能力的不同而引入了额外的bias，导致了性能的下降。
As shown in Figure~\ref{fig:main-4o-ab}, replacing greedy sampling with probabilistic sampling $f^{ts}_p$ results in a significant performance drop. This is likely because the randomness introduced by $f^{ts}_p$ hinders the achievement of multiple optimization objectives. In terms of preference aggregation strategies, the Elo rating system $f^{pa}_{Elo}$ shows a slight performance decline compared to the BT model due to its higher instability \citep{eloun}. Moreover, the strategy of directly using the average win rate $f^{pa}_{avg}$ may introduce additional bias, as it fails to consider the varying strengths of the opponents faced by different models, leading to a performance decrease.

\subsection{Generalizability under Different Settings}
\label{sec:exp-3}
\begin{figure}[h]
    \centering
    \subfigure{
    \label{fig:351m}
        \includegraphics[width=0.31\textwidth]{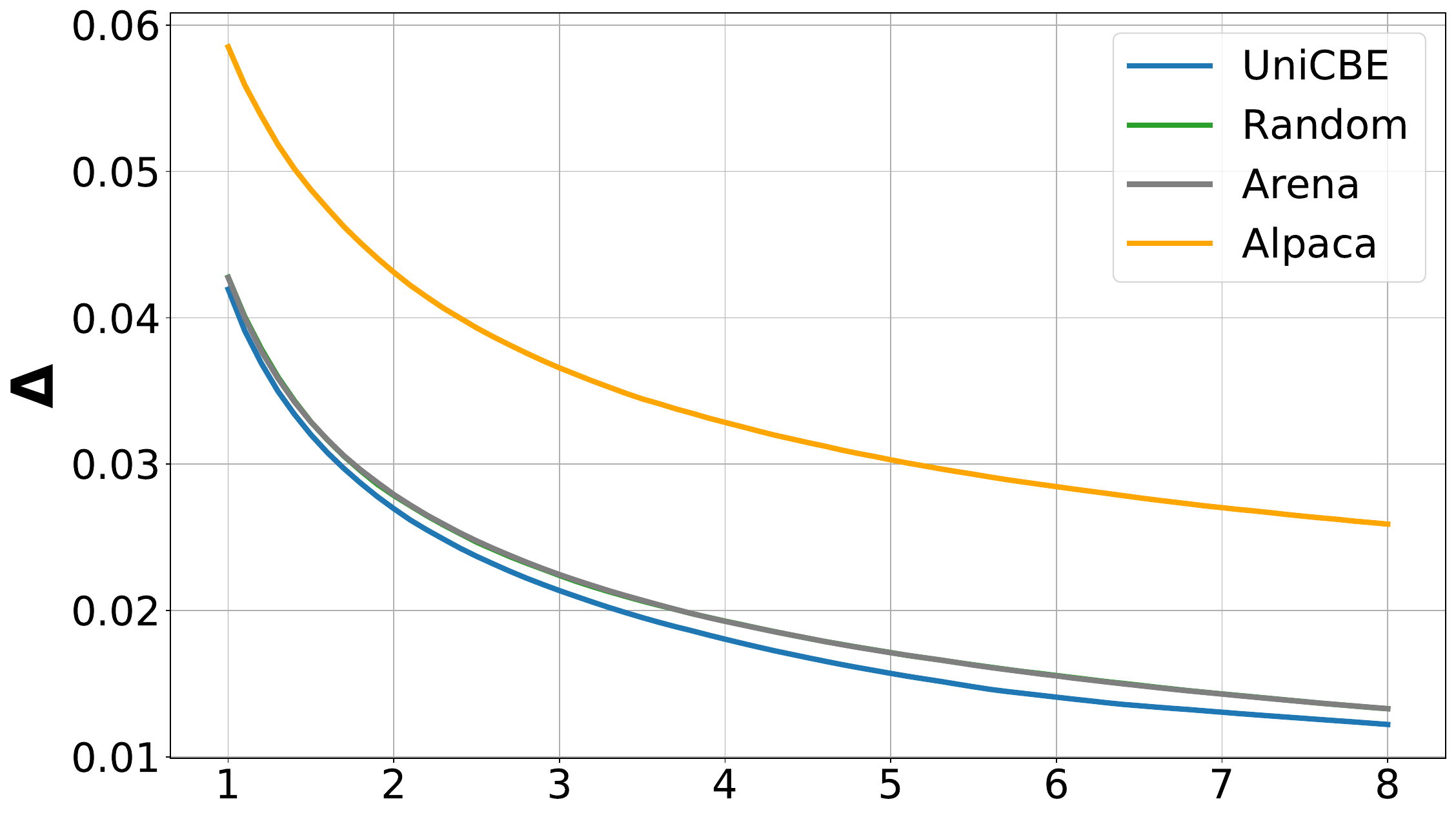}
    }
    \hfill
    \subfigure{
    \label{fig:355m}
        \includegraphics[width=0.31\textwidth]{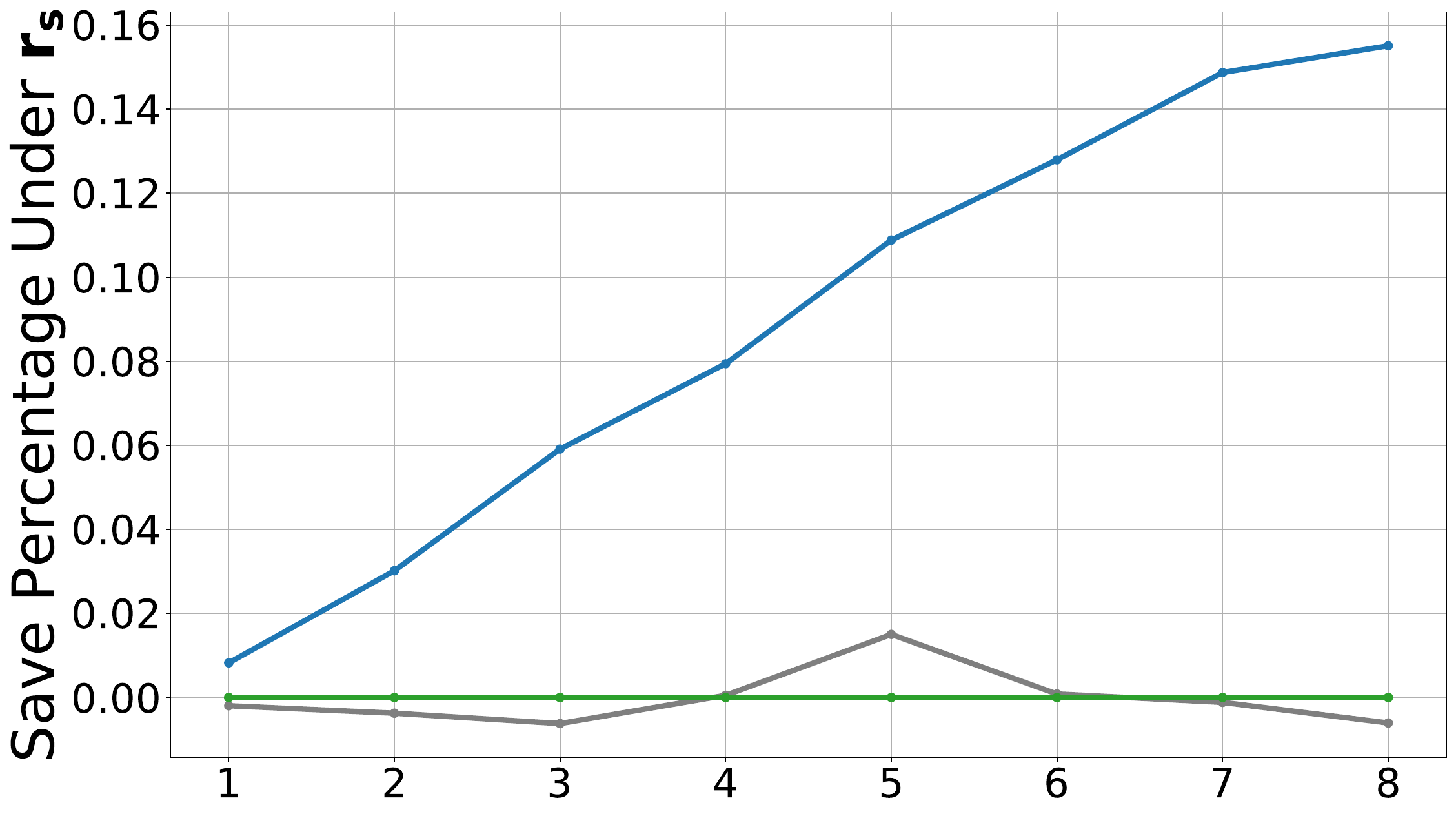}
    }
    \hfill
    \subfigure{
    \label{fig:356m}
        \includegraphics[width=0.31\textwidth]{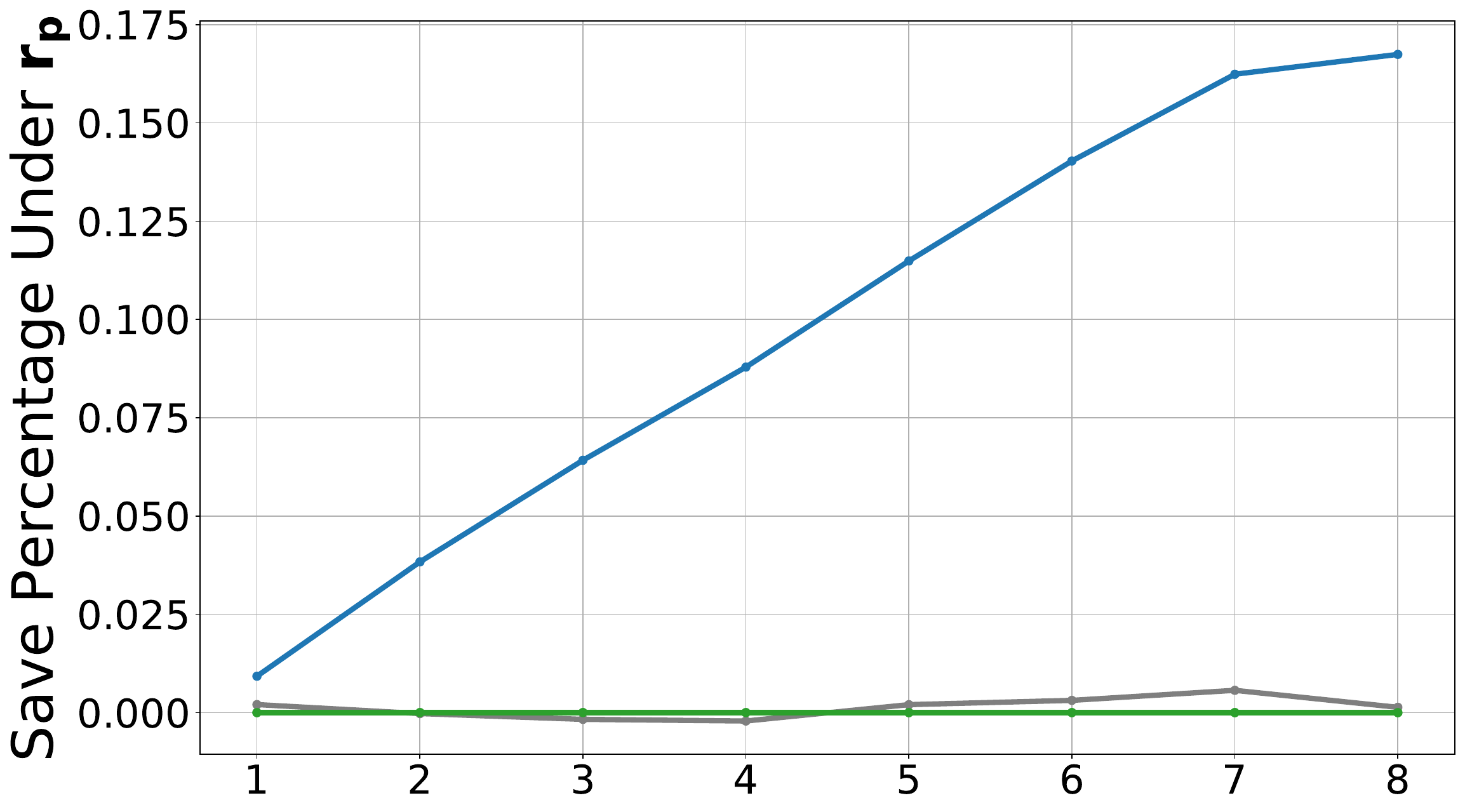}
    } \\
    \subfigure{
    \label{fig:mt1m}
        \includegraphics[width=0.31\textwidth]{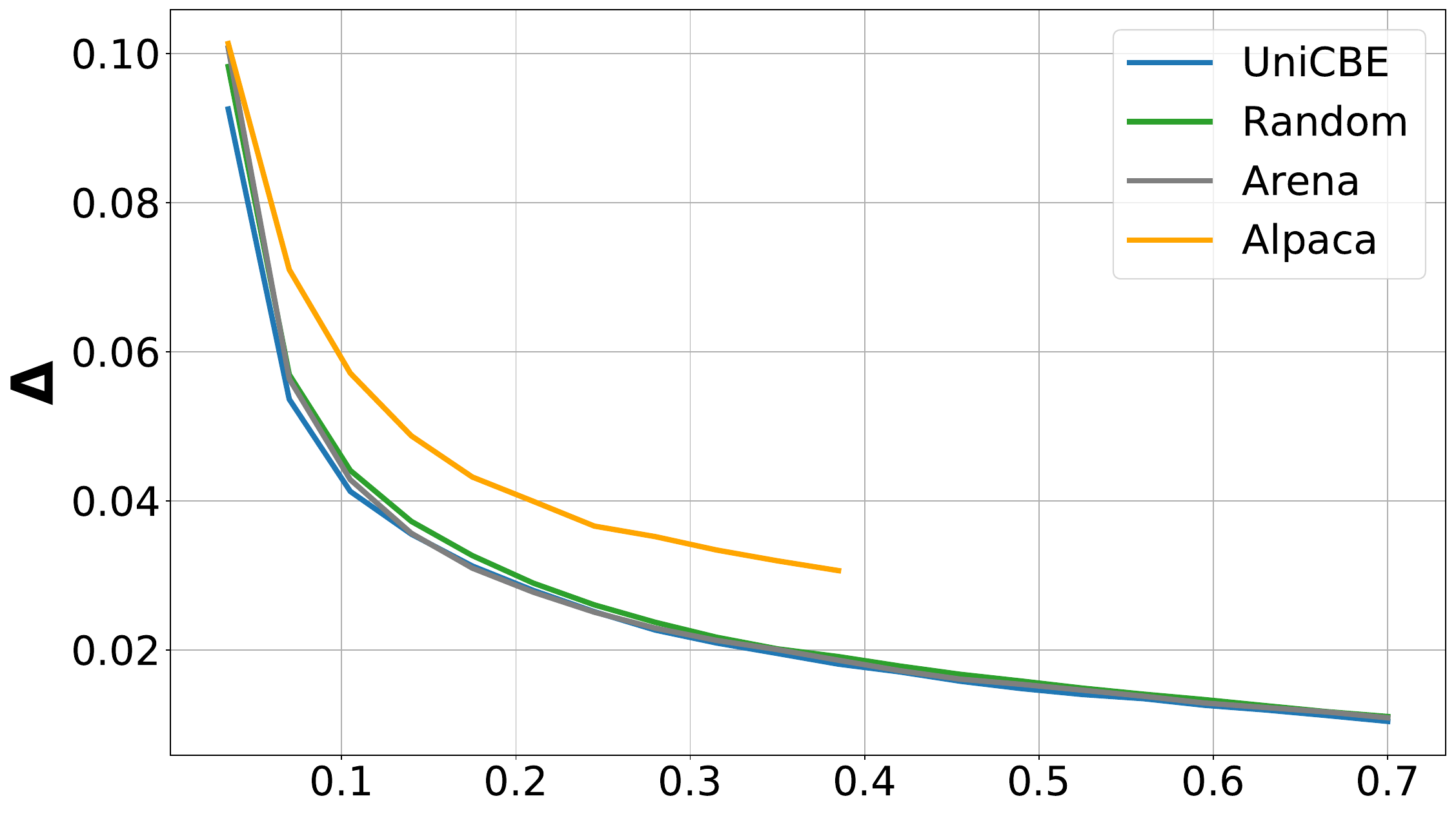}
    }
    \hfill
    \subfigure{
    \label{fig:mt5m}
        \includegraphics[width=0.31\textwidth]{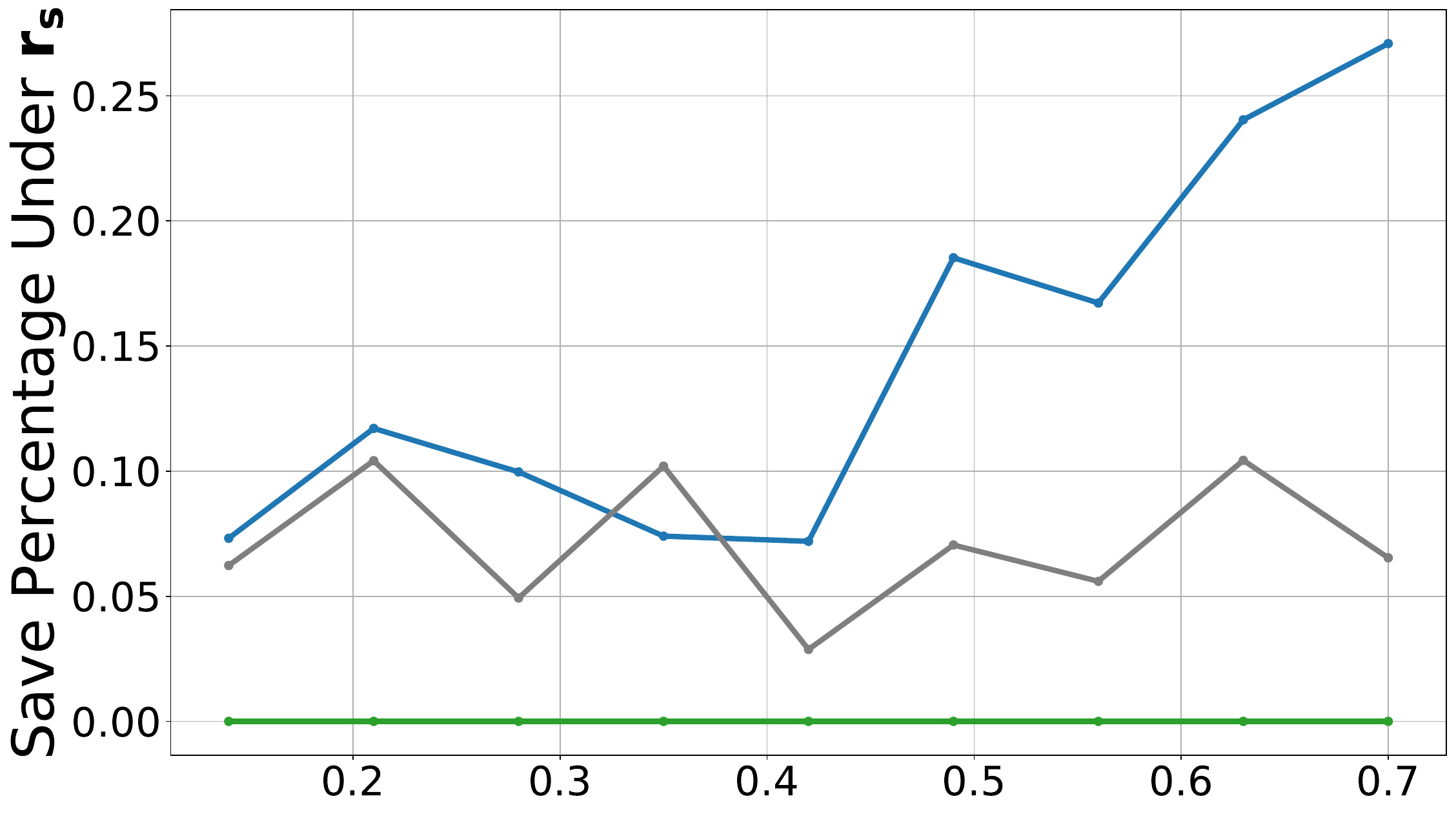}
    }
    \hfill
    \subfigure{
    \label{fig:mt6m}
        \includegraphics[width=0.31\textwidth]{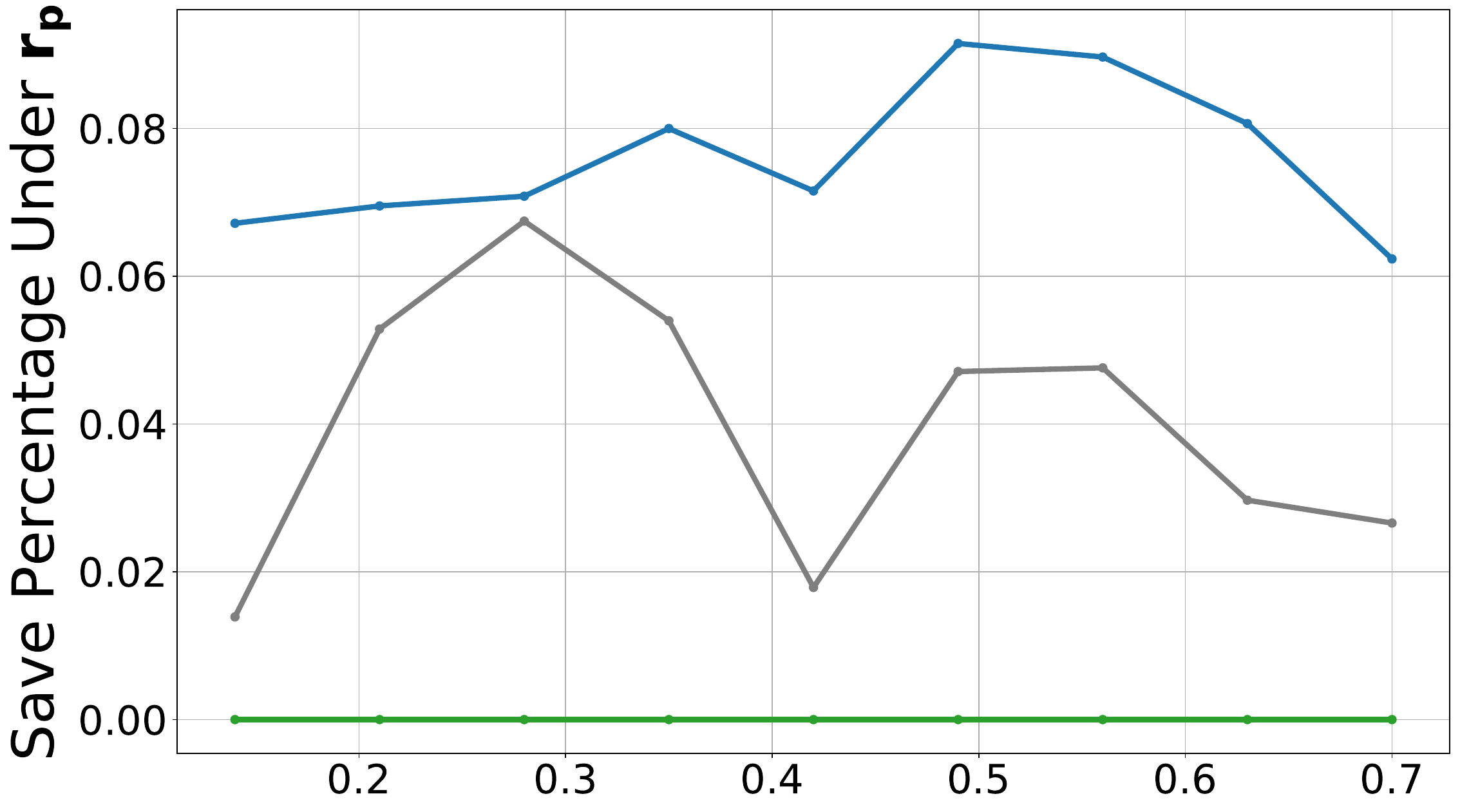}
    }
    \vspace{-0.2cm}
    \caption{Results of compared CBE methods with GPT-3.5-turbo as the judge on AlpacaEval (above) and human as the judge on MT-Bench (below). }
    \vspace{-0.4cm}
    \label{fig:main-diffjudge}
\end{figure}

\begin{figure}[htbp]
    \centering
    \vspace{-0.2cm}
    \subfigure[$M=15,N=805$]{
    \label{fig:4ohy1}
        \includegraphics[width=0.31\textwidth]{figs/performance-4o-pre_wins_theta.pdf}
    }
    \hfill
    \subfigure[$M=12,N=805$]{
    \label{fig:4ohy2}
        \includegraphics[width=0.31\textwidth]{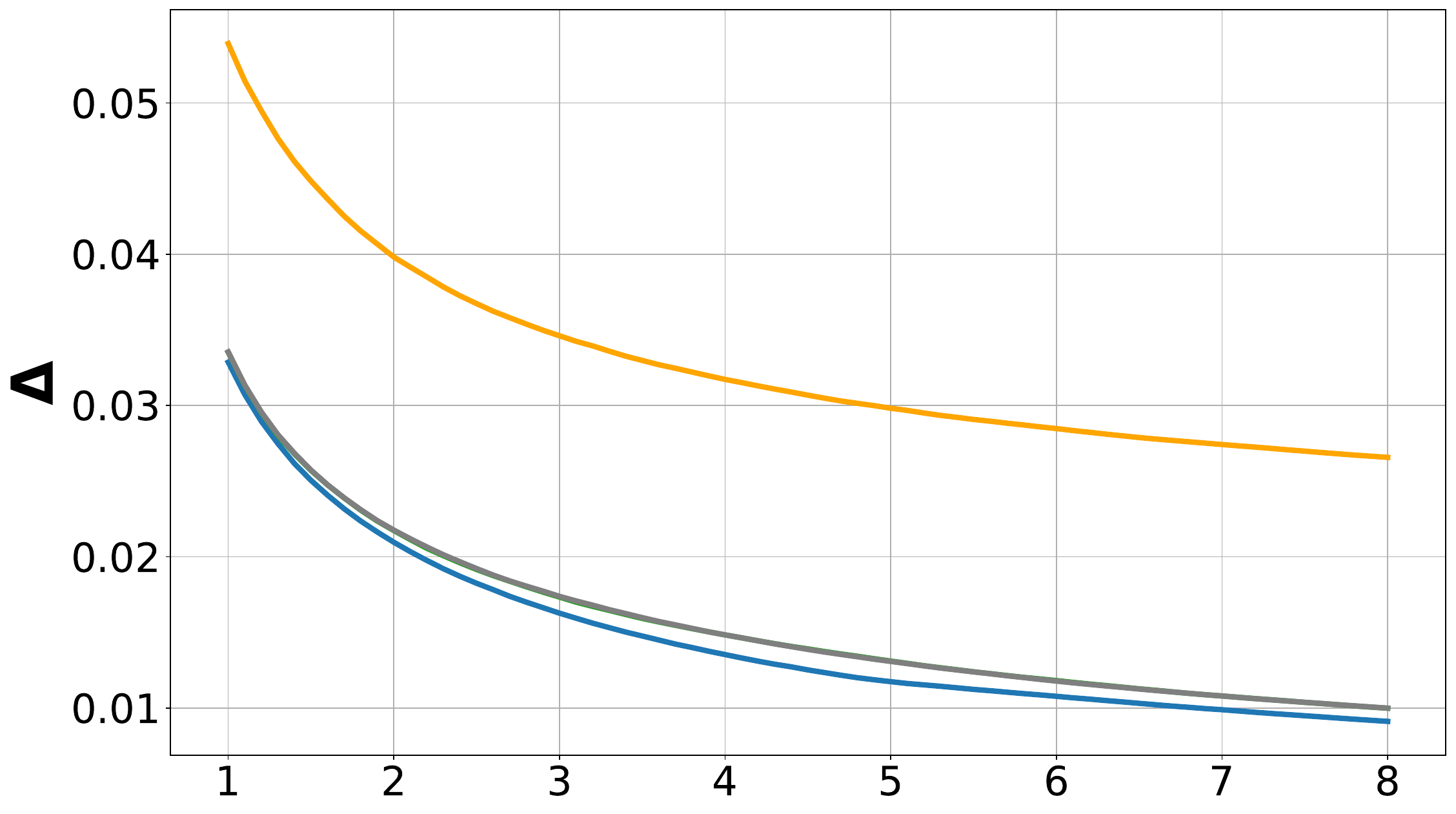}
    }
    \hfill
    \subfigure[$M=15,N=600$]{
    \label{fig:4ohy3}
        \includegraphics[width=0.31\textwidth]{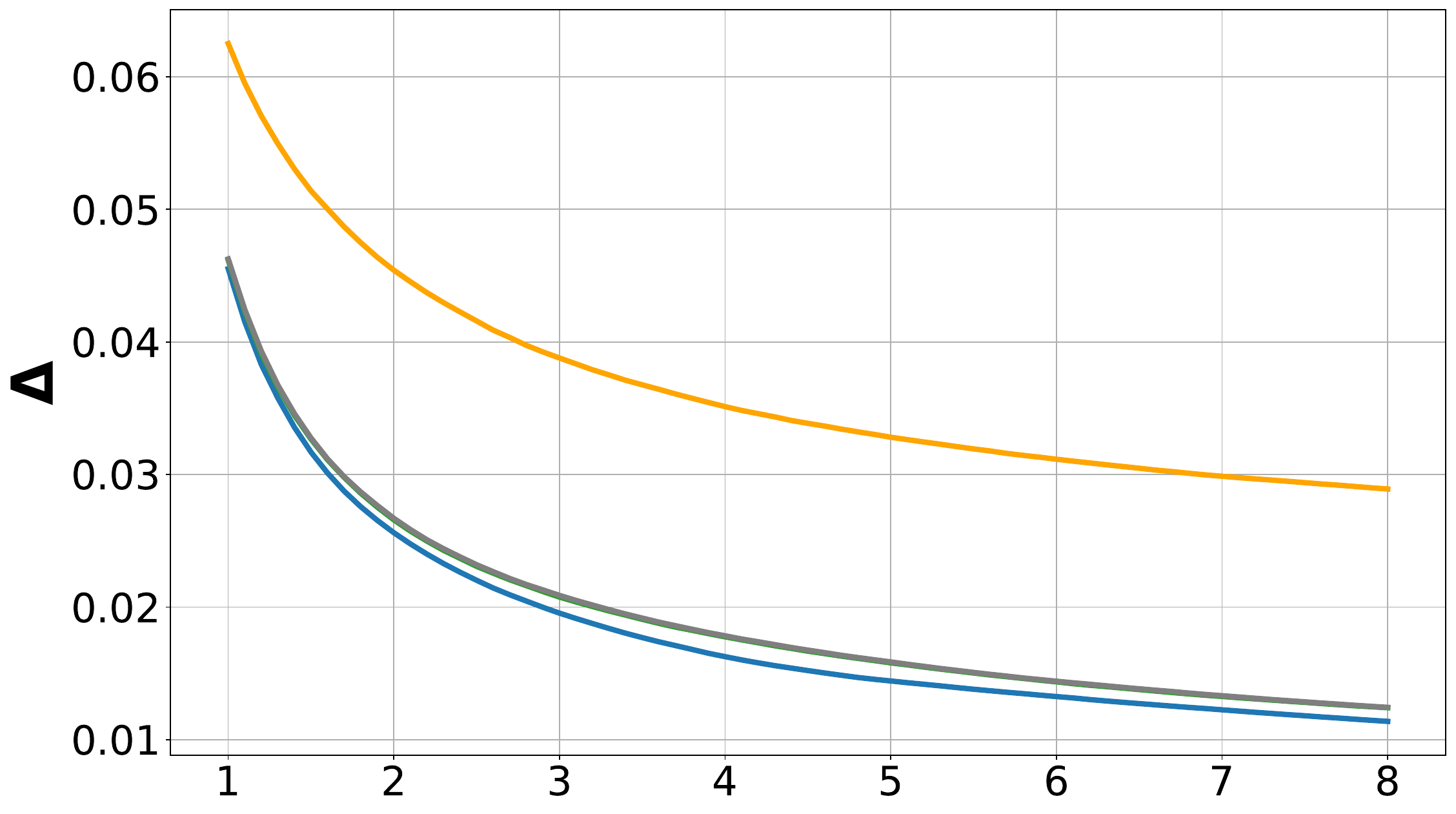}
    } \\
    \subfigure[$M=15,N=805$]{
    \label{fig:4ohy4}
        \includegraphics[width=0.31\textwidth]{figs/save-4o-pre_wins_theta.pdf}
    }
    \hfill
    \subfigure[$M=12,N=805$]{
    \label{fig:4ohy5}
        \includegraphics[width=0.31\textwidth]{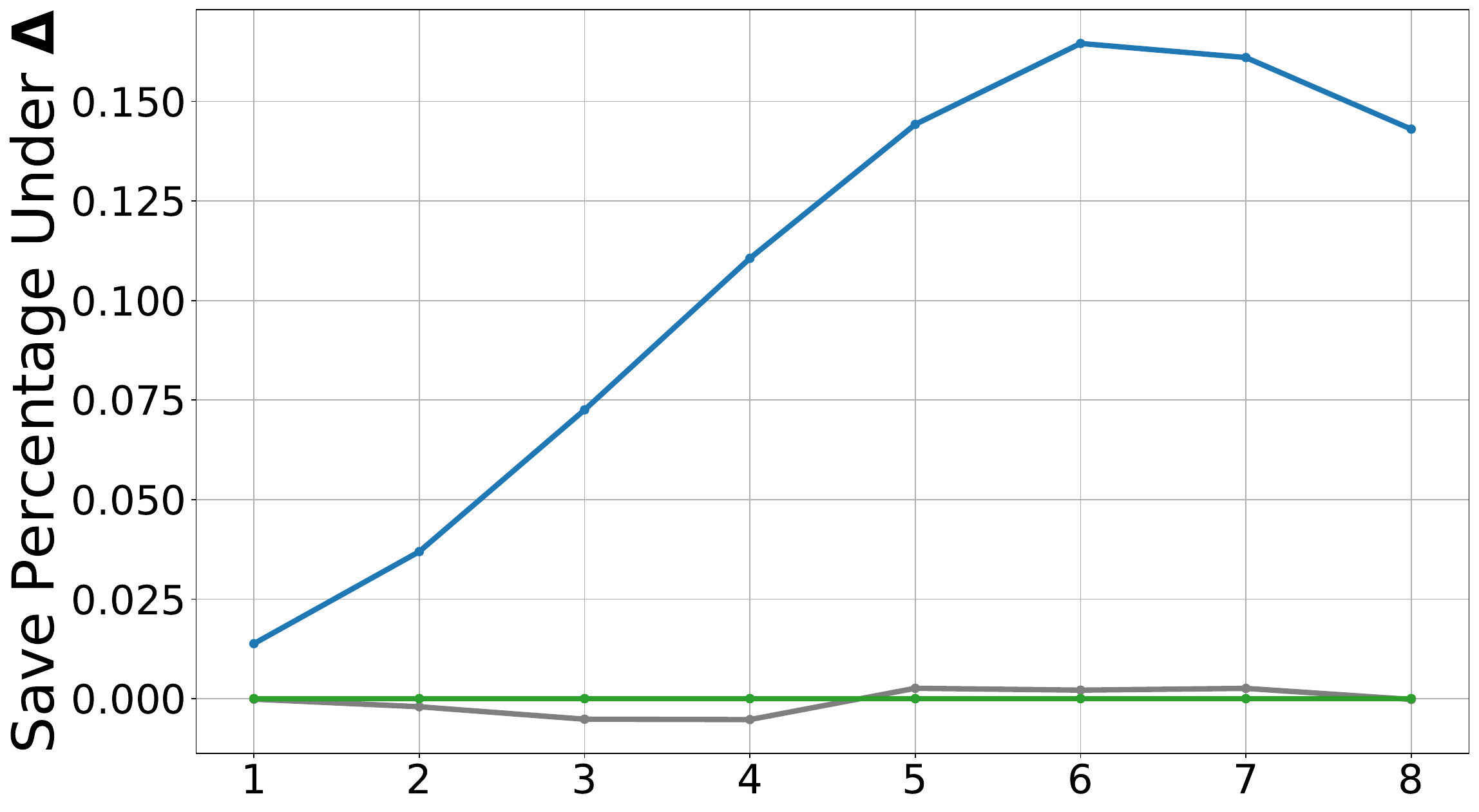}
    }
    \hfill
    \subfigure[$M=15,N=600$]{
    \label{fig:4ohy6}
        \includegraphics[width=0.31\textwidth]{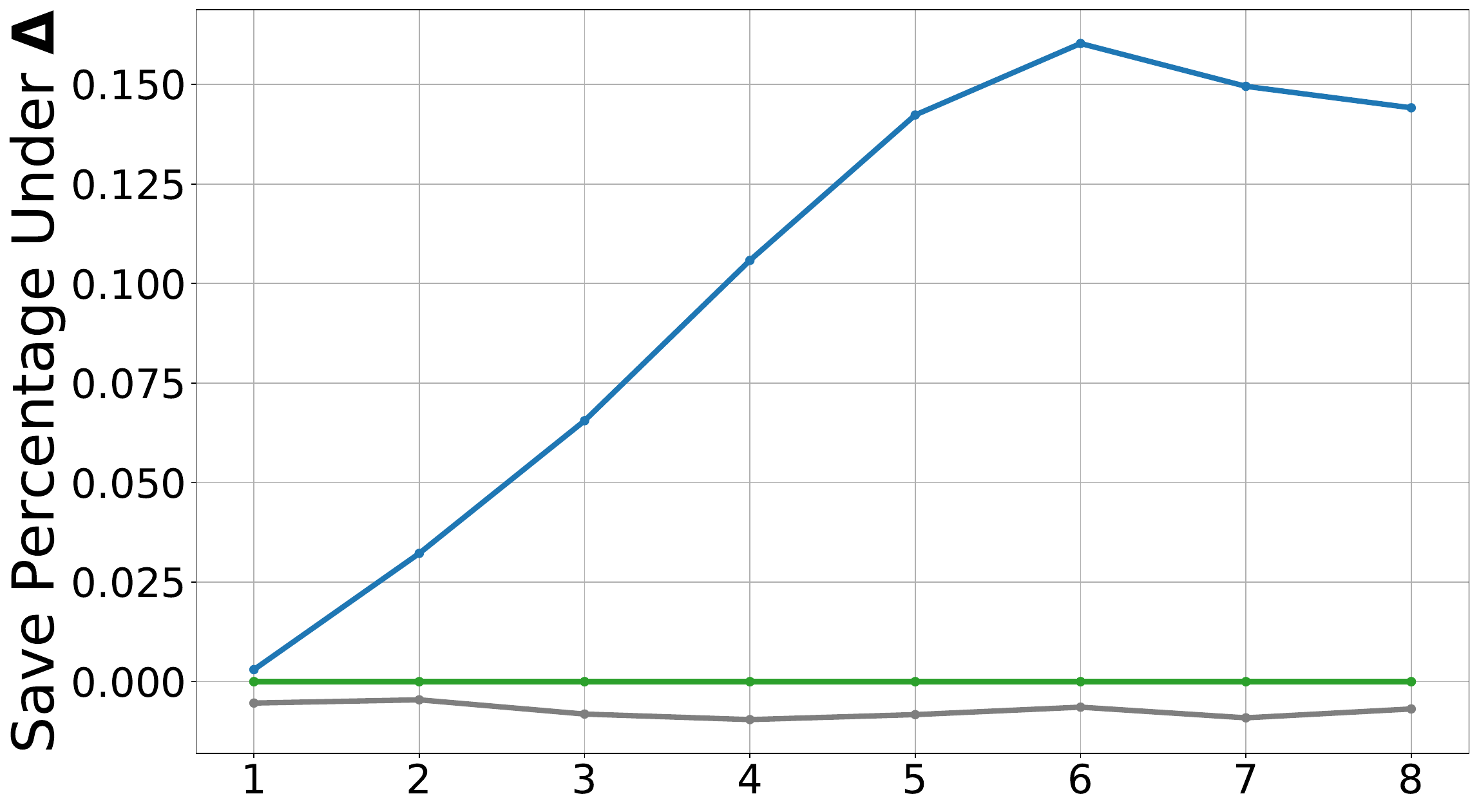}
    }
    \vspace{-0.2cm}
    \caption{Effects of number of models under evaluation $M$ and number of samples $N$. The results are obtained with GPT-4o as the judge on AlpacaEval.}
    \vspace{-0.4cm}
    \label{fig:main-4o-hyper}
\end{figure}

\begin{figure}[htbp]
    \centering
    \subfigure{
    \label{fig:4o31}
        \includegraphics[width=0.31\textwidth]{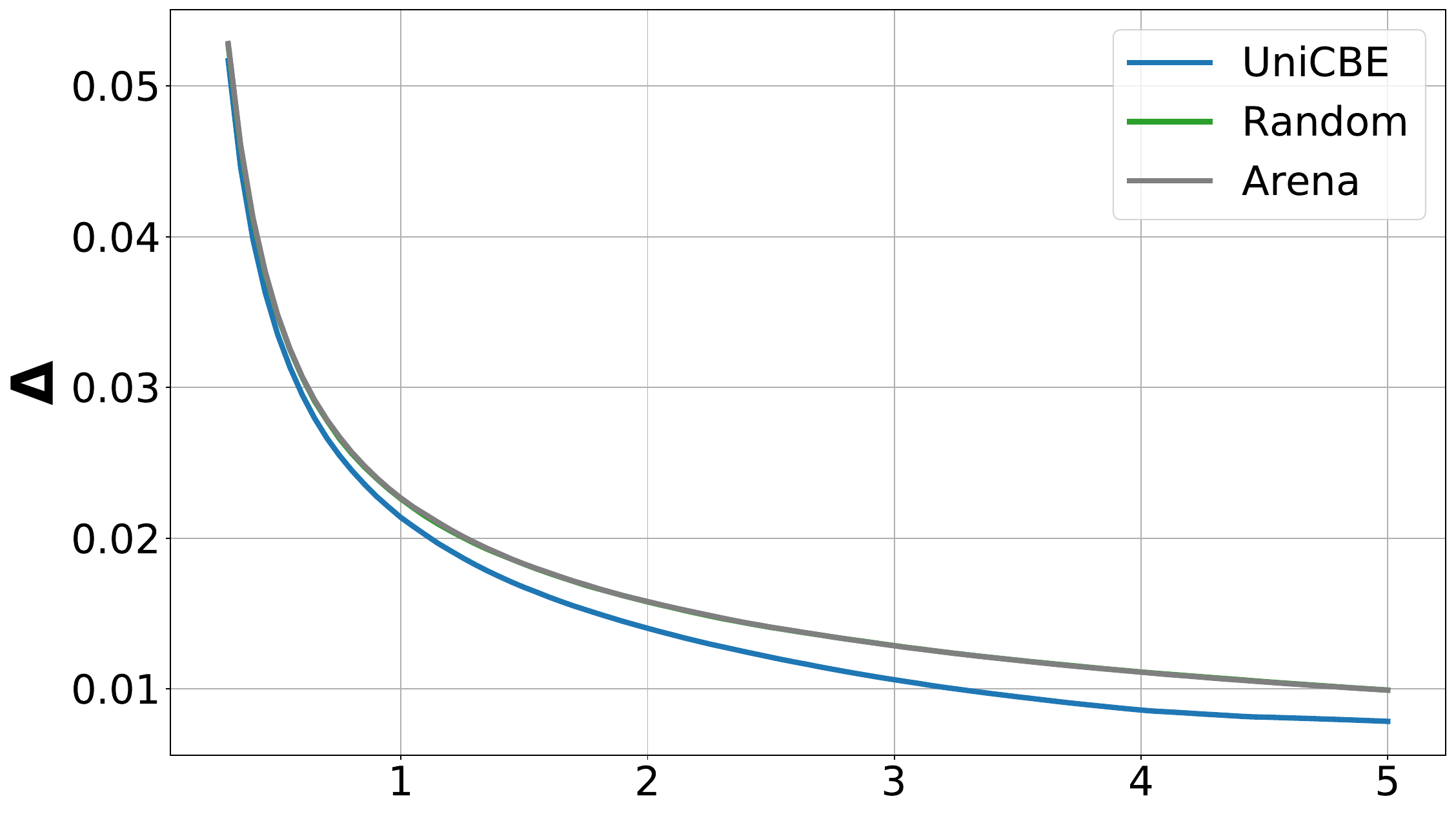}
    }
    \hfill
    \subfigure{
    \label{fig:4o32}
        \includegraphics[width=0.31\textwidth]{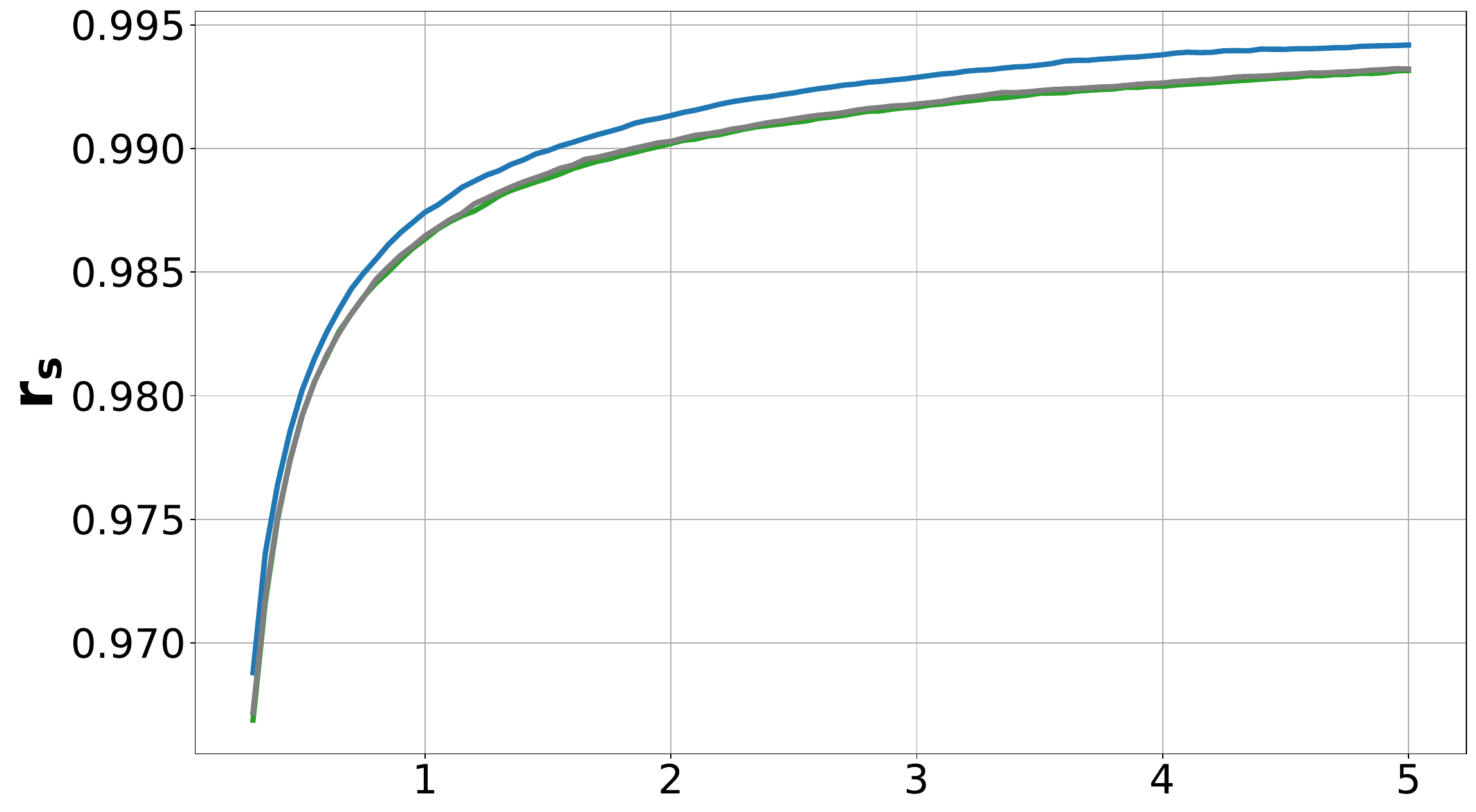}
    }
    \hfill
    \subfigure{
    \label{fig:4o33}
        \includegraphics[width=0.31\textwidth]{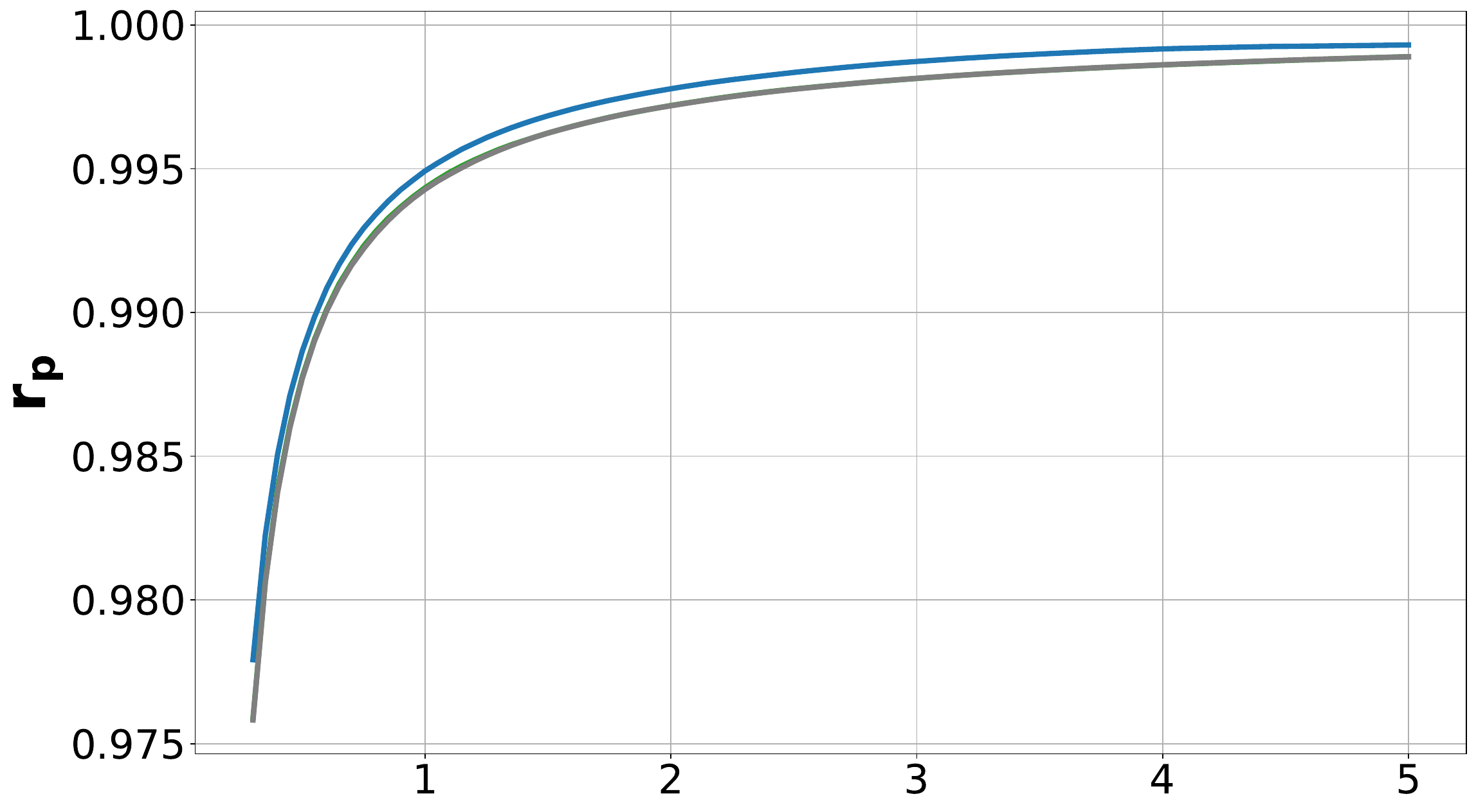}
    } \\
    \subfigure{
    \label{fig:4o34}
        \includegraphics[width=0.31\textwidth]{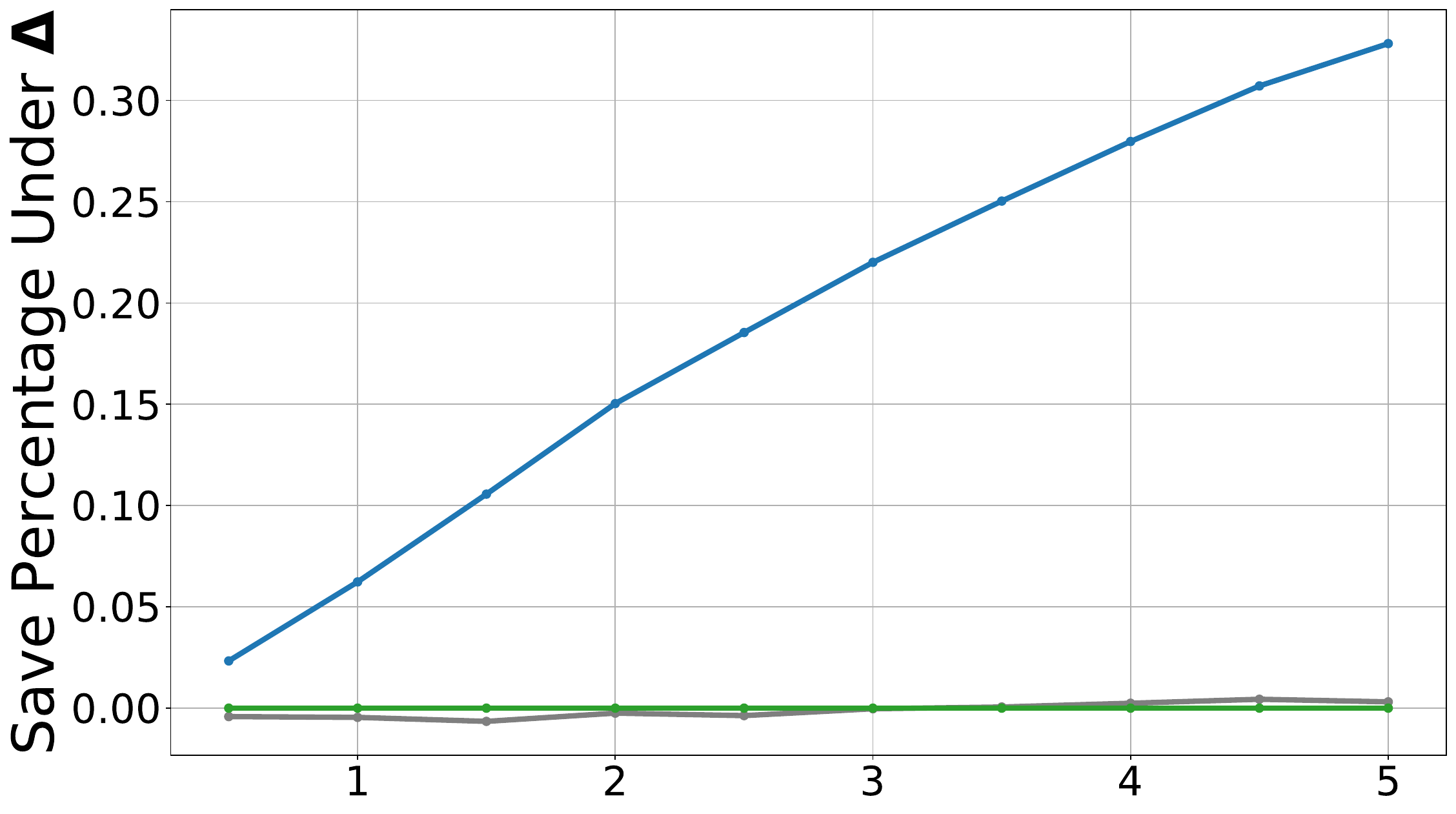}
    }
    \hfill
    \subfigure{
    \label{fig:4o35}
        \includegraphics[width=0.31\textwidth]{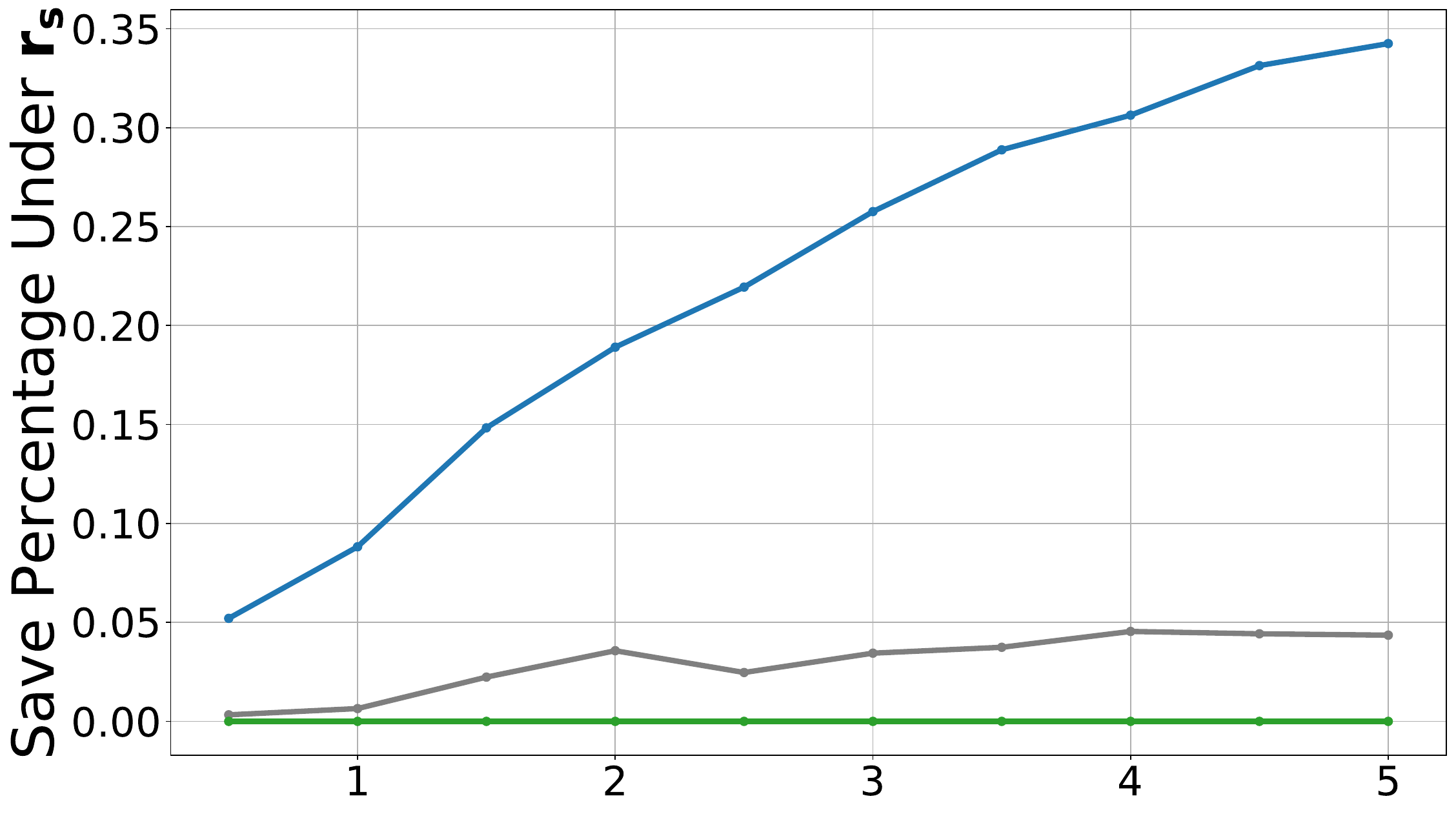}
    }
    \hfill
    \subfigure{
    \label{fig:4o36}
        \includegraphics[width=0.31\textwidth]{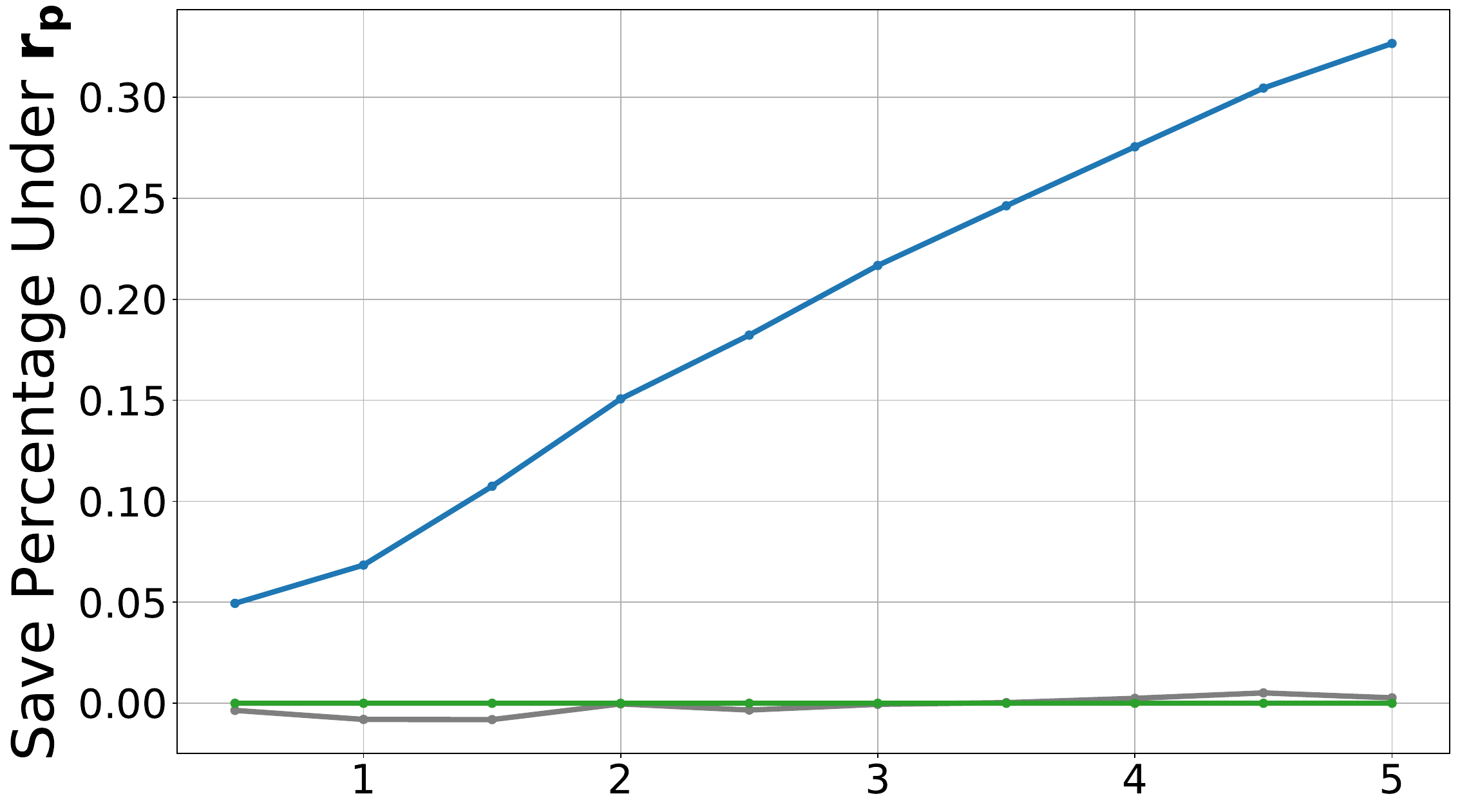}
    }
    \vspace{-0.2cm}
    \caption{Performance of CBE methods with list-wise preference of GPT-4o on AlpacaEval.}
    \vspace{-0.4cm}
    \label{fig:main-4o-3}
\end{figure}

\paragraph{Different Judges.} Apart from GPT-4o, we also experiment with GPT-3.5-turbo and Qwen-Plus as judge on AlpacaEval, and human as judge on MT-Bench. As shown in the above part of Figure~\ref{fig:main-diffjudge}, 
the overall conclusion with GPT-3.5-turbo is similar to GPT-4o, except for: (1) The performance of \textsc{Arena} no longer shows advantage over \textsc{Random}. (2) There is a certain decline in the performance of all methods, which is likely due to the increased noise in the preferences provided by GPT-3.5-turbo, leading to slower convergence. 
Similar trends are observed with Qwen-Plus (See Figure~\ref{fig:qwen}).
%在MT-Bench上的实验结果如图所示。
Results on MT-Bench are shown in the below part of Figure~\ref{fig:main-diffjudge}, where \textsc{UniCBE} also demonstrates better performance compared to other methods. However, due to the limited preference data included in MT-Bench, the experimental results show relatively larger fluctuations.
The results above demonstrate the good generalizability of \textsc{UniCBE} across different judges and the data domain.
% \textsc{UniCBE} 相比于其他方法同样呈现了更好的表现。不过由于MT-Bench所包含的preference数据较少，实验结果的波动也相对较大。
\paragraph{Varied Number of Models and Samples.}
%最后，如图所示，我们改变了模型数量$M$和sample 数量$N$进行实验. 可以看到\textsc{UniCBE}在这些settings下相比于各类baselines都取得了显著更好的效果，特别是在$M$和$N$较大时。
Finally, as shown in Figure~\ref{fig:main-4o-hyper}, we conducte experiments by varying the number of models \(M\) and samples \(N\). It can be observed that \textsc{UniCBE} achieves significantly better results compared to all the baselines under these settings, especially when \(M\) and \(N\) are large.
\paragraph{List-wise Preference.} 
\label{sec:5.4-2}
%\textsc{UniCBE}同样可以适用于list-wise的preference。假设judge每次给出$K$个模型的preference排序结果，我们可以类似~\eqref{eq:14}计算$K+1$维的P，并采样得到tuple((m^{l\text{-}1},...,m^{l\text{-}K},s^l)). 我们再由judge对该tuple的排序结果得到K(K-1)/2个pair-wise的preference。
%图3展示了K=3时各方法的结果，可以看到\textsc{UniCBE}在该setting下相对于\textsc{Random}的节省比例甚至超过30%。
\textsc{UniCBE} can also be applied to list-wise preference. Suppose the judge provides a preference ranking for $K$ models each time. We can compute a $K+1$ dimensional $P$, similar to equation~\eqref{eq:14}, and sample to obtain a tuple \((m^{l\text{-}1},...,m^{l\text{-}K},s^l)\). From the judge's ranking of this tuple, we derive \(\frac{K(K-1)}{2}\) pair-wise preferences.
Figure~\ref{fig:main-4o-3} shows the results for the case where \(K=3\). It can be seen that \textsc{UniCBE} achieves a savings compared to \textsc{Random} of over 30\% in this setting.
%这可能是因为List-wise Preference会使得\(\frac{K(K-1)}{2}\) pair-wise preferences集中在K个模型之间，加剧了采样bias，更需要\textsc{UniCBE}方法来抑制
This may be due to the fact that list-wise preference results in \(\frac{K(K-1)}{2}\) pair-wise preferences concentrated among the $K$ models, exacerbating the sampling bias. Therefore, \textsc{UniCBE} is even more needed to suppress this effect with list-wise preference.
%

%整体的结论和在GPT4-turbo作为judge时相似，除了：（1）Arena的表现相比Random不再有显著的优势。（2）各类方法的performance都有一定的下滑，这可能主要是因为GPT-3.5-turbo提供的preference中有更多的噪声导致了较慢的收敛。

\section{Conclusions}
%现有的CBE方法因其单一的优化目标而无法有效利用宝贵的prference signals. 我们深入分析了提升CBE的关键在于mitigate sampling bias，balance the escent process of uncertainty和抑制 updating uncertainty。在此基础上我们提出UniCBE方法，通过解耦地构建导向上述优化目标的采样矩阵并进行集成，从而获得了更好的accuracy，convergence和scalability. 综合的实验和分析验证了UniCBE的strong effectiveness，improved 可扩展性和良好的泛化性。
The existing comparing-based evaluation methods are ineffective in fully utilizing valuable preference signals due to their constrained optimization objectives. 
Our in-depth analysis reveals that the key to enhancing CBE lies in mitigating sampling bias, balancing the descent process of uncertainty, and suppressing the updating uncertainty.
Based on this, we propose the \textsc{UniCBE} framework that simultaneously optimizes the aforementioned objectives by promoting uniformity in corresponding aspects to enhance accuracy, convergence, and scalability.
Comprehensive experiments and analyses confirm the strong effectiveness, improved scalability, and good generalizability of \textsc{UniCBE}.

\clearpage
% This work is supported by Beijing Natural Science Foundation (No.4222037, L181010).

\bibliography{iclr2025_conference}
\bibliographystyle{iclr2025_conference}

\clearpage
\appendix
% \clearpage
% \appendix
\section{Proof of Theorem in \S\ref{sec:3.2}}
\label{app:proof}
Given that $\sum_{i=1}^U X_i = 0$, we want to attain sampling set $\mathcal{S}$ that satisfies $|\mathcal{S}|=V$ and:
\begin{equation}
    \mathcal{S} = argmin_{\mathcal{S}} | \sum\limits_{i \in \mathcal{S}} X_i | = argmin_{\mathcal{S}} (\sum\limits_{i \in \mathcal{S}} X_i )^2
    \label{eq-ap:1}
\end{equation}
Firstly, it is easy to know that for any sampling set $\mathcal{S}$:
\begin{equation}
    \mathbb{E}[ \sum\limits_{i \in \mathcal{S}} X_i ]=0
    \label{eq-ap:2}
\end{equation}
Thus, 
\begin{equation}
    \mathbb{E}[(\sum\limits_{i \in \mathcal{S}} X_i)^2] = 
    \mathbb{E}[(\sum\limits_{i \in \mathcal{S}} X_i-0)^2] =
    \mathbb{E}[(\sum\limits_{i \in \mathcal{S}} X_i-\mathbb{E}[ \sum\limits_{i \in \mathcal{S}} X_i ])^2] =\mathbb{E}[  \mathrm{Var}[ \sum\limits_{i \in \mathcal{S}} X_i ] ]
    \label{eq-ap:3}
\end{equation}
Considering that:
\begin{equation}
    \mathrm{Var}[ \sum\limits_{i \in \mathcal{S}} X_i ] =  \sum\limits_{i \in \mathrm{set}(\mathcal{S})} c_i^2\mathrm{Var}(X_i)
    \label{eq-ap:4}
\end{equation}
where $c_i$ denotes the number of $X_i$ in $\mathcal{S}$. On this basis, we derive that:
\begin{equation}
\begin{aligned}
    \mathbb{E}[(\sum\limits_{i \in \mathcal{S}} X_i)^2] &= \mathbb{E}[\mathrm{Var}(X)]\sum\limits_{i \in \mathrm{set}(\mathcal{S})} c_i^2\\
&\geq \mathbb{E}[\mathrm{Var}(X)](\sum\limits_{i \in \mathrm{set}(\mathcal{S})} c_i)^2 |\mathrm{set}(\mathcal{S})|^{-1}\\
&=V^2\mathbb{E}[\mathrm{Var}(X)] |\mathrm{set}(\mathcal{S})|^{-1} \\
&\geq V^2\mathbb{E}[\mathrm{Var}(X)]\mathrm{min}(U,V)^{-1}
\end{aligned}
\label{eq-ap:5}
\end{equation}
The equality condition of the first inequality is: the number of samples taken from each category is equal. The equality condition of the second inequality is: the number of sampled categories equals to $\mathrm{min}(U,V)$. These two conditions imply that a completely uniform sampling strategy is optimal.

\section{Introduction of Elo Rating System and Bradley-Terry Model}
\subsection{Elo Rating System}
\label{app:elo}
The Elo rating system \citep{elo} is widely used to rank participants based on their relative performance in competitive settings. Given two models, $A$ and $B$, with initial ratings $R_A$ and $R_B$, the expected score of model $A$ in a pairwise comparison is calculated as:
\[
E_A = \frac{1}{1 + 10^{(R_B - R_A)/400}}
\]
Similarly, the expected score for model $B$ is:
\[
E_B = \frac{1}{1 + 10^{(R_A - R_B)/400}}
\]
After the comparison, the actual results are used to update the ratings. If model $A$ wins, its new rating $R_A'$ is updated as:
\[
R_A' = R_A + K(S_A - E_A)
\]
where $S_A$ is the actual result of the match (1 for a win, 0 for a loss, and 0.5 for a draw), and $K$ is a constant that controls the sensitivity of the rating adjustment. Model $B$'s rating is updated in a similar way:
\[
R_B' = R_B + K(S_B - E_B)
\]
where $S_B$ is the actual result for model $B$.

When extending the Elo rating system to multiple models, we consider a set of $n$ models. Pairwise comparisons between the models are conducted, resulting in $\binom{n}{2}$ unique pairs:
\[
\binom{n}{2} = \frac{n(n-1)}{2}
\]
Each pair is evaluated using the Elo score update rules, and the results are iteratively applied to adjust the ratings, ensuring that each model’s rating reflects its relative performance within the set.

The extension to multiple models leverages the transitive property. For any three models $i, j, k \in \{1, 2, \dots, n\}$, if $R_i > R_j$ and $R_j > R_k$, the transitivity implies:
\[
R_i > R_j \quad \text{and} \quad R_j > R_k \quad \implies \quad R_i > R_k
\]
This property ensures consistency in the rankings, even when individual match outcomes vary. By iterating over all $\binom{n}{2}$ comparisons, the Elo scores converge to reflect the overall capabilities of the models, with higher scores indicating stronger performance.

\subsection{Bradley-Terry Model}
\label{app:bt}
The Bradley-Terry model \citep{BT} estimates the probability that one model outperforms another in pairwise comparisons. For two models $M_i$ and $M_j$ with strength parameters $\xi_i$ and $\xi_j$, the probability that model $i$ beats model $j$ is modeled as:
\begin{equation}
\label{eq:bt}
P(M_i>M_j) = \frac{1}{1 + e^{\xi_j - \xi_i}}    
\end{equation}
where $\xi$ is an $|M|$-length vector of Bradley-Terry coefficients. Given a set of comparing results $\mathcal{S}=\{(M_i^t,M_j^t,R^t)\}_{t=1}^{T}$ where $R^t$ represents the degree $M_i^t$ wins over $M_j^t$. We set $\rm{mean}(\xi) = 0$.
After attaining the BT scores using $f^{pa}$, we calculate the estimated win rate matrix with~\eqref{eq:bt}.

\section{More Experimental Analyses}
\begin{figure}[h]
    \centering
    \subfigure{
    \label{fig:351m}
        \includegraphics[width=0.31\textwidth]{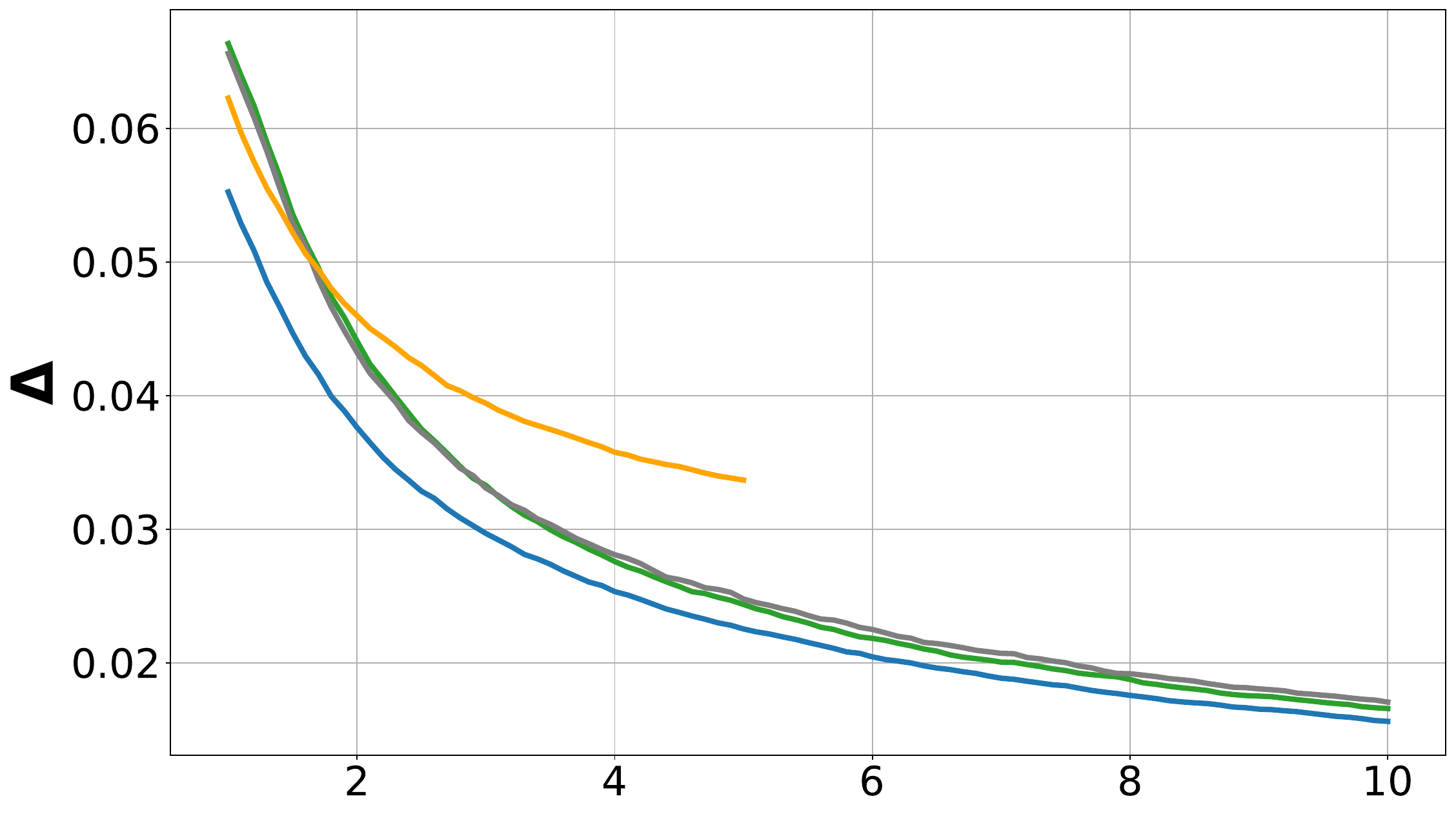}}
    \hfill
    \subfigure{
    \label{fig:355m}
        \includegraphics[width=0.31\textwidth]{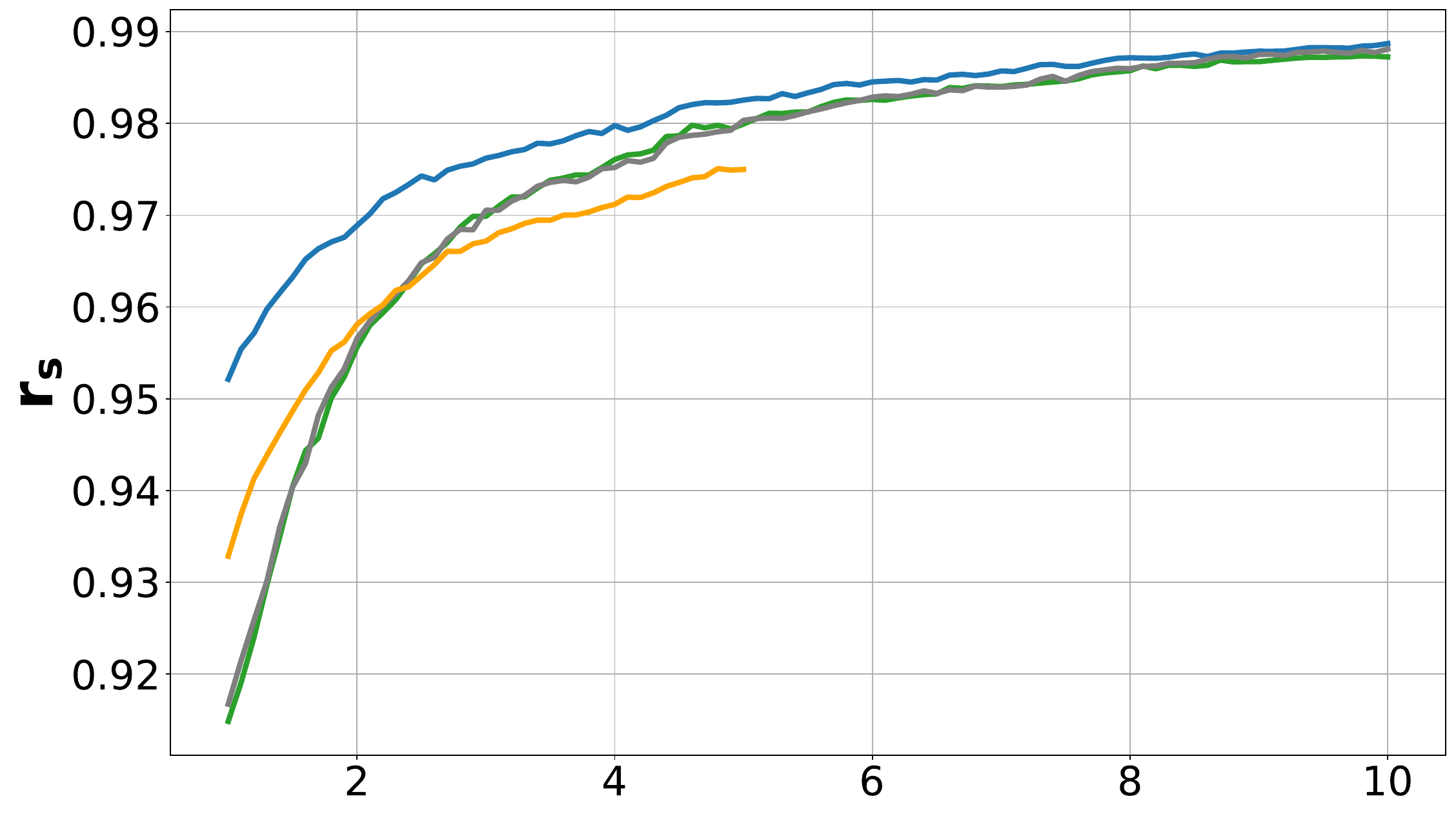}
    }
    \hfill
    \subfigure{
    \label{fig:356m}
        \includegraphics[width=0.31\textwidth]{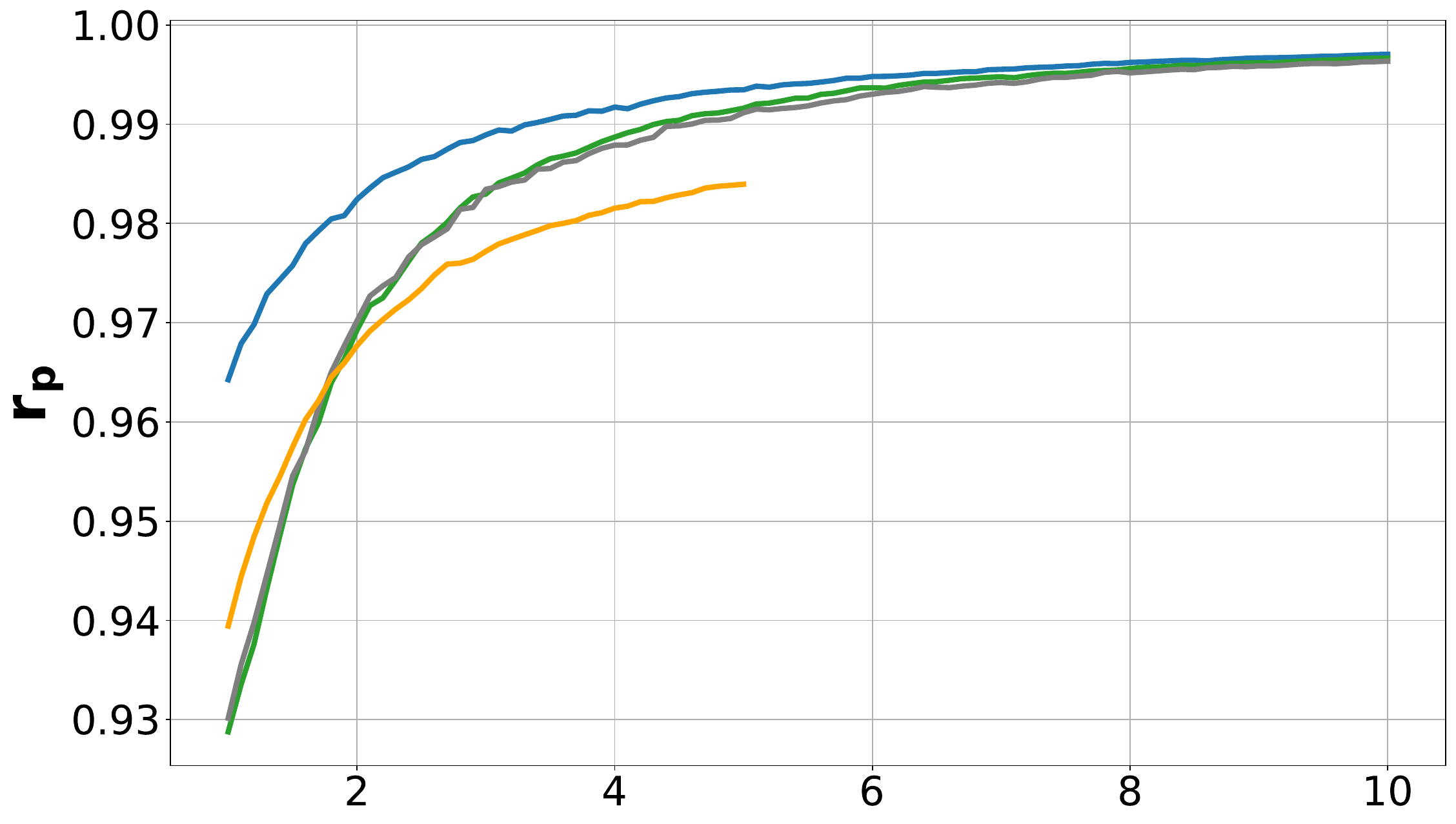}
    } \vspace{-0.2cm}\\
    \subfigure{
    \label{fig:mt1m}
        \includegraphics[width=0.31\textwidth]{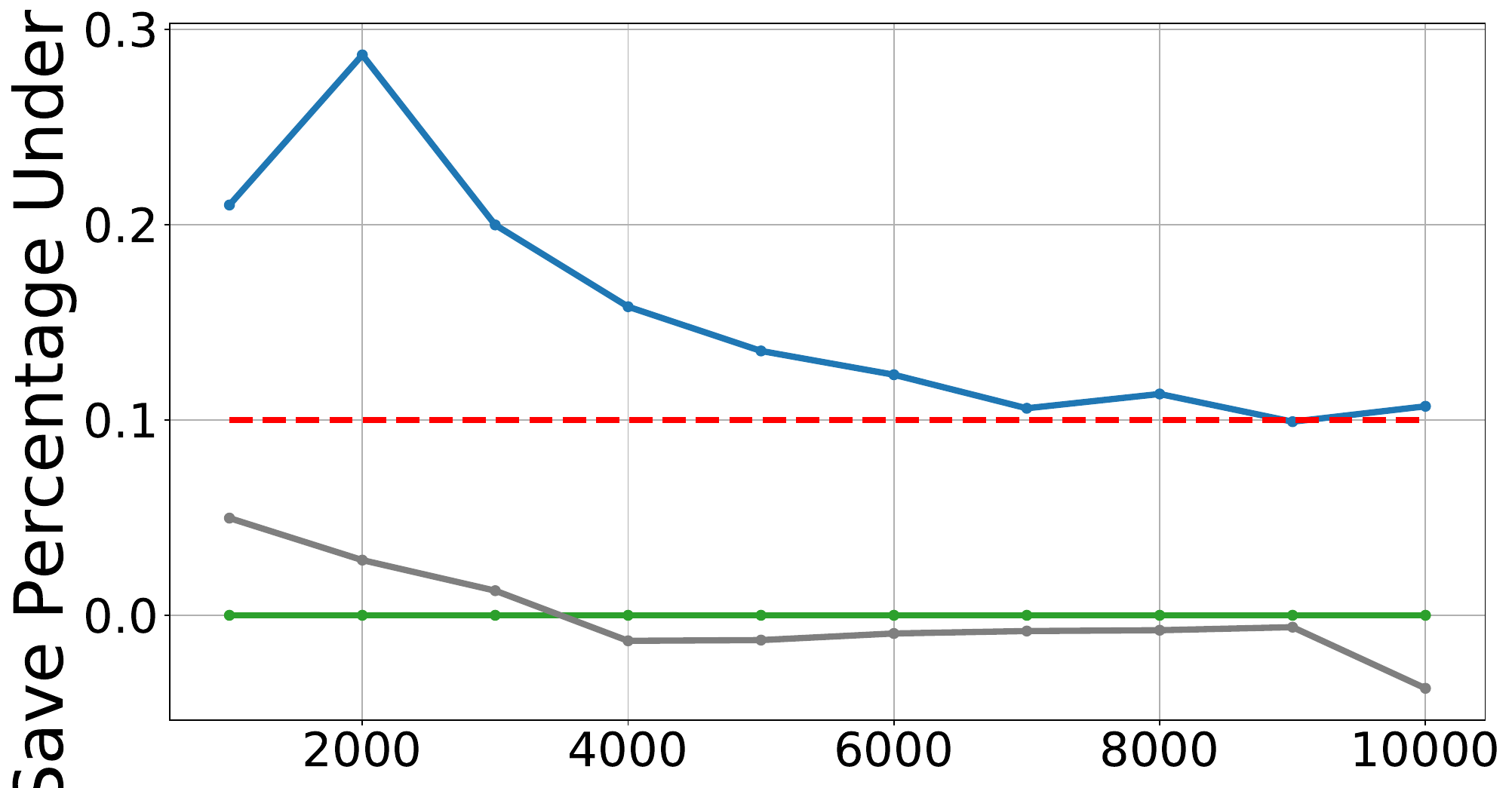}
    }
    \hfill
    \subfigure{
    \label{fig:mt5m}
        \includegraphics[width=0.31\textwidth]{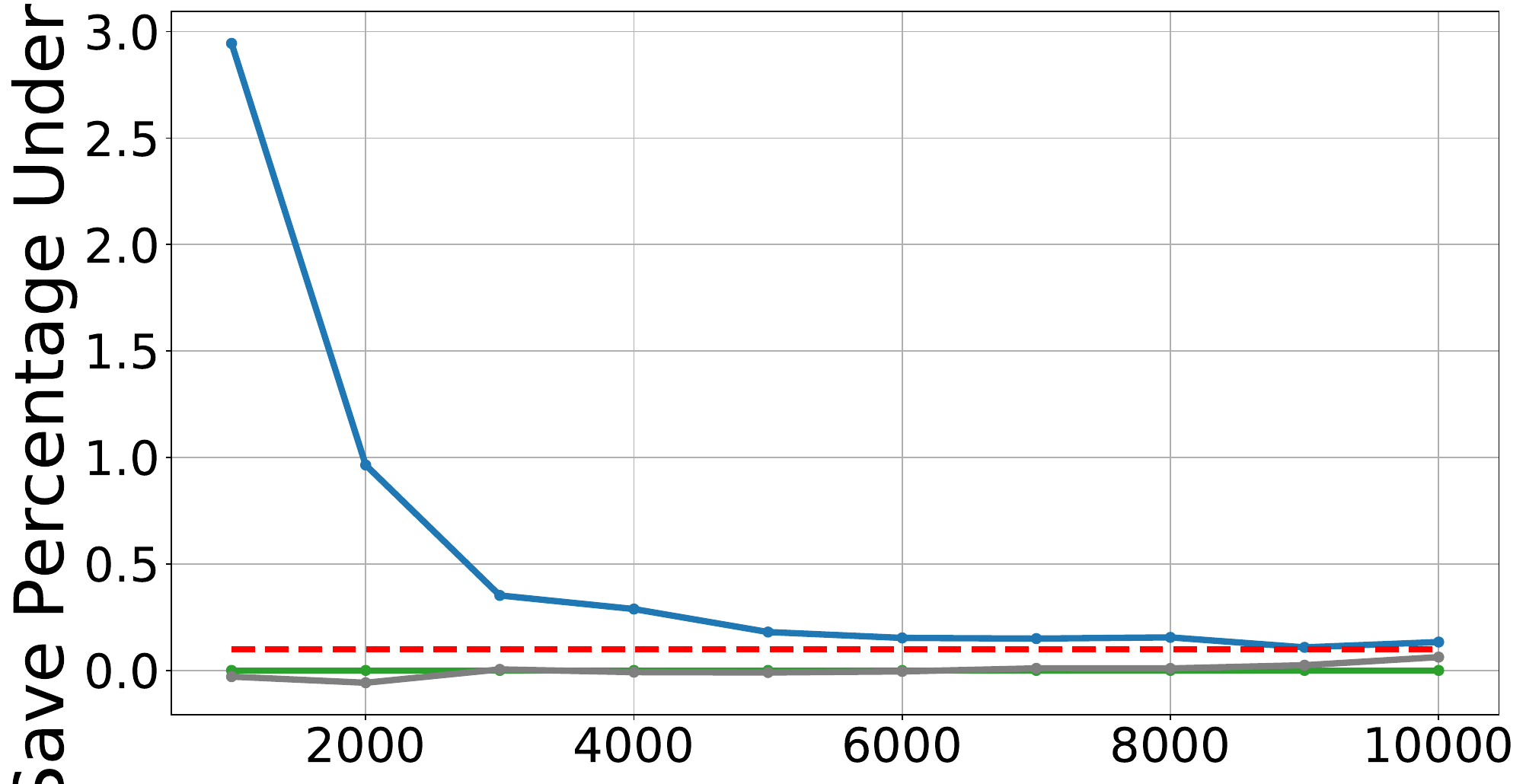}
    }
    \hfill
    \subfigure{
    \label{fig:mt6m}
        \includegraphics[width=0.31\textwidth]{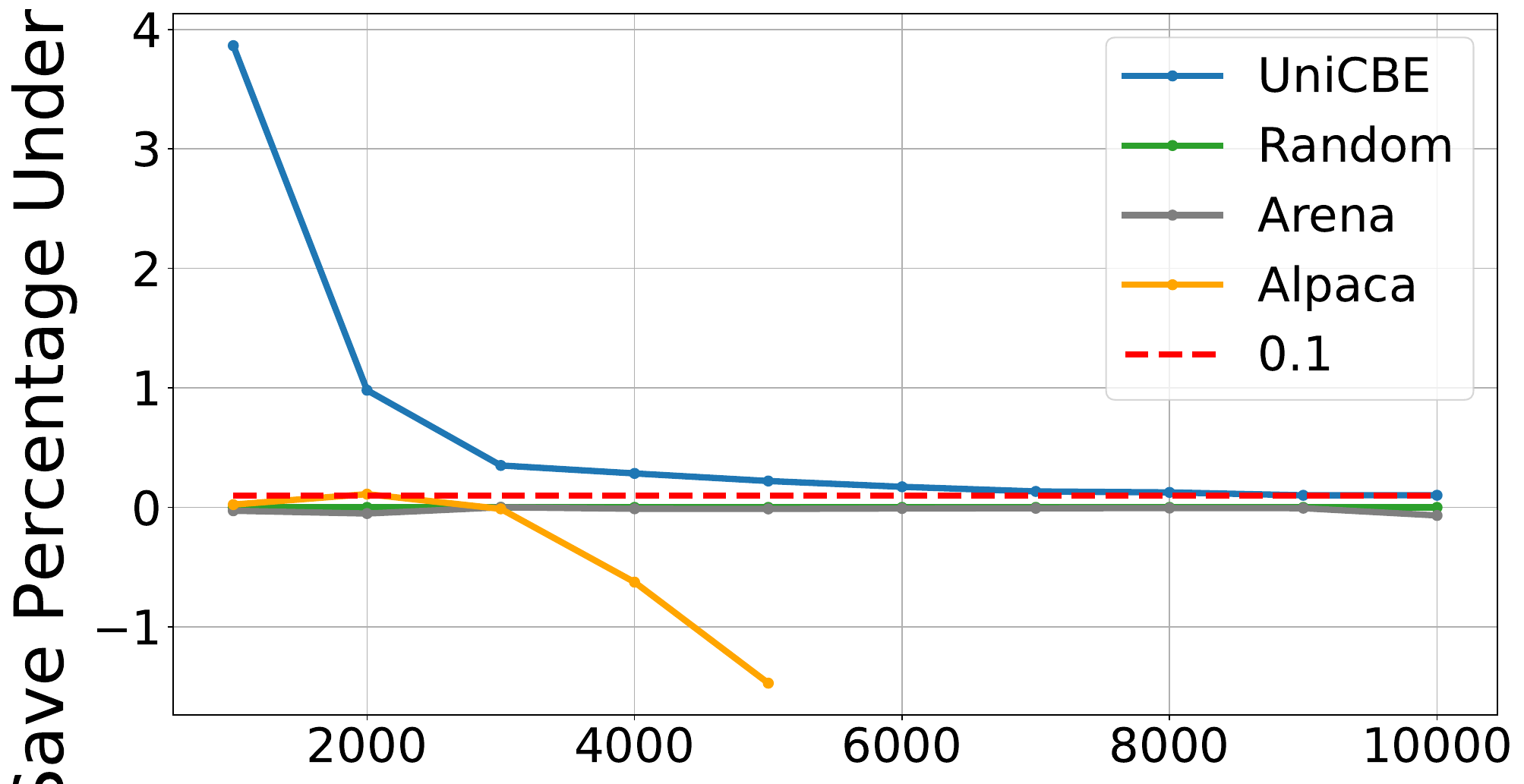}
    }
    \vspace{-0.2cm}
    \caption{Results of compared CBE methods in a scenario where models and samples are dynamically added or removed at a random frequency. }
    \vspace{-0.3cm}
    \label{fig:dynamic}
\end{figure}
\subsection{Performance in Scenarios Close to Reality}
We think that conducting experiments in settings that are closer to real-world scenarios (highly dynamic and requiring real-time evaluation) can help us more comprehensively assess \textsc{UniCBE} and the baseline methods. To this end, we perform the following experiments: Starting with a sample size of $N=600$ and model number of $M=12$, we execute a random operation at each time step. The operations included: adding one model to be evaluated with a probability of 0.01, removing one model with a probability of 0.01, adding one potential sample with a probability of 0.01, randomly deleting one sample with a probability of 0.01, and taking no action with a probability of 0.96. Based on the experimental results shown in Figure~\ref{fig:dynamic}, we have the following observations:
\begin{itemize}[leftmargin=20pt]
\setlength{\itemsep}{0pt}
\setlength{\parsep}{0pt}
\setlength{\parskip}{0pt}
\item The convergence speed of all baseline methods significantly slowed down compared to Figure~\ref{fig:main-4o}. None of the baseline methods achieve a Spearman correlation coefficient of 0.96 or a Pearson correlation coefficient of 0.97 by $T=2000$, highlighting the difficulty of model evaluation in this setting. In contrast, \textsc{UniCBE} achieve rapid convergence, reaching a Spearman coefficient of approximately 0.97 and a Pearson coefficient exceeding 0.98 by $T=2000$.
\item Over the long term, as $T$ increases, \textsc{UniCBE} consistently demonstrates over 10\% savings in preference budget across all metrics, even under this challenging setting, showcasing its strong practicality.
\item An interesting observation is that \textsc{AlpacaEval} exhibits better convergence in the early stages compared to \textsc{Random} and \textsc{Arena}, supporting our previous conclusions in Table~\ref{tb:intro}. However, as $T$ increases, \textsc{AlpacaEval}'s lack of accuracy optimization objective leads to its performance being surpassed by \textsc{Random} and \textsc{Arena}.
\end{itemize} 

\subsection{Ablation Study of Uniformity Constraints}
\begin{figure}[h]
    \centering
    \subfigure{
    \label{fig:351m}
        \includegraphics[width=0.31\textwidth]{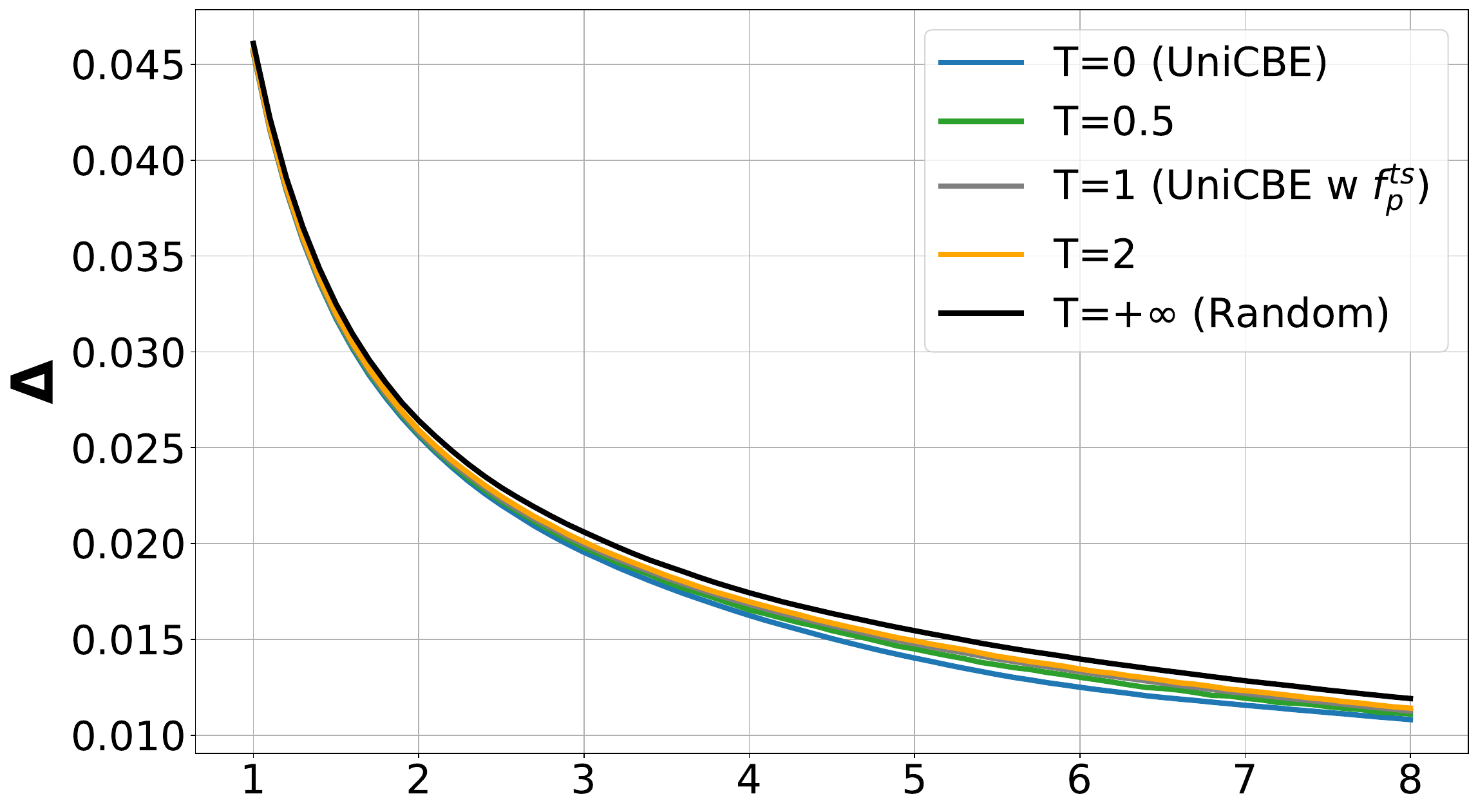}}
    \hfill
    \subfigure{
    \label{fig:355m}
        \includegraphics[width=0.31\textwidth]{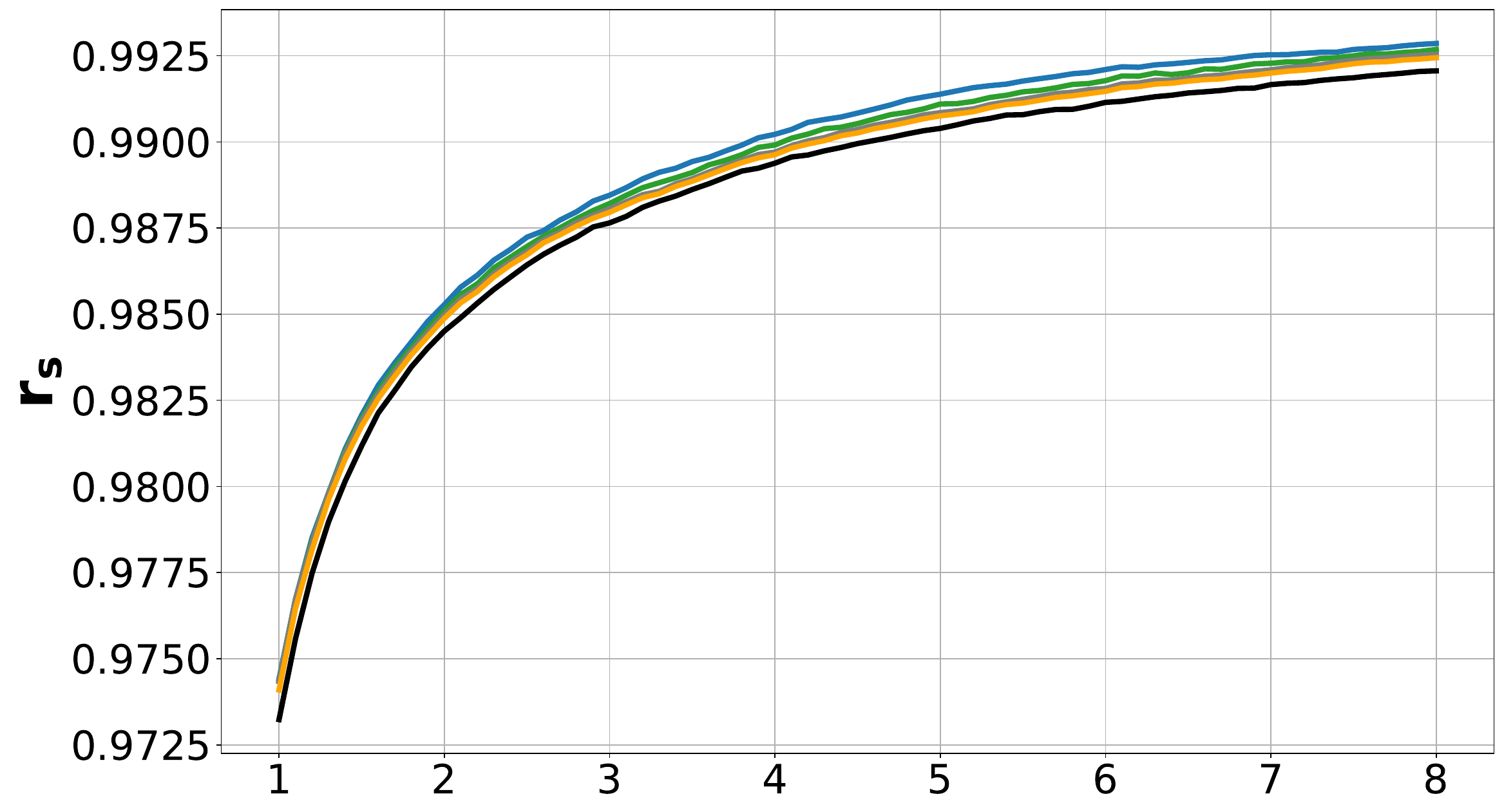}
    }
    \hfill
    \subfigure{
    \label{fig:356m}
        \includegraphics[width=0.31\textwidth]{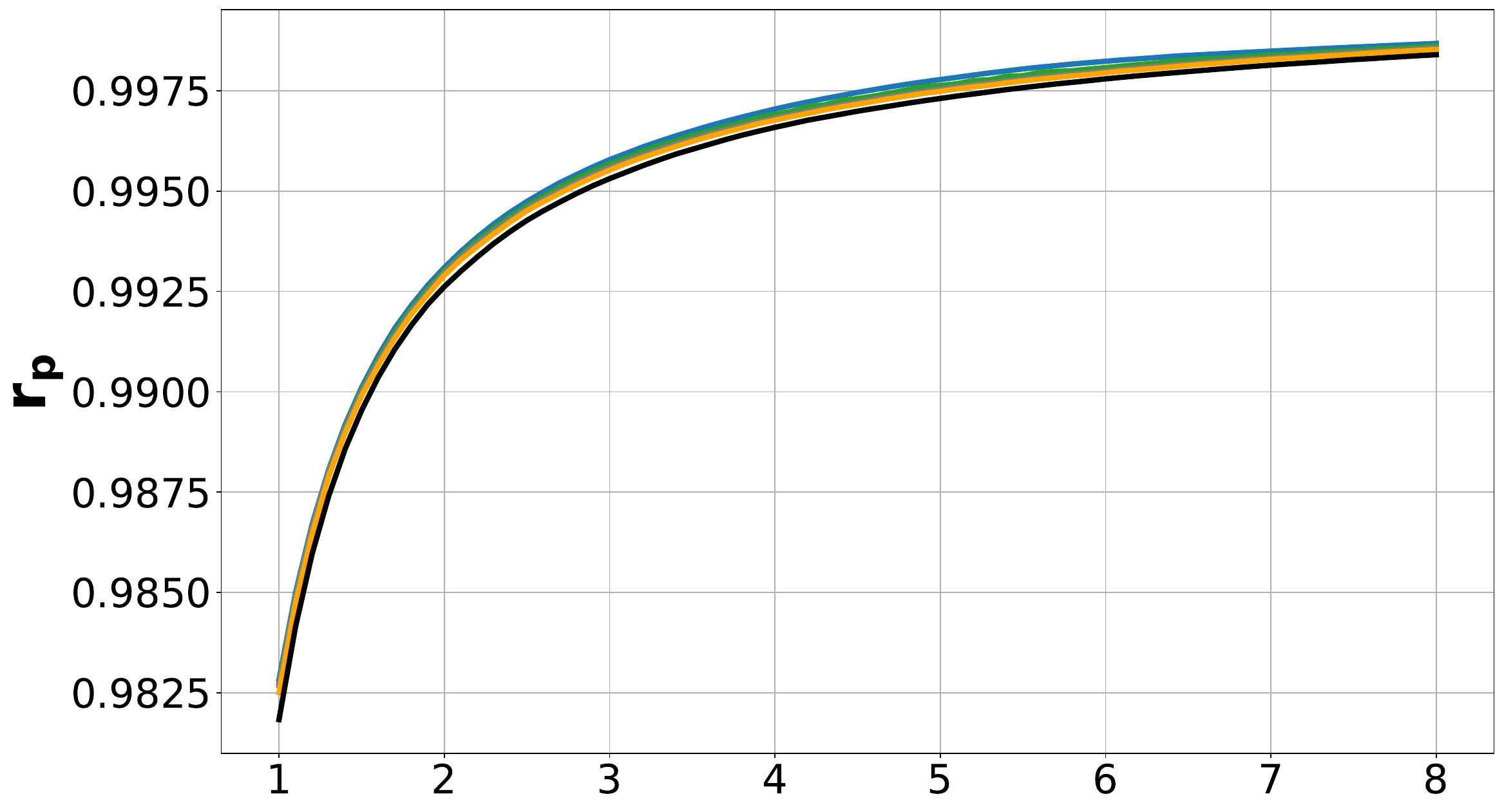}
    } \vspace{-0.2cm}\\
    \subfigure{
    \label{fig:mt1m}
        \includegraphics[width=0.31\textwidth]{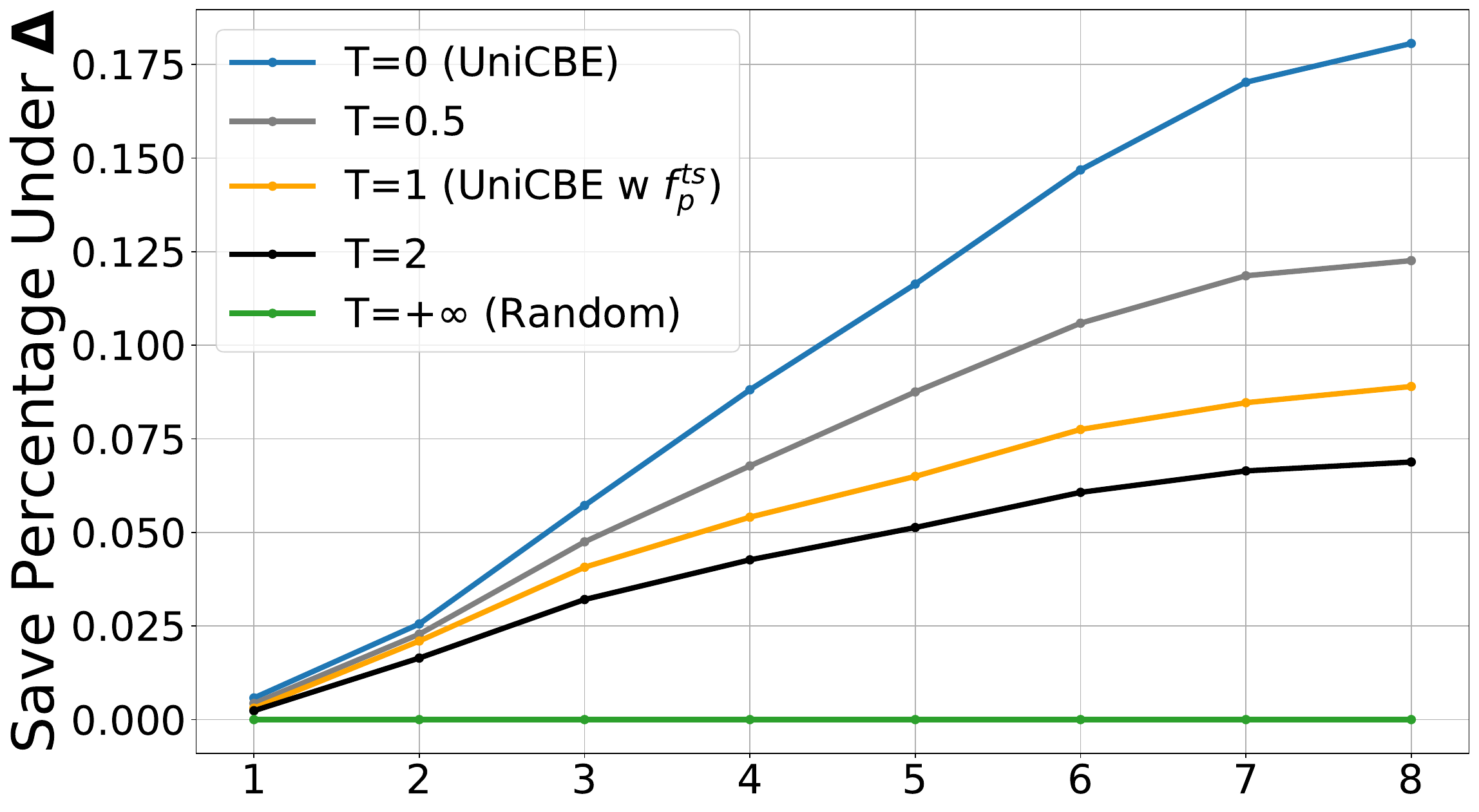}
    }
    \hfill
    \subfigure{
    \label{fig:mt5m}
        \includegraphics[width=0.31\textwidth]{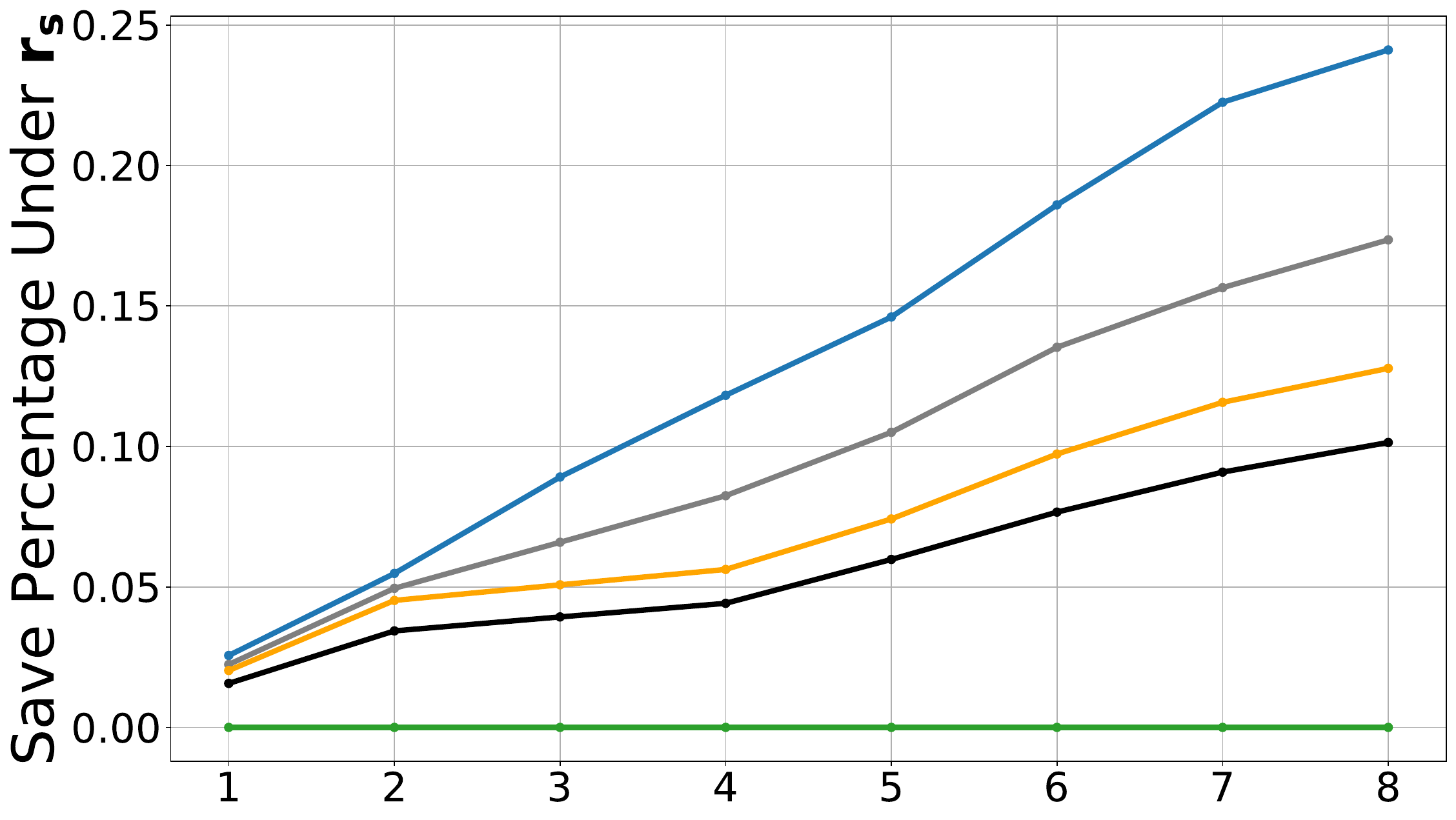}
    }
    \hfill
    \subfigure{
    \label{fig:mt6m}
        \includegraphics[width=0.31\textwidth]{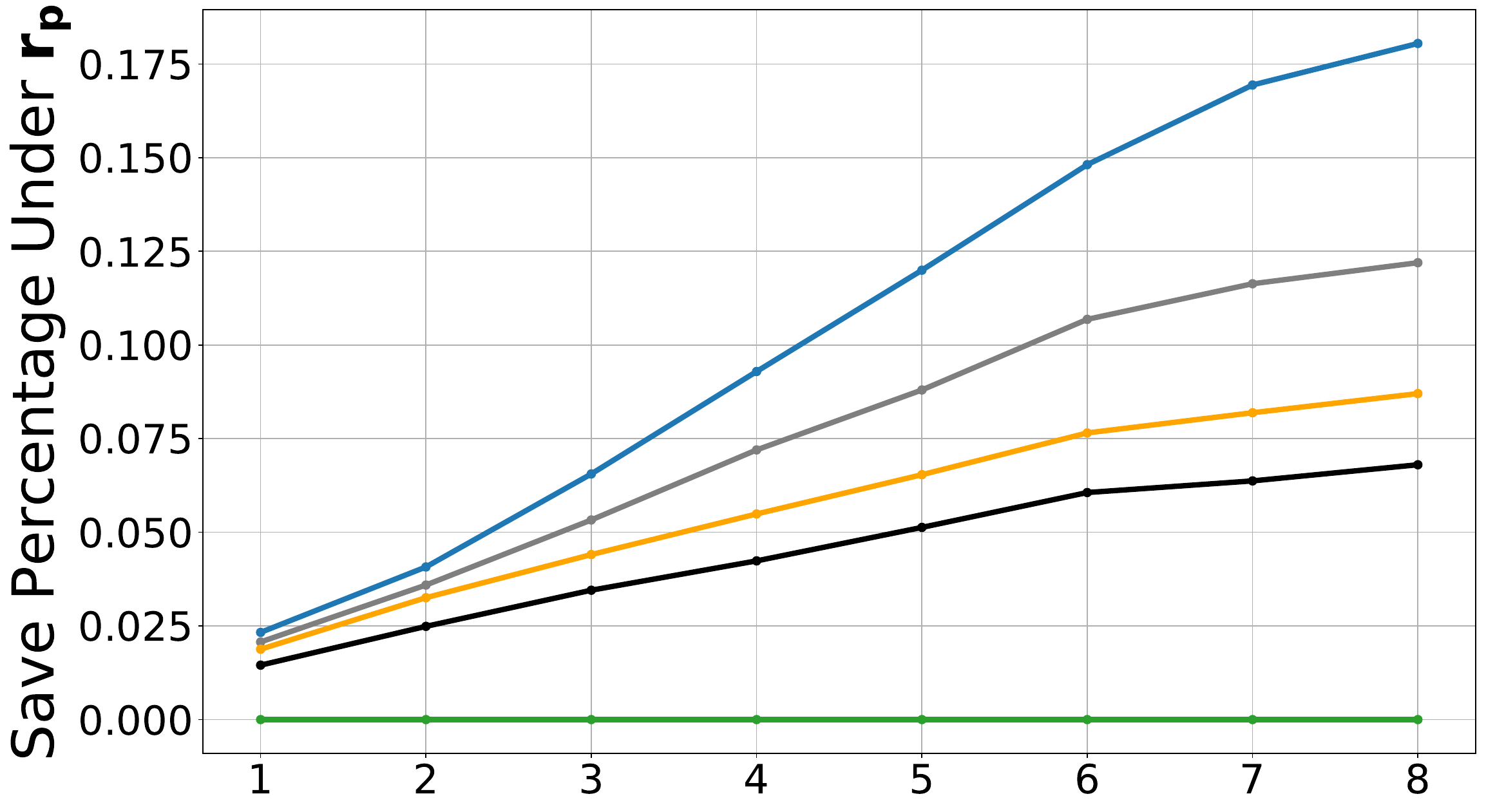}
    }
    \vspace{-0.2cm}
    \caption{Results of \textsc{UniCBE} with different sampling temperatures. }
    \vspace{-0.4cm}
    \label{fig:T}
\end{figure}

Based on our analysis in \S\ref{sec:3}, the degree to which uniformity is achieved is positively correlated with performance in terms of accuracy, convergence, and scalability. To explore the empirical relationship between the degree of uniformity constraints and the final outcomes, we draw inspiration from the concept of temperature-based control in random sampling. By adjusting the temperature $T$ in the following formula for sampling $f^{ts}_T$, we regulate the extent of uniformity constraints according to $P^l$ in ~\eqref{eq:14}:
\begin{equation}
    f^{ts}_T(i,j,k) =\frac{(P^l_{i,j,k})^{-T}}{\sum (P^{l})^{-T}}
\end{equation}
As \(T\) increases, the uniformity constraints become more relaxed. When \(T=0\), it corresponds to greedy sampling \(f^{ts}_g\), which imposes the strictest uniformity constraints. When \(T=1\), it corresponds to probabilistic sampling \(f^{ts}_p\), which imposes general uniformity constraints. When \(T=+\infty\), it corresponds to random sampling, where no uniformity constraints are applied. 
Our experimental results are shown in Figure~\ref{fig:T}. 
As \(T\) increases from 0 to \(T=+\infty\), the evaluation results progressively deteriorate. This indicates that adopting greedy sampling to impose the strictest uniformity constraints yields the optimal evaluation performance. This observation also validates the correctness of our conclusions in \S\ref{sec:3}.

\subsection{Adjusting the Weights of Optimization Objectives}
\begin{figure}[h]
    \centering
    \subfigure{
    \label{fig:351m}
        \includegraphics[width=0.31\textwidth]{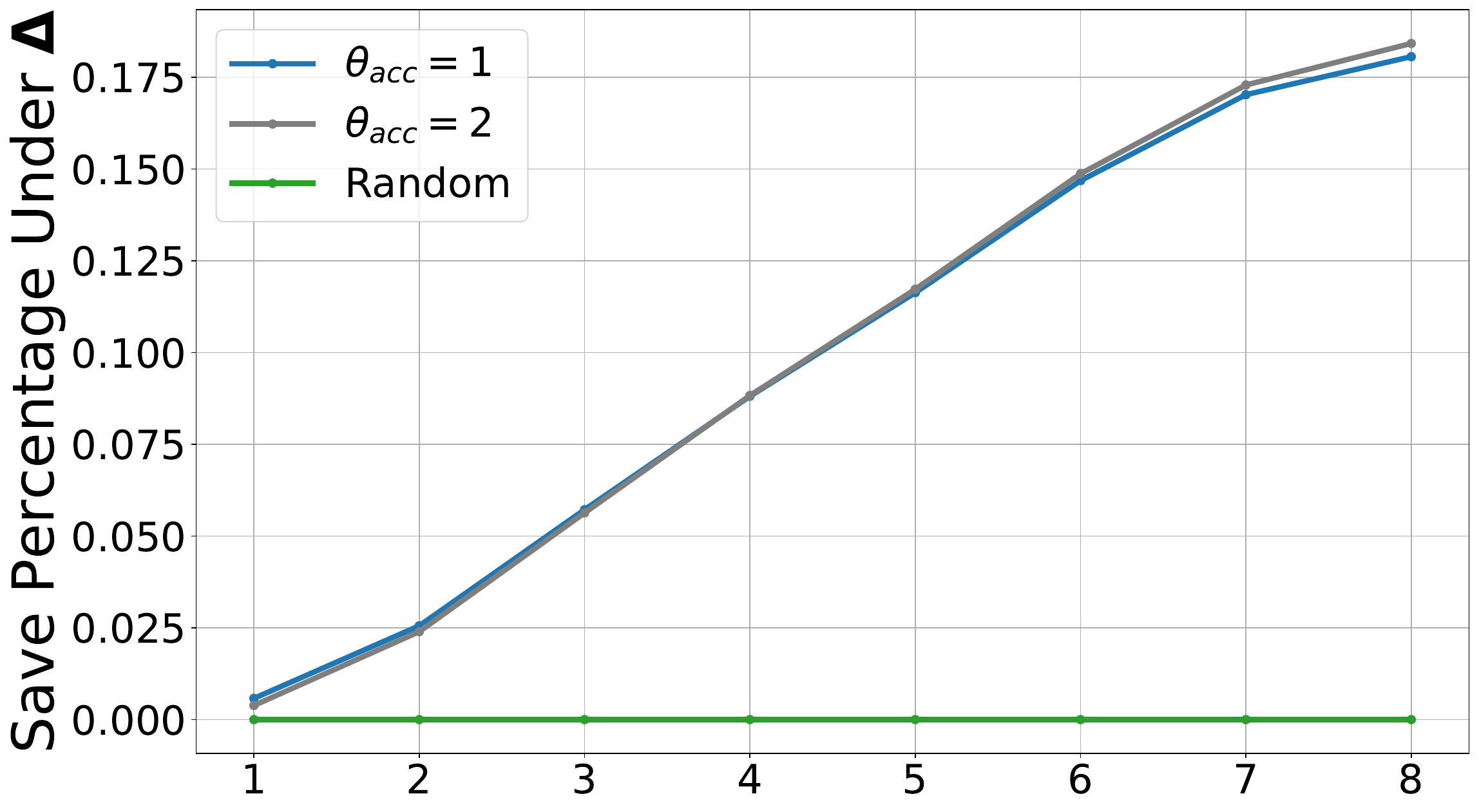}}
    \hfill
    \subfigure{
    \label{fig:355m}
        \includegraphics[width=0.31\textwidth]{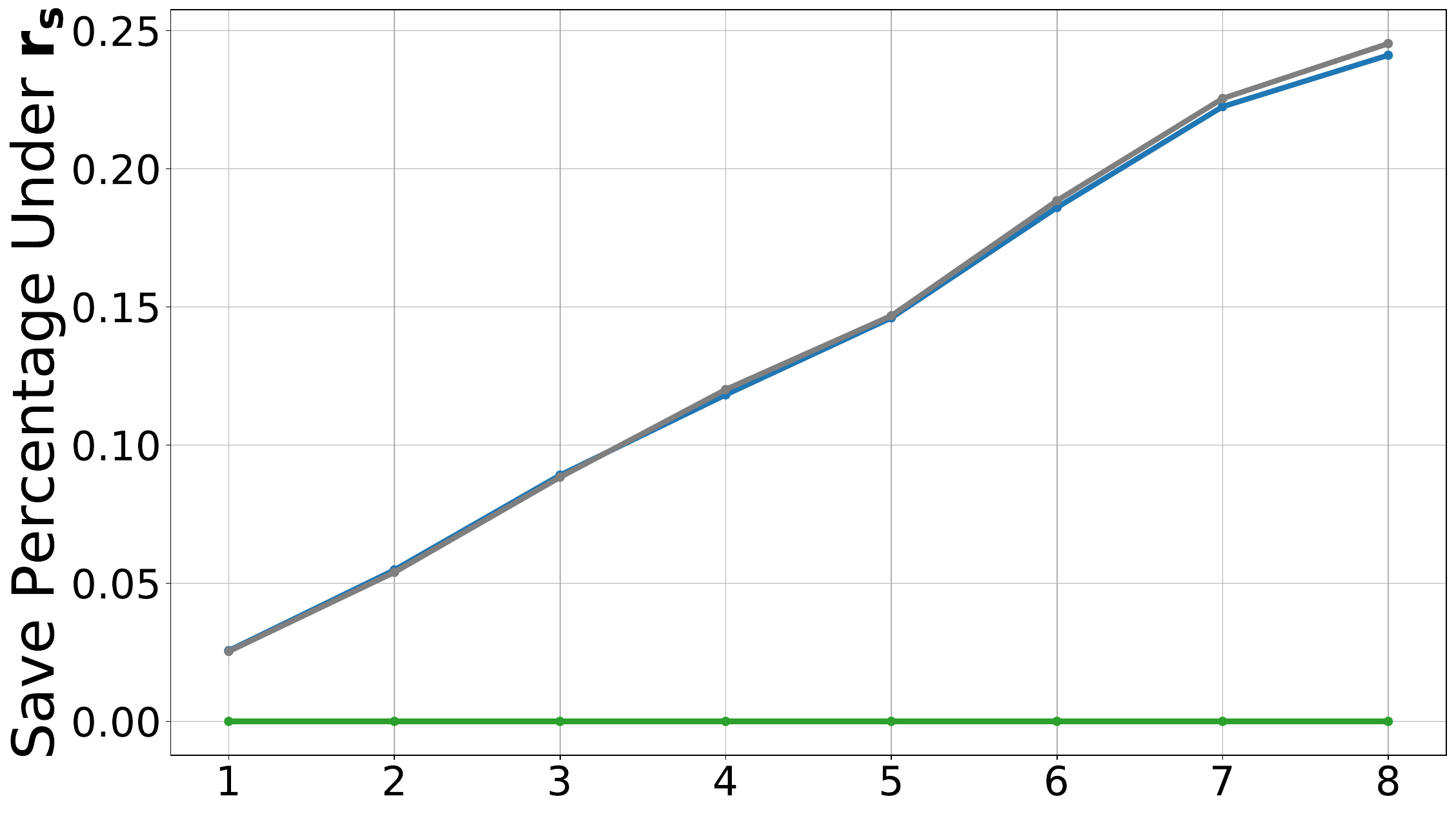}
    }
    \hfill
    \subfigure{
    \label{fig:356m}
        \includegraphics[width=0.31\textwidth]{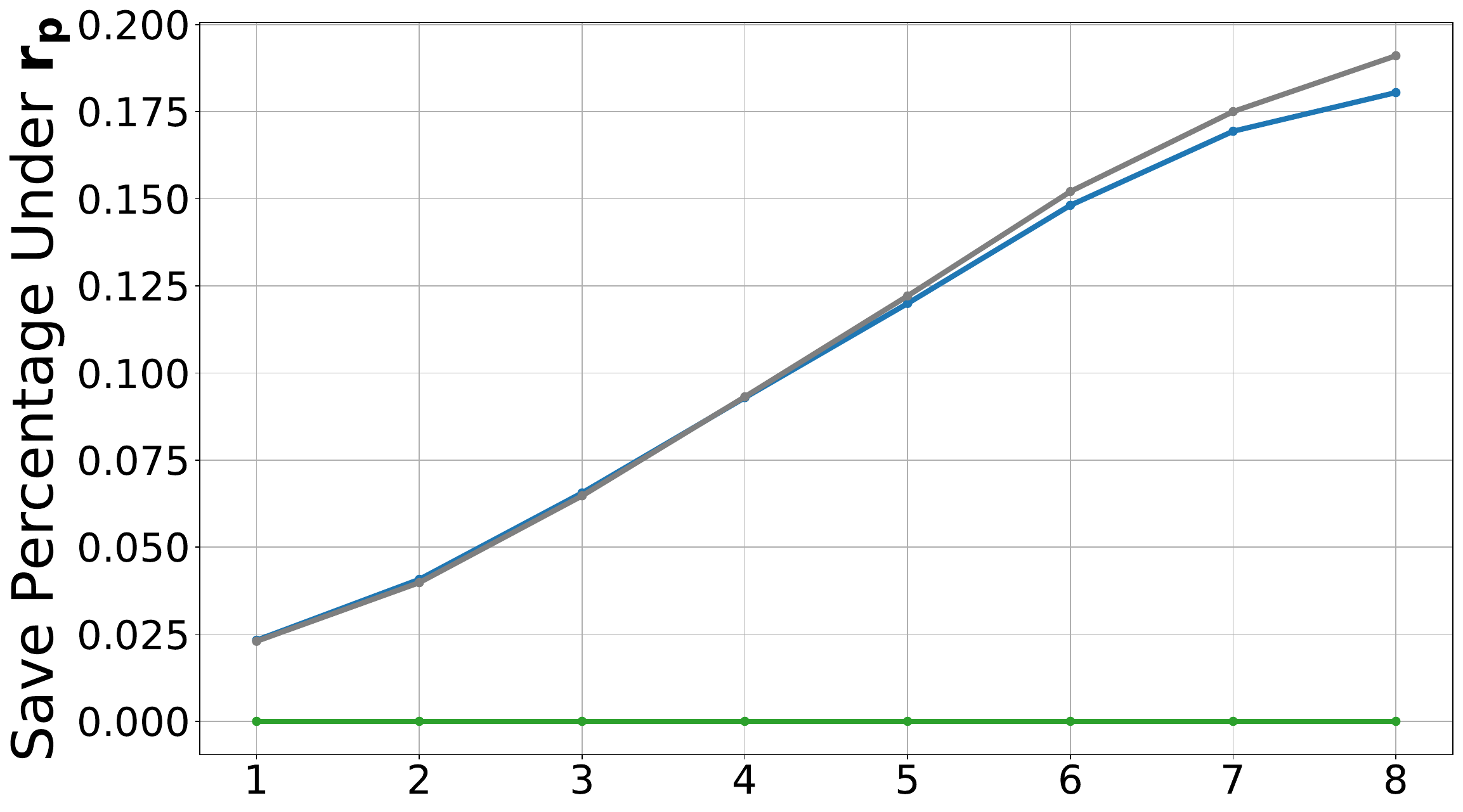}
    }
    \vspace{-0.2cm}
    \caption{Results of \textsc{UniCBE} with different $\theta_{acc}$. }
    \vspace{-0.4cm}
    \label{fig:theta}
\end{figure}

In ~\eqref{eq:14}, we integrate sampling matrices targeting different optimization objectives with equal weights. In practice, when faced with varying requirements, it is straightforward to prioritize a specific objective by adjusting the weights \(\theta_{acc}\), \(\theta_{con}\), and \(\theta_{sca}\) for these matrices, as shown in ~\eqref{eq:theta}. 
\begin{equation}
    P^{l} = \frac{(P^{acc\text{-}l})^{\theta_{acc}} \circ (P^{con\text{-}l})^{\theta_{con}}  \circ (P^{sca\text{-}l})^{\theta_{sca}} }{\sum ((P^{acc\text{-}l})^{\theta_{acc}} \circ (P^{con\text{-}l})^{\theta_{con}}  \circ (P^{sca\text{-}l})^{\theta_{sca}})}
    \label{eq:theta}
\end{equation}
As demonstrated in Table~\ref{tab:beta_theta}, we set different settings and calculate the degree of achievement \(\beta\) for each optimization objective following the procedure described in \S\ref{app:beta}. Compared to equal-weight integration, users can increase the corresponding \(\beta\) (e.g., \(\beta_{acc}\)) by assigning a larger weight to a specific optimization objective (\(\theta_{acc}\)), thereby better meeting their practical needs (accuracy).
We also observe that enhancing a specific optimization objective often comes with a slight decrease in the achievement of other objectives. In Figure~\ref{fig:theta}, we illustrate an example of improving accuracy, where \(\theta_{acc}\) is increased from 1 to 2. We find that the increased focus on accuracy objective slightly slows down the convergence speed. As a result, when \(T\) is relatively small, the performance of \(\theta_{acc} = 2\) lags behind that of \(\theta_{acc} = 1\). However, in the later stages, after convergence, the enhanced accuracy objective enables \(\theta_{acc} = 2\) to outperform \(\theta_{acc} = 1\), resulting in greater savings in the preference budget.
\begin{table}[h]
\renewcommand\arraystretch{1.4}
\centering
\setlength{\tabcolsep}{0.6em} 
\vspace{-0.1cm}
\caption{The measurement results of the achievement of objectives in \S\ref{sec:3} for \textsc{UniCBE} with varied hyperparameters.}
\begin{tabular}{lcccc}
\toprule
\multirow{3}{*}{Settings}&$\theta_{acc} = 2$&$\theta_{acc} = 1$&$\theta_{acc} = 1$&$\theta_{acc} = 1$\\
\multirow{3}{*}{}&$\theta_{con} = 1$&$\theta_{con} = 2$&$\theta_{con} = 1$&$\theta_{con} = 1$\\
\multirow{3}{*}{}&$\theta_{sca} = 1$&$\theta_{sca} = 1$&$\theta_{sca} = 2$&$\theta_{sca} = 1$\\
\midrule
$\beta_{acc}$&.7380\tiny \textcolor{mygreen}{(+.0016)}&.7355\tiny \textcolor{red}{(-.0009)}&.7351\tiny \textcolor{red}{(-.0013)}&\textbf{.7364}\\
$\beta_{con}$&.9221\tiny \textcolor{red}{(-.0007)}&.9235\tiny \textcolor{mygreen}{(+.0007)}&.9217\tiny \textcolor{red}{(-.0011)}&\textbf{.9228}\\
$\beta_{sca}$&.9996\tiny \textcolor{red}{(-.0001)}&.9997\tiny \textcolor{gray}{(.0000)}&.9998\tiny \textcolor{mygreen}{(+.0001)}&\textbf{.9997}\\
\bottomrule
\end{tabular}
\vspace{-0.4cm}
\label{tab:beta_theta}
\end{table}

\section{Prompt for Having an LLM Act as Judge}
\label{app:prompt}
We follow AlpacaEval\footnote{\url{https://github.com/tatsu-lab/alpaca_eval}} to instruct the LLMs act as judge with the following prompt:

\begin{quote}
{\itshape
You are a helpful assistant, that ranks models by the quality of their answers.
\\
\\
I want you to create a leaderboard of different of large-language models. To do so, I will give you the instructions (prompts) given to the models, and the responses of two models. Please give the winner model based on which responses would be preferred by humans. All inputs and outputs should be python dictionaries.
\\
\\
Here is the prompt:
\\
\{
\\
    "instruction": """\{instruction\}""",
\\
\}
\\
\\
Here are the outputs of the models:
\\
    \{
\\
        "model": "$model_1$",
\\
        "answer": """\{$output_1$\}"""
\\
    \},
\\
    \{
\\
        "model": "$model_2$",
\\
        "answer": """\{$output_2$\}"""
\\
    \}
\\
\\
Now please give the winner model according to the quality of their answers, so that the winner model has the best output. Then return the winner model in the following format if $model_x$ is the winner:
winner: $model_x$
\\
\\
You need to strictly follow the format above. Please provide the ranking that the majority of humans would give.
}
\end{quote}

\section{The Calculation Process of $\beta$ in Table~\ref{tab:beta}}
\label{app:beta}
%我们计算各CBE方法的$\beta$值来衡量其在多大程度上贴合了第三节我们所分析出的优化目标：ensures uniformity across tuples, uniformity across model pairs in win-rate uncertainty, and uniformity across model。
We calculate the $\beta$ values for each CBE method to measure how well they align with the optimization objectives we analyzed in \S\ref{sec:3}: ensuring uniformity across tuples, uniformity across model pairs in win-rate uncertainty, and uniformity across models.
Specifically, we first construct $U^{acc1}$, $U^{acc2}$, $U^{con}$ and $U^{sca}$ as follows:
\begin{equation}
\begin{gathered}
    U^{acc1}_{i,j} = \begin{cases} 0, \ \ \rm{if}\ \ i=j\\
    \frac{1}{M(M-1)}, \ \ \rm{else}
    \end{cases}\\
    U^{acc2}_{i,k} = \frac{1}{MN}
\end{gathered}
\end{equation}
\begin{equation}
    U^{con}_{i,j} = \begin{cases} 0, \ \ \rm{if}\ \ i=j\\
    \frac{1}{M(M-1)}, \ \ \rm{else}
    \end{cases}
\end{equation}
\begin{equation}
    U^{sca}_{i} = \frac{1}{M}
\end{equation}
On this basis, we calculate $\beta_{acc}$, $\beta_{con}$ and $\beta_{sca}$ as follows:
\begin{equation}
    \beta_{acc} = \rm{CosineSim}(U^{acc1},C.\rm{mean}(\rm{dim}=-1))\times \rm{CosineSim}(U^{acc2},C.\rm{mean}(\rm{dim}=0))
\end{equation}
\begin{equation}
    \beta_{con} = \rm{CosineSim}(U^{con},\epsilon)
\end{equation}
\begin{equation}
    \beta_{sca} = \rm{CosineSim}(U^{sca},C.\rm{mean}(\rm{dim}=-1).\rm{mean}(\rm{dim}=-1))
\end{equation}

\section{Further Discussions}
\subsection{Affinity of the Three Optimization Objectives}
\label{app:dis-aff}
We have discussed in \S\ref{sec:3} that the keys to strengthen the accuracy, convergence and scalability of CBE are: ensuring uniformity across tuples, uniformity across model pairs in win-rate uncertainty, and uniformity across models.
%我们以下来讨论他们的兼容性关系。首先，ensuring uniformity across models 可以视作是 ensuring uniformity across tuples的一个子目标，二者有着良好的亲和度。前者可以视作对于后者在model uniformity方面的一个加强。
Below we discuss their compatibility. Firstly, ensuring uniformity across models can be seen as a sub-goal of ensuring uniformity across tuples, therefore they exhibit a strong affinity. The former can be considered a refinement of the latter, specifically focusing on model uniformity. 
%此外，从公式2可以看出，模型间胜率的不确定度和模型间的比较次数是呈反比的，因此提升uniformity across model pairs in win-rate uncertainty也将有助于提升模型间的比较次数的均匀性。这进一步意味着确保uniformity across model pairs in win-rate uncertainty和ensuring uniformity across tuples you着兼容的目标和良好的亲和性。
Moreover, as shown in\eqref{eq:9}, the uncertainty in win-rate between models $\epsilon_{i,j}$ is inversely proportional to the number of comparisons between models. Therefore, improving uniformity across model pairs in win-rate uncertainty $\epsilon$ will also contribute to a more uniform distribution of comparisons between models. This further implies that ensuring uniformity across model pairs in win-rate uncertainty and ensuring uniformity across tuples are compatible goals with a strong affinity.
%此外从Table2我们可以看出相比于基线SaAcCon的\beta都得到了提升，这也从实验角度验证了优化三个目标可以相互增益的事实。基于以上分析，我们可推断改变$\alpha$其实是在改变对不同目标的优化力度。但由于三者是相互促进的，因此改变$\alpha$所带来的效果就会很小。
Furthermore, as shown in Table~\ref{tab:beta}, all $\beta$ values of \textsc{UniCBE} are improved compared to the baselines, experimentally validating the fact that optimizing the three objectives can be mutually beneficial. Based on this analysis, we can infer that changing $\alpha$ essentially adjusts the optimization emphasis on different objectives. However, since the three objectives are mutually reinforcing, the effect of changing $\alpha$ will be relatively small.

\subsection{Discussion on Sampling Bias in Incomplete Sampling Scenarios}

Previous studies have discussed the risks of introducing sampling bias in incomplete sampling scenarios. Specifically, \citet{samplebias1} demonstrated through simulation experiments that K-fold cross-validation (K-fold CV) can produce significant performance estimation bias when dealing with small sample sizes. This bias persists even when the sample size reaches 1000. In contrast, methods like nested cross-validation (Nested CV) and train/test split have been shown to provide robust and unbiased performance estimates regardless of sample size. \citet{samplebias2} introduced a weighting scheme, as described in \citep{samplebias3}, to mitigate sampling bias in active testing scenarios. \citet{AP} proposed leveraging information obtained from source models to select representative samples from the test set, thereby reducing sampling bias. Additionally, \citet{tiny} employed Item Response Theory \citep{irt} to correct sample bias in addressing this issue.

These studies inspired us to investigate the bias problem in the CBE scenario. Unlike the aforementioned studies, we found that in CBE scenario, not only does sample bias exist, but model bias also plays a role, and the two are coupled. This coupling poses greater challenges for analyzing and mitigating these biases. To address this, based on the analyses outlined in \S\ref{sec:3}, we propose the \textsc{UniCBE} method, which effectively alleviates biases in this scenario.

\begin{figure}[htbp]
    \centering
    \subfigure{
    \label{fig:351}
        \includegraphics[width=0.31\textwidth]{figs/performance-full-pre_wins_theta.pdf}
    }
    \hfill
    \subfigure{
    \label{fig:352}
        \includegraphics[width=0.31\textwidth]{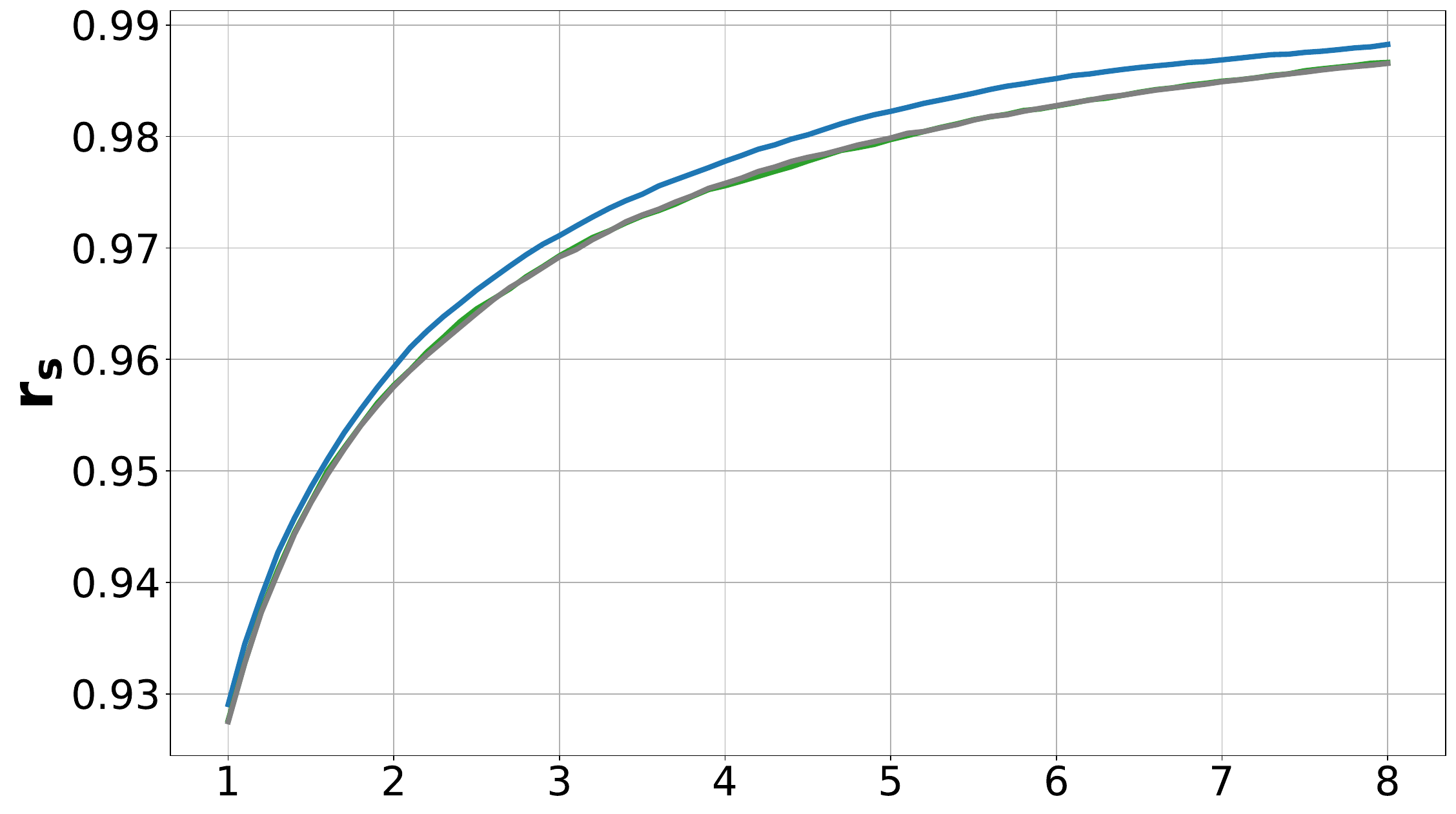}
    }
    \hfill
    \subfigure{
    \label{fig:353}
        \includegraphics[width=0.31\textwidth]{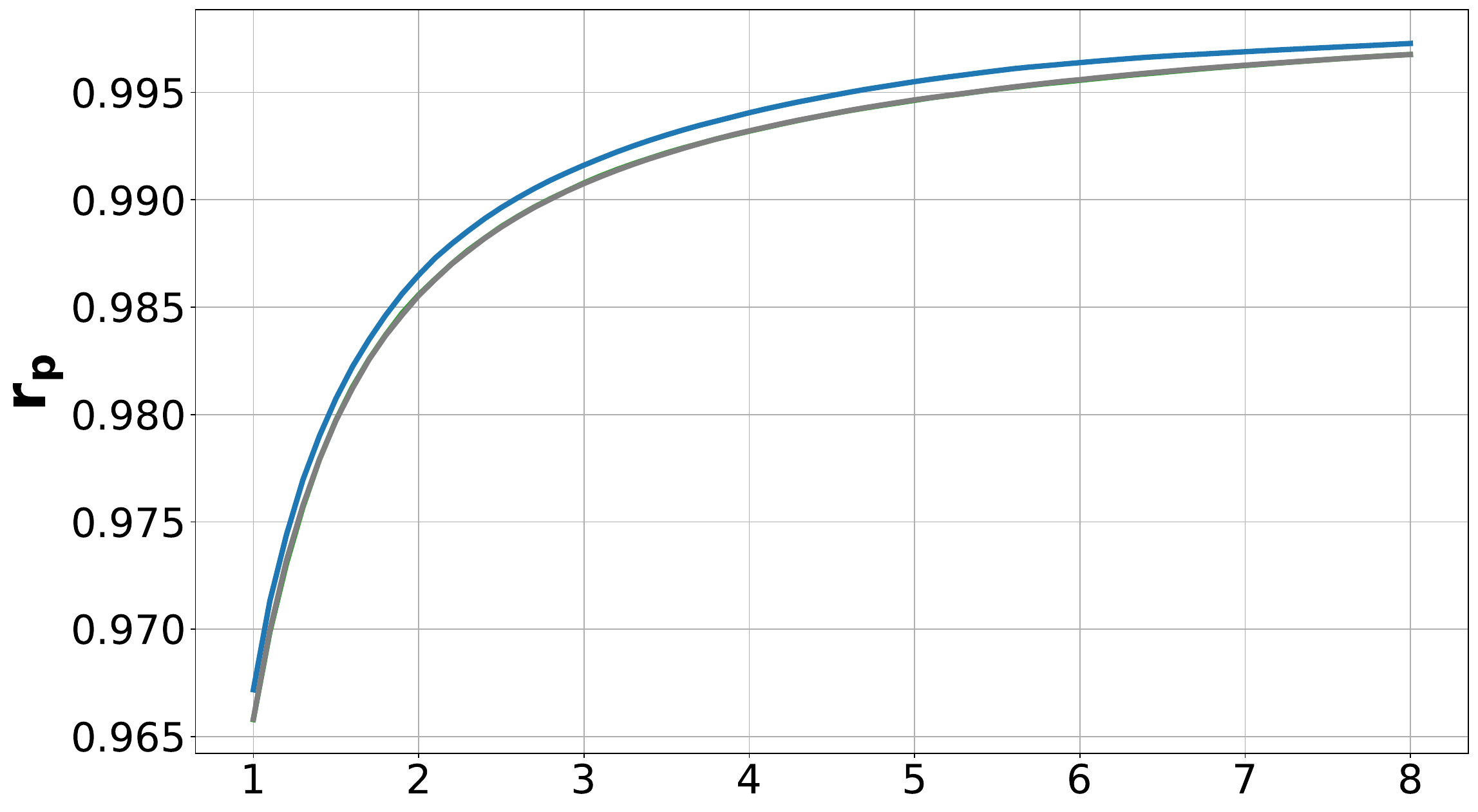}
    } \\
    \subfigure{
    \label{fig:354}
        \includegraphics[width=0.31\textwidth]{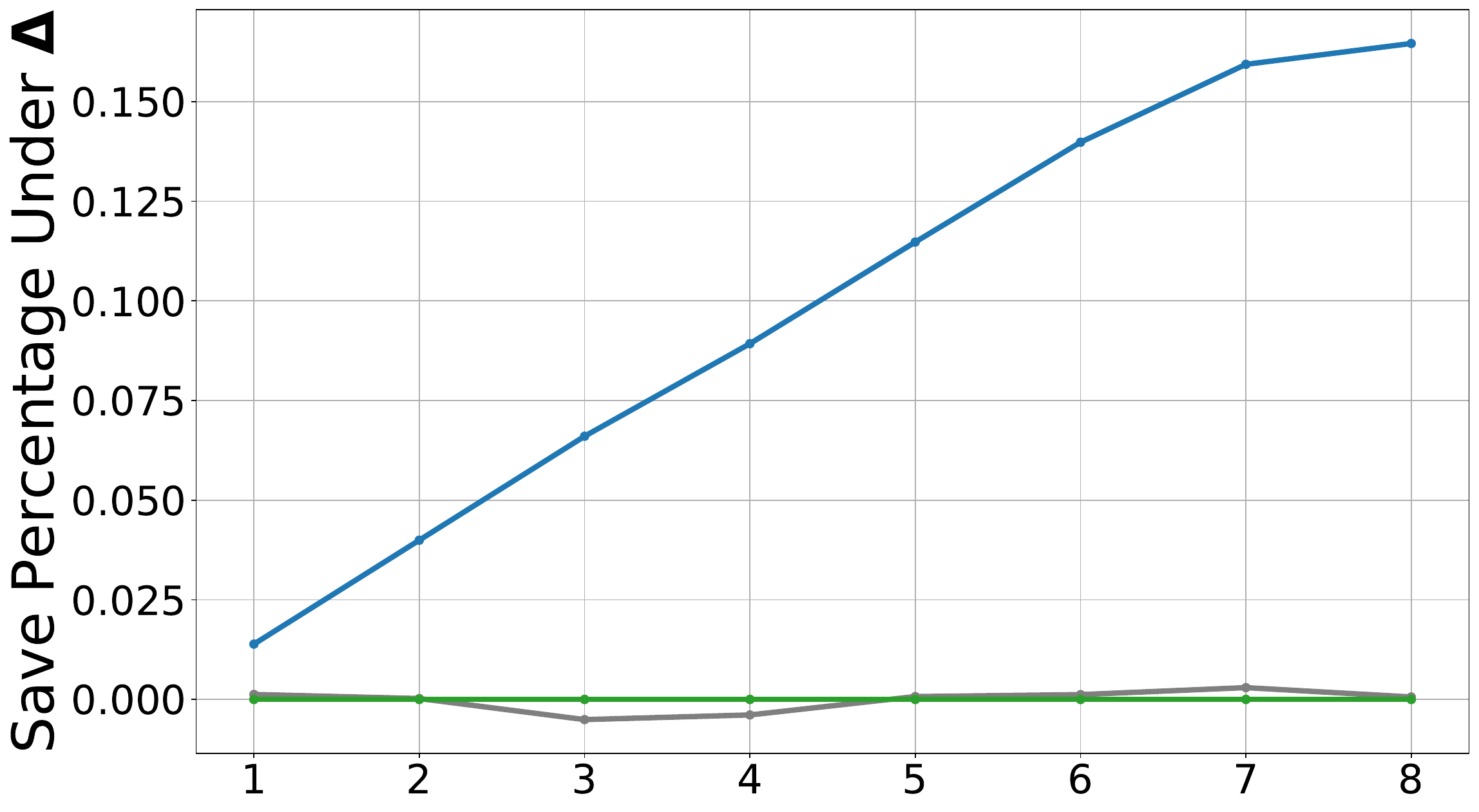}
    }
    \hfill
    \subfigure{
    \label{fig:355}
        \includegraphics[width=0.31\textwidth]{figs/save-full-s.pdf}
    }
    \hfill
    \subfigure{
    \label{fig:356}
        \includegraphics[width=0.31\textwidth]{figs/save-full-p.pdf}
    }
    \caption{Results of compared CBE methods with GPT-3.5-turbo as the judge on AlpacaEval. }
    \label{fig:main-35}
\end{figure}

\begin{figure}[htbp]
    \centering
    \subfigure{
    \label{fig:mt1}
        \includegraphics[width=0.31\textwidth]{figs/performance-mt-pre_wins_theta.pdf}
    }
    \hfill
    \subfigure{
    \label{fig:mt2}
        \includegraphics[width=0.31\textwidth]{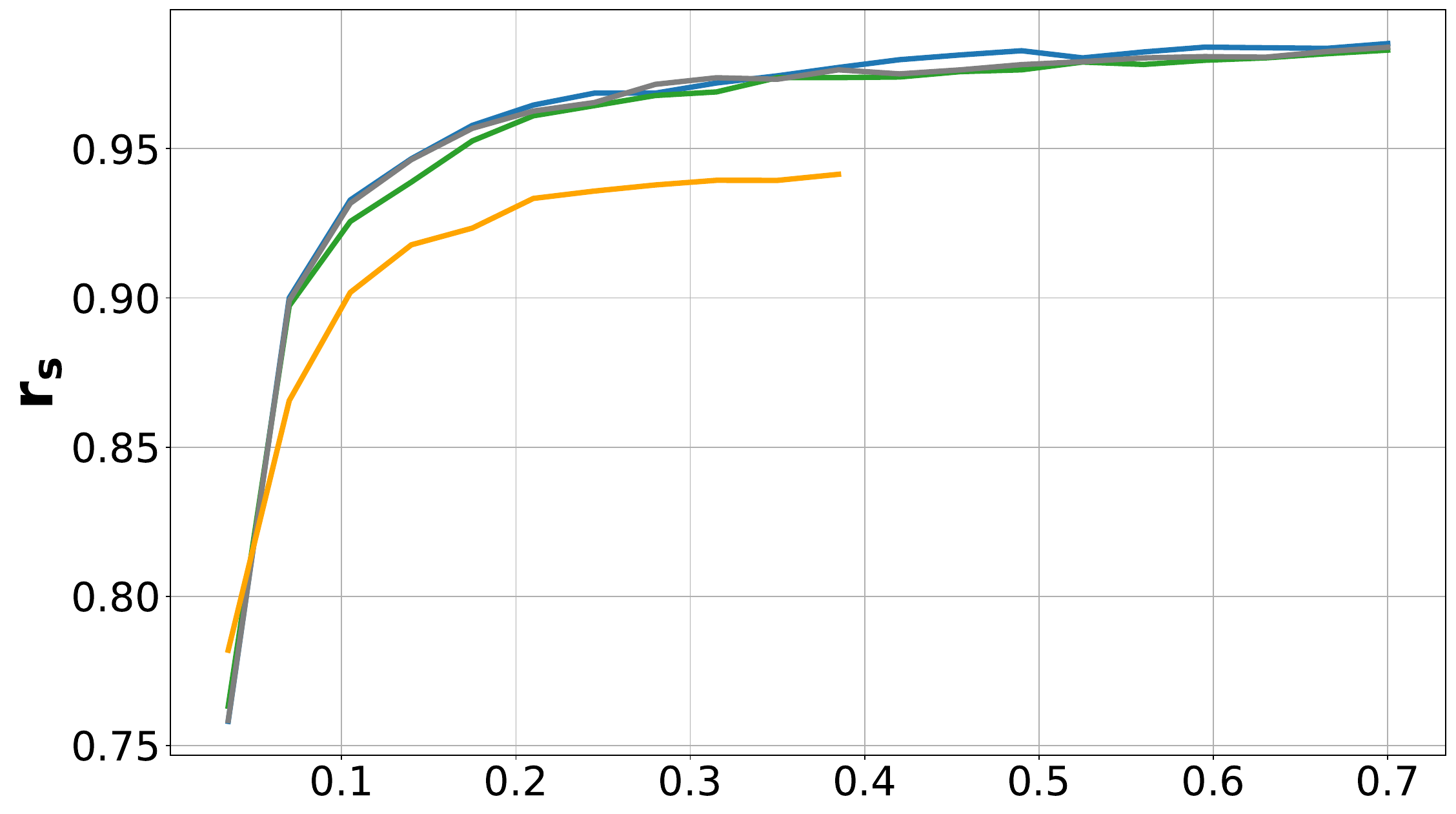}
    }
    \hfill
    \subfigure{
    \label{fig:mt3}
        \includegraphics[width=0.31\textwidth]{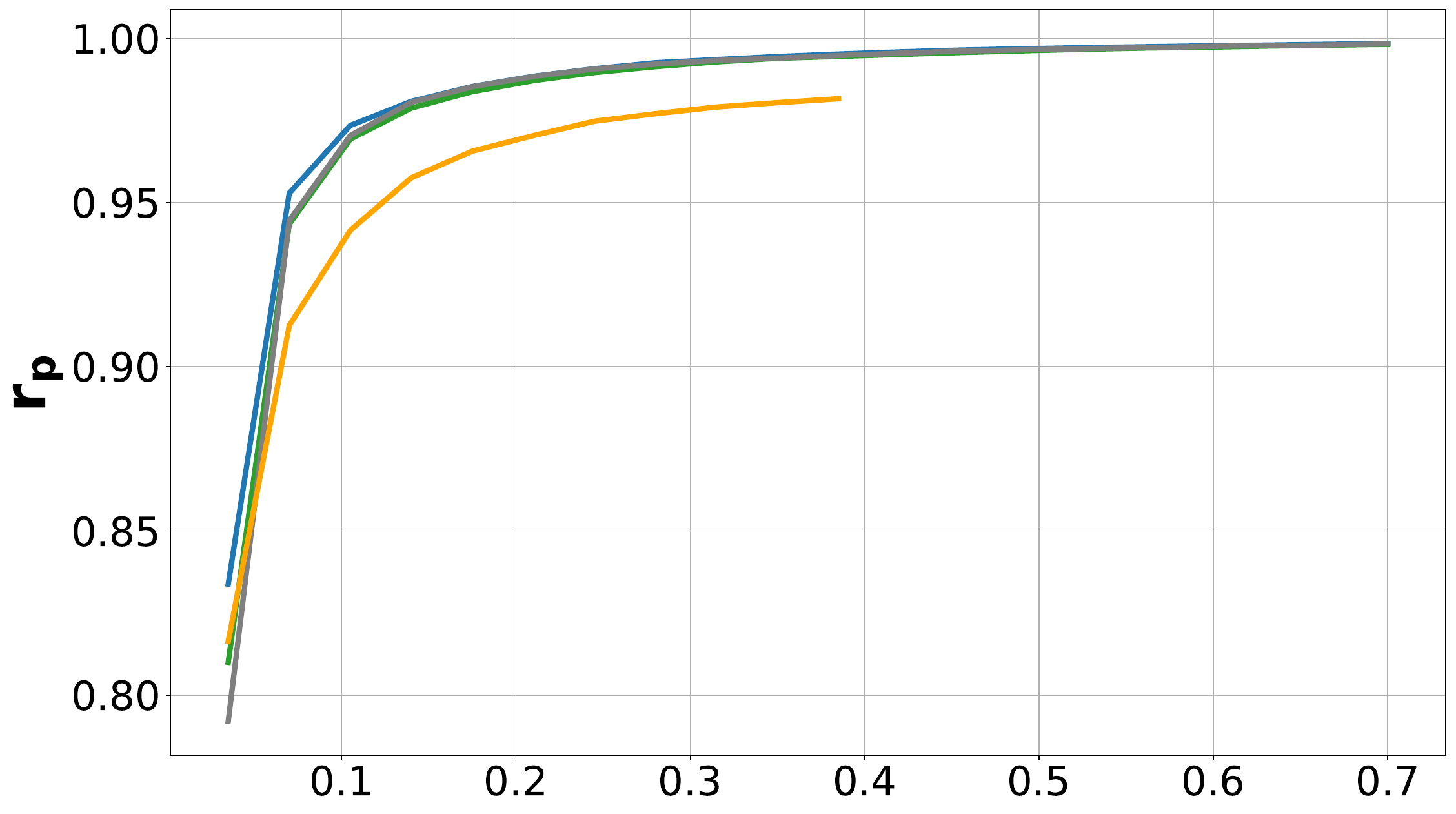}
    } \\
    \subfigure{
    \label{fig:mt4}
        \includegraphics[width=0.31\textwidth]{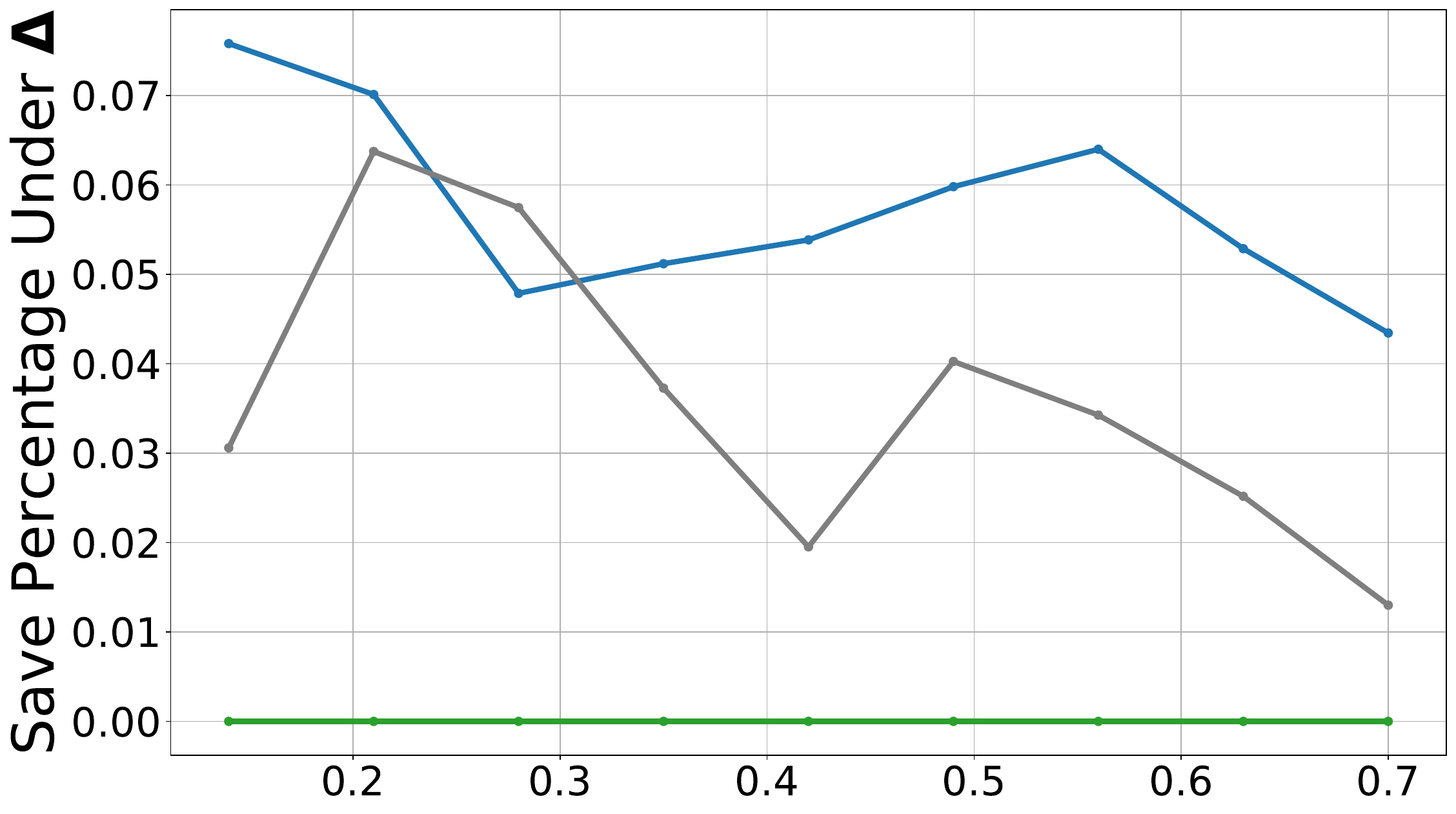}
    }
    \hfill
    \subfigure{
    \label{fig:mt5}
        \includegraphics[width=0.31\textwidth]{figs/save-mt-s.pdf}
    }
    \hfill
    \subfigure{
    \label{fig:mt6}
        \includegraphics[width=0.31\textwidth]{figs/save-mt-p.pdf}
    }
    \caption{Results of compared CBE methods with human as the judge on MT-Bench.}
    \label{fig:main-mt}
\end{figure}

\begin{figure}[h]
    \centering
    \subfigure{
    \label{fig:351m}
        \includegraphics[width=0.31\textwidth]{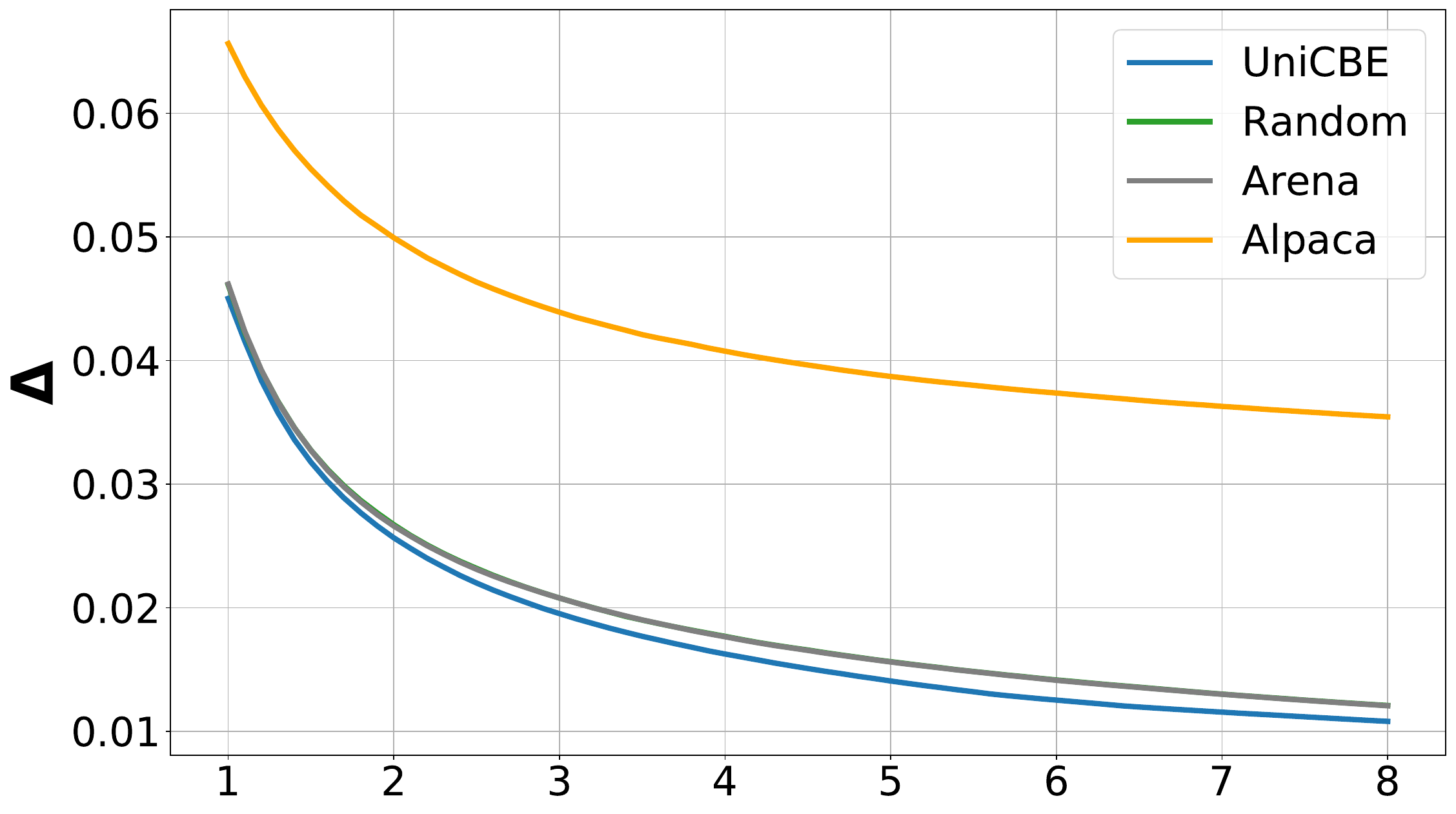}}
    \hfill
    \subfigure{
    \label{fig:355m}
        \includegraphics[width=0.31\textwidth]{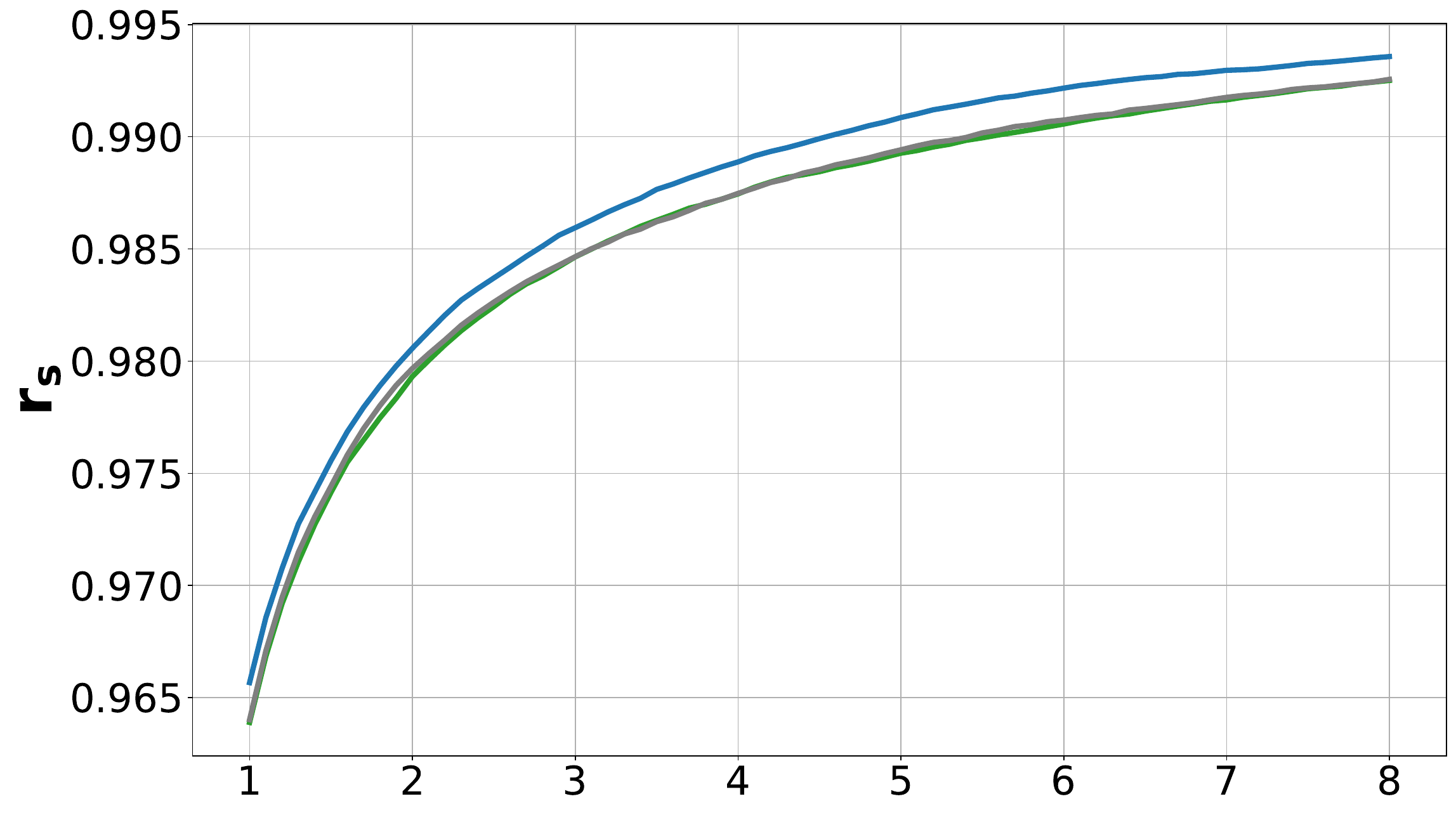}
    }
    \hfill
    \subfigure{
    \label{fig:356m}
        \includegraphics[width=0.31\textwidth]{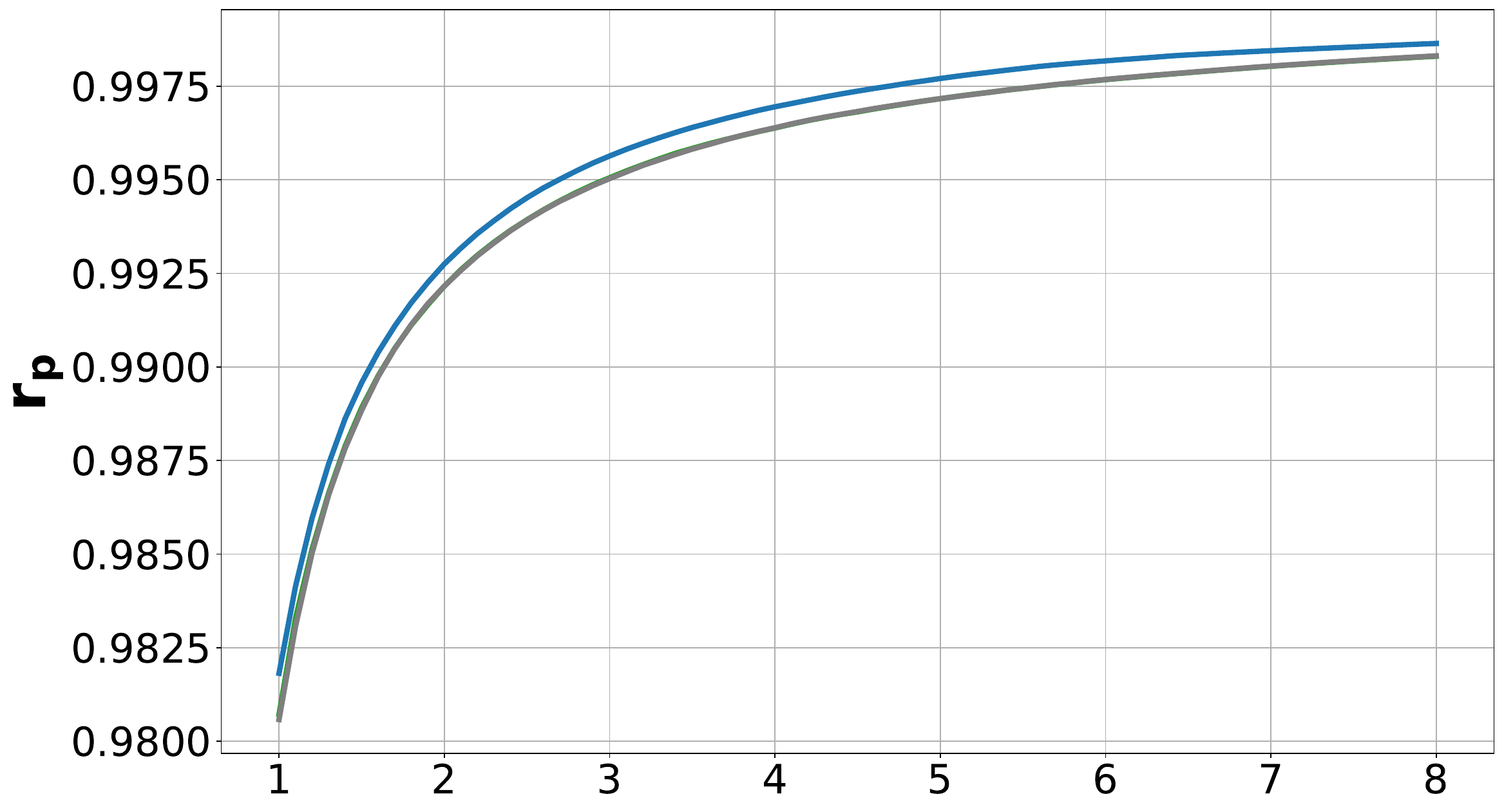}
    } \vspace{-0.2cm}\\
    \subfigure{
    \label{fig:mt1m}
        \includegraphics[width=0.31\textwidth]{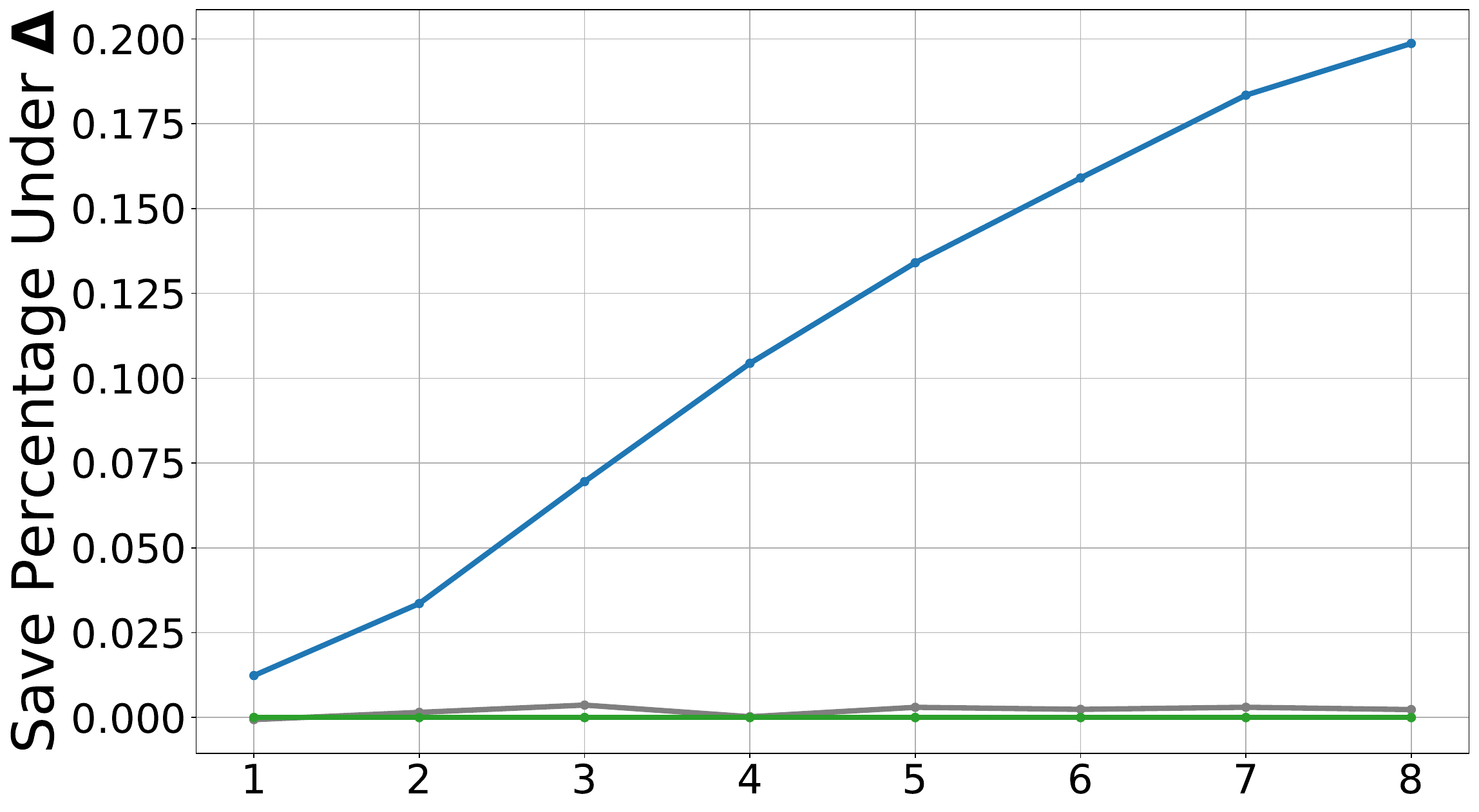}
    }
    \hfill
    \subfigure{
    \label{fig:mt5m}
        \includegraphics[width=0.31\textwidth]{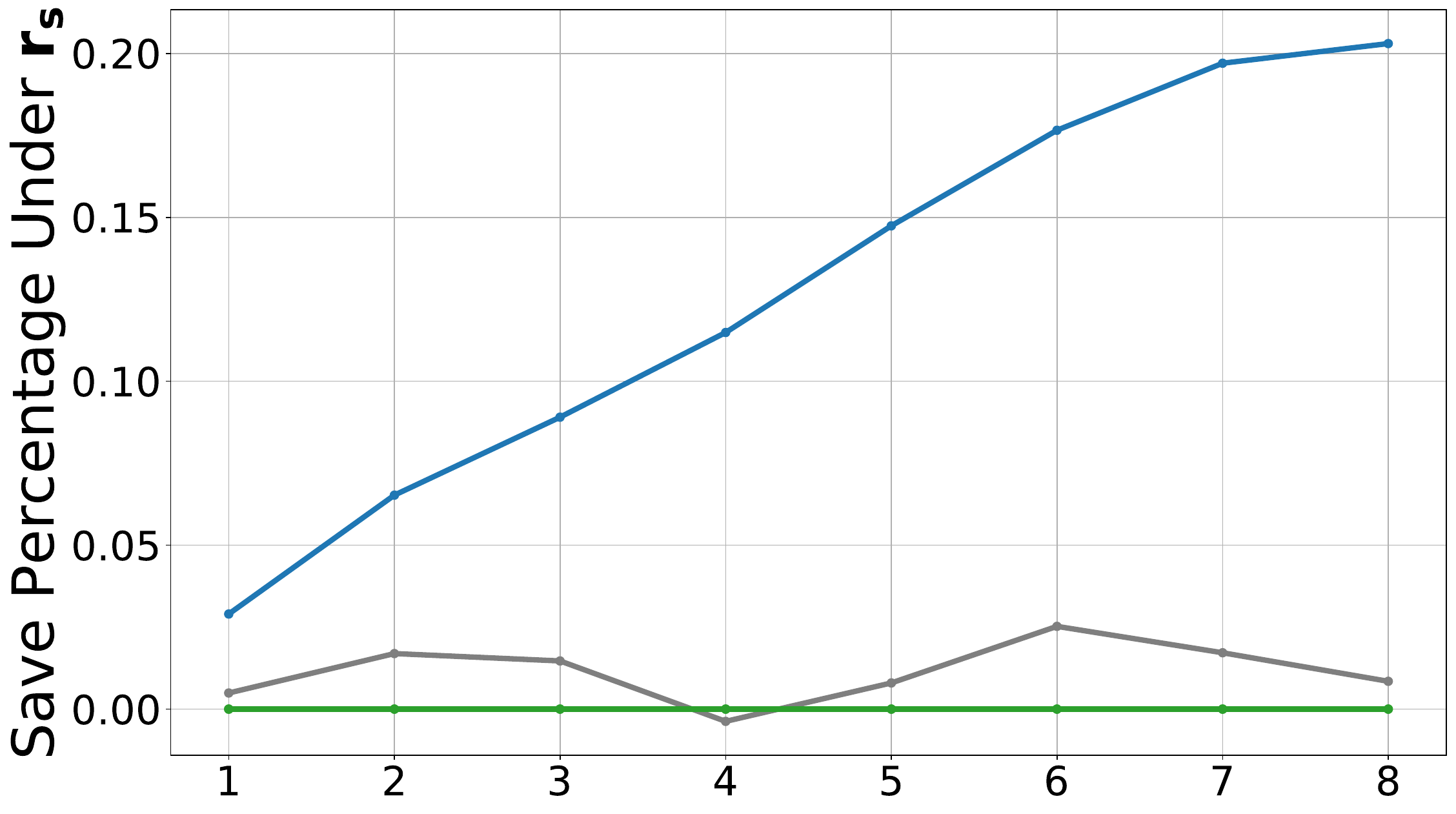}
    }
    \hfill
    \subfigure{
    \label{fig:mt6m}
        \includegraphics[width=0.31\textwidth]{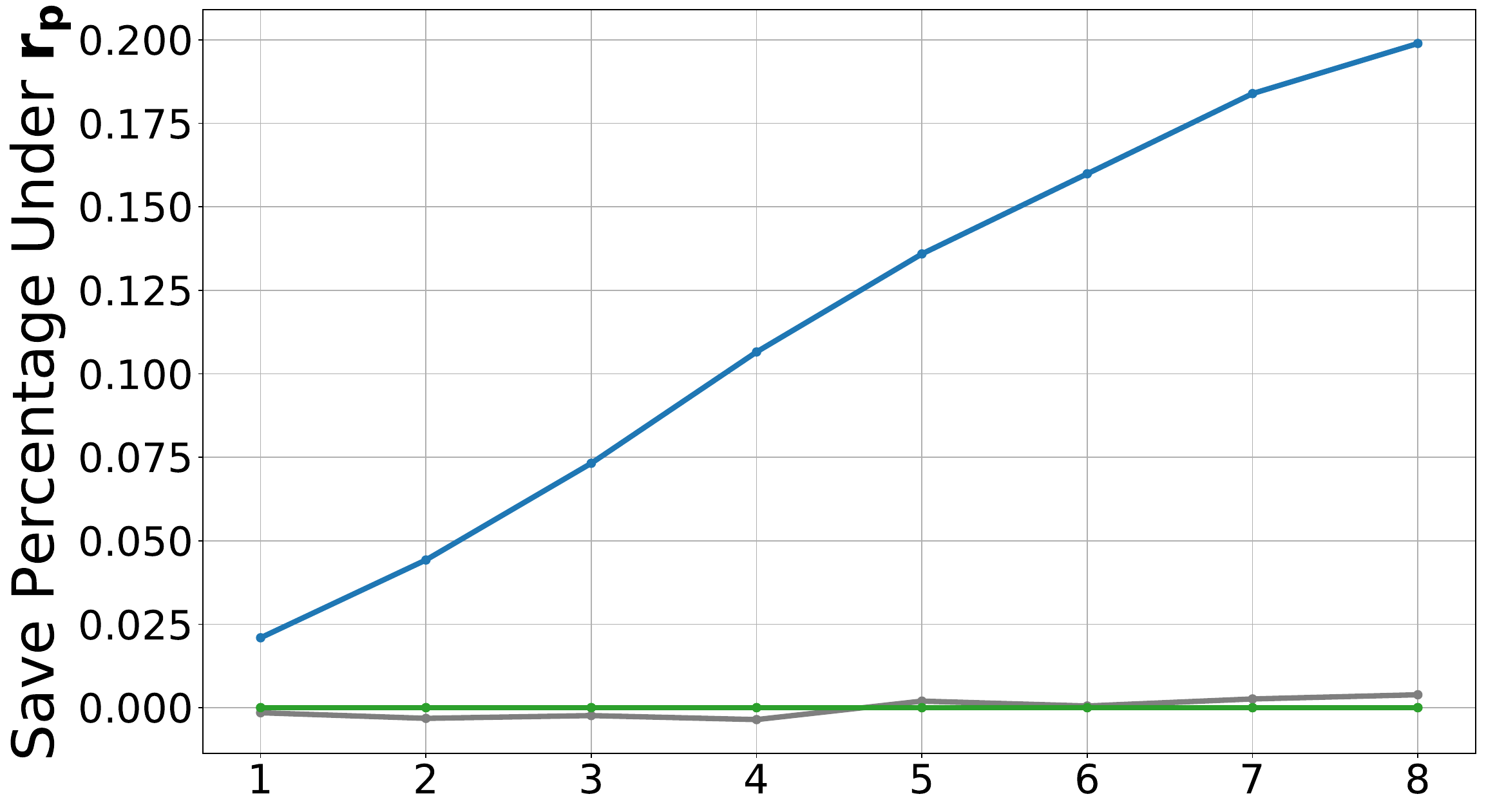}
    }
    \vspace{-0.2cm}
    \caption{Results of compared CBE methods with Qwen-Plus as the judge on AlpacaEval. }
    \vspace{-0.3cm}
    \label{fig:qwen}
\end{figure}

\end{document}